\documentclass[10pt,twocolumn,letterpaper]{article}

\usepackage[pagenumbers]{iccv} 

\usepackage[dvipsnames]{xcolor}

\usepackage{multirow}
\usepackage{tabularx}
\usepackage{booktabs}
\usepackage{makecell}
\usepackage{pifont}
\usepackage{graphicx}
\usepackage{tablefootnote}
\usepackage{float}
\usepackage{colortbl}
\usepackage{svg}
\usepackage{marvosym}
\definecolor{iccvblue}{rgb}{0.21,0.49,0.74}
\usepackage[pagebackref,breaklinks,colorlinks,allcolors=iccvblue]{hyperref}

\def\paperID{14650} 
\def\confName{ICCV}
\def\confYear{2025}

\title{GauS-SLAM: Dense RGB-D SLAM with Gaussian Surfels}
 
\author{Yongxin Su$^{1}$ \qquad Lin Chen$^{2}$ \qquad Kaiting Zhang$^{1}$ \qquad {Zhongliang Zhao}\textsuperscript{1,\Letter}  \\[4pt]
Chenfeng Hou$^{1}$ \qquad Ziping Yu$^{1}$ \\[4pt]
$^{1}$ Beihang University \qquad $^{2}$ Northwestern Polytechnical University \\
\url{https://gaus-slam.github.io}
}

\definecolor{top1}{HTML}{BFE8C1}
\definecolor{top2}{HTML}{FFF5B3}
\definecolor{top3}{HTML}{FFD9B3}

\begin{document}
\twocolumn[{%
\renewcommand\twocolumn[1][]{#1}%
\maketitle
\begin{center}
    \centering
    \captionsetup{type=figure}
    \includegraphics[width=\linewidth]{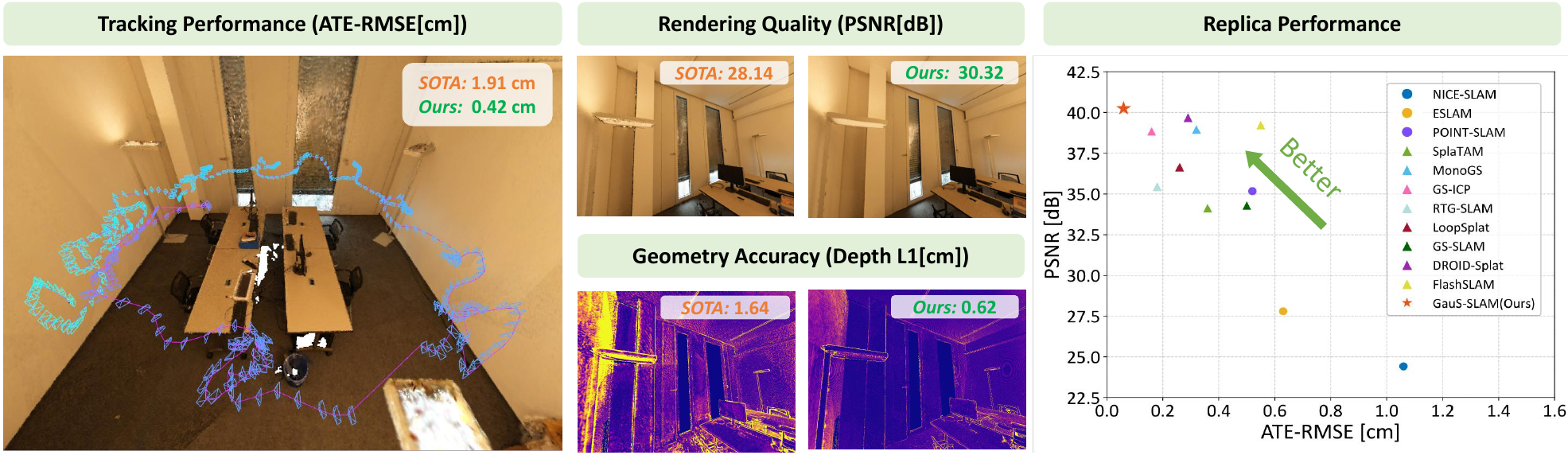}
    \captionof{figure}{GauS-SLAM is a dense SLAM system using 2D Gaussian surfels\cite{2DGS}, capable of simultaneously achieving high-precision localization and high-fidelity  reconstruction. As shown in the left figure, GauS-SLAM exhibits millimeter-level tracking accuracy on a challenging real-world scenario(\texttt{b20a261fdf} in ScanNet++\cite{scannetpp} dataset), significantly outperforming the SOTA approach, SplaTAM\cite{SplaTAM}. The right figure demonstrates GauS-SLAM's SOTA performance on the Replica\cite{Replica} dataset, achieving an absolute trajectory error(ATE-RMSE) of $0.06cm$ and 40.25 dB in rendering quality.}
    \label{fig:reconstruction-fig}
\end{center}%
}]

\maketitle

\begin{abstract}
We propose GauS-SLAM, a dense RGB-D SLAM system that leverages 2D Gaussian surfels to achieve robust tracking and high-fidelity mapping. Our investigations reveal that Gaussian-based scene representations exhibit geometry distortion under novel viewpoints, which significantly degrades the accuracy of Gaussian-based tracking methods. These geometry inconsistencies arise primarily from the depth modeling of Gaussian primitives and the mutual interference between surfaces during the depth blending. To address these, we propose a 2D Gaussian-based incremental reconstruction strategy coupled with a Surface-aware Depth Rendering mechanism, which significantly enhances geometry accuracy and multi-view consistency. Additionally, the proposed local map design dynamically isolates visible surfaces during tracking, mitigating misalignment caused by occluded regions in global maps while maintaining computational efficiency with increasing Gaussian density. Extensive experiments across multiple datasets demonstrate that GauS-SLAM outperforms comparable methods, delivering superior tracking precision and rendering fidelity. The project page will be made available at https://gaus-slam.github.io.

\end{abstract}    
\section{Introduction}
\label{sec:intro}




Over the past decade, dense visual simultaneous localization and mapping (SLAM) has remained a fundamental research area in computer vision. Recent advances in map representation have increasingly focused on integrating neural models with traditional 3D features—such as points, voxels, and surfels—enabling more flexible and accurate map construction. Despite these innovations, current methods still face considerable challenges in areas like pose optimization, convergence difficulties, and catastrophic forgetting during continual learning.

Explicit representation using 3D Gaussian Splatting\cite{3DGS} (3DGS) has shown promising results in both 3D reconstruction and dense SLAM. Pioneering works \cite{MonoGS, SplaTAM, Gaussian-SLAM, GS-SLAM} propose tracking and mapping pipelines based on Gaussian representation. However, these Gaussian-based tracking approaches often suffer from inaccurate pose estimation and convergence problems. In contrast, some advances \cite{Photo-SLAM, DROID-Splat, HI-SLAM2, RP-SLAM, g2mapping} address these issues by decoupling the tracking from the Gaussian model and leveraging mature odometry methods \cite{ORB-SLAM2, DROID-SLAM}. While this decoupled design improves real-time performance, it inherently lacks the mutual reinforcement between reconstruction and localization that a coupled system could offer.

\begin{figure}
    \centering
    \includegraphics[width=\linewidth]{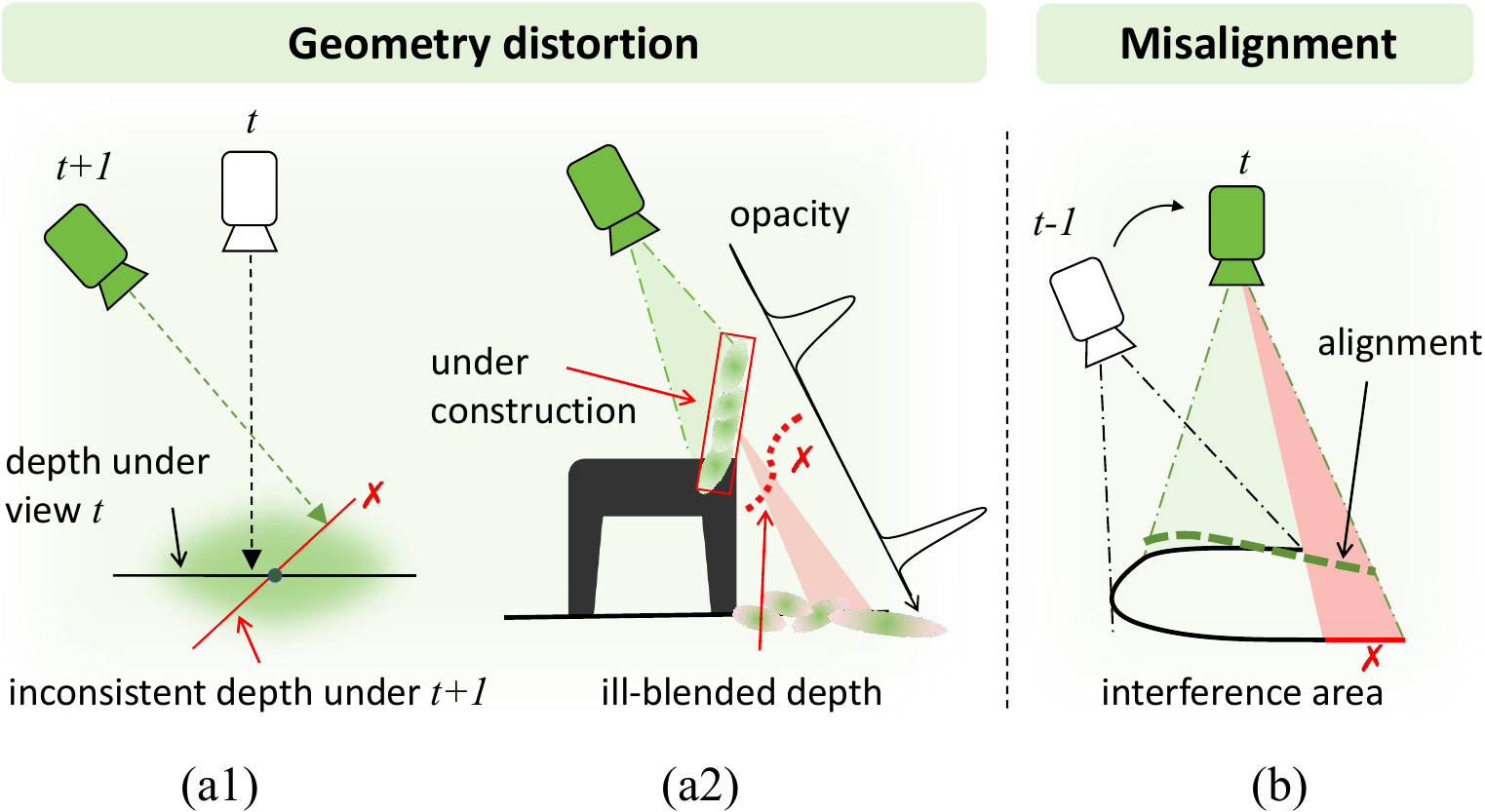}
    \caption{Two challenges in Gaussian-based tracking methods. (a1) illustrates geometry distortions caused by center depth model of the 3D Gaussian. (a2) shows ill-blended depth arising from depth rendering between different surfaces. (b) demonstrates that, during the alignment, certain interference area exhibit high accumulated opacity making them challenging to be masked out as outliers.}
    \label{fig:mutliview_challenges}
\end{figure}

In this paper, we focus on two critical challenges encountered in coupled Gaussian-based SLAM frameworks, as illustrated in \cref{fig:mutliview_challenges}. The first issue is geometry distortion. In most Gaussian-based tracking methods, the camera transformation is estimated by aligning the observation with the rendering result from the current viewpoint. During this process, perspective-induced geometry distortion leads to misalignment between the frame and the Gaussian model, consequently degrading the tracking accuracy.

We attribute geometry distortion to two primary factors. First, inherent inconsistencies exist in the Gaussian-based depth representation model, where the center depth model of 3D Gaussian primitives exhibits multi-view inconsistent depth estimations (as visualized in \cref{fig:mutliview_challenges}(a1)), whereas 2D Gaussian surfels effectively address this inconsistency through intersection depth model \cite{2DGS}.
The second arises from the mutual interference between different surfaces during depth blending. As illustrated in \cref{fig:mutliview_challenges}(a2), when rendering the backrest of the chair under reconstruction, the distant floor with greater depth results in ill-blended depth. To resolve depth blending ambiguities, we propose an incremental reconstruction strategy based on Gaussian surfels \cite{2DGS}, coupled with a Surface-aware Depth Rendering scheme, which significantly enhances the geometry accuracy and view consistency of the Gaussian scene.



The second challenge lies in the outlier removal during the alignment between frames and the Gaussian model. As demonstrated by SplaTAM\cite{SplaTAM}, outlier elimination is crucial, and their approach achieves this by masking regions with low accumulated opacity. However, as shown in Fig.\ref{fig:mutliview_challenges}(b), interference regions with high accumulated opacity remain challenging to mask, especially when the camera moves around the object. Our approach confines camera tracking to a small local map, thereby isolating these interference regions from the global map. Additionally, by periodically resetting the local map, we ensure that camera tracking always operates within a subset of Gaussian primitives, avoiding the degradation of tracking efficiency as the number of Gaussians increasing.



We propose GauS-SLAM, a dense SLAM system that leverages 2D Gaussian surfels within a tightly coupled front-end/back-end framework to address these challenges, achieving superior localization accuracy and view synthesis quality on RGB-D datasets. Our contributions are as follows:

\begin{itemize}
\item  We propose a 2D Gaussian-based incremental reconstruction strategy and Surface-aware Depth Rendering mechanism that effectively mitigates geometry distortions and improves tracking accuracy.
\item We propose a dense SLAM system with a front-end/back-end architecture, and incorporates a local map design that ensures tracking accuracy and efficiency.
\item We conduct extensive experiments that demonstrate the superiority of our approach, both in tracking accuracy and in reconstruction quality, compared to SOTA methods.
\end{itemize}

\begin{figure*}
\centering
\includegraphics[width=1\textwidth]{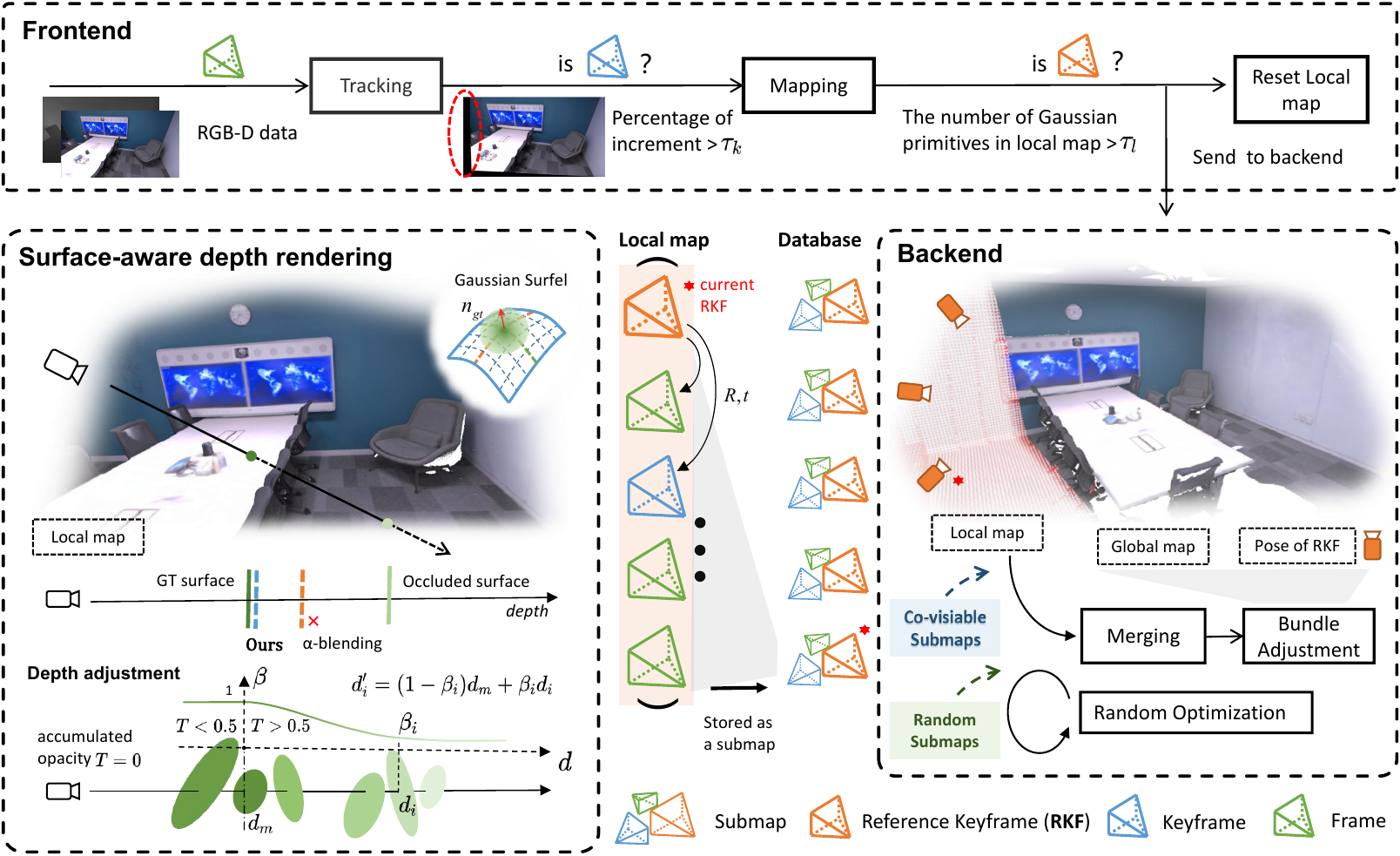}
\caption{\textbf{Overview of GauS-SLAM.} This framework consists of a front-end that performs tracking and mapping using a single local map, and a back-end responsible for merging the local map into the global map and submap-based global optimization.}
\label{fig:overview}
\end{figure*}

\section{Related Works}
\paragraph{Neural based dense SLAM}
Leveraging the powerful modeling capability of deep neural networks, \cite{Vox-Fusion, DI-Fusion, NeuralFusion} utilize implicit representation of 3D scenes through neural voxel and neural grid frameworks. iSDF\cite{iSDF} integrates neural networks with traditional Signed Distance Functions, enabling the adaptive adjustment of detail levels. iMap\cite{iMap} demonstrates that the MLP-based neural implicit representation can be effectively utilized for localization and reconstruction in dense SLAM. \cite{Nice-SLAM, ESLAM} employ NeRF\cite{NeRF} to represent 3D scenes, and introduce hierarchical multi-feature grids for coarse-to-fine optimization of camera poses. NeRF-SLAM \cite{NeRF-SLAM} integrates the end-to-end DROID-SLAM\cite{DROID-SLAM} with NeRF, achieving accurate pose estimation and efficient reconstruction. Point-SLAM\cite{Point-SLAM} represents scenes as neural point clouds, enhancing the ability to capture details near scene surfaces. Considering that rendering implicit neural radiance fields requires sampling along rays, the slow rendering speed limits its applicability in practical scenarios.

\paragraph{3DGS based dense SLAM}
3D Gaussian primitives offer a concise and flexible ellipsoid representation compared to Point-NeRF\cite{Point-NeRF}. For camera pose estimation, some 3DGS-based approaches \cite{MonoGS, GS-SLAM, g2mapping} implement a renderer that is differentiable with respect to camera pose. SplaTAM\cite{SplaTAM} and Gaussian-SLAM\cite{Gaussian-SLAM} consider that the gradient of the camera pose can be approximated as the gradient of the Gaussian pose relative to the camera pose. Other dense SLAM systems \cite{Photo-SLAM, RTG-SLAM, RP-SLAM, HI-SLAM2} incorporate traditional feature-based visual odometry to balance real-time performance with precision. DROID-SLAM\cite{DROID-SLAM} and Splat-SLAM\cite{Splat-SLAM} integrate end-to-end systems for global camera pose optimization. GS-ICP\cite{GS-ICP-SLAM} utilizes point-to-Gaussian ICP registration, achieving ultrafast and accurate localization. 

However, the geometry consistency of Gaussian representation has emerged as a critical research focus in the field of 3D scene reconstruction. Some researches such as 2DGS\cite{2DGS}, GOF\cite{GOF}, and RaDe-GS\cite{RaDe-GS} attempt to flatten the Gaussian ellipsoid and employ equivalent planes along with unbiased depth estimates to improve the multi-view geometry consistency for Gaussian representation. Unlike previous works that focus on reconstruction quality of Gaussian model, we investigate the impact of geometry consistency on the camera tracking. and propose a novel framework to integrate 2D Gaussian primitives into dense SLAM systems. And we explore incorporating 2D Gaussians and research on geometry consistency from the field of 3D reconstruction into Gaussian-based dense SLAM. 

\section{Method}

\subsection{Gaussian Surfel-based Representation}
Compared to 3D Gaussians, 2D Gaussian surfels offer superior surface modeling capabilities and enhanced geometry accuracy. Theoretically, 2D Gaussian primitives are attached to the tangent planes of the scene surface, which can be represented by the central point $\mu$ of the 2D Gaussian and two tangential vectors ($\mathbf{e_u}$ and $\mathbf{e_v}$) of the space. A point $p=(x,y,z,1)^T$ in the plane can be described as:

\begin{equation}
p = \left[ \begin{array}{cccc}                                                    
     \mathbf{e}_u & \mathbf{e}_v & 0 & \mathbf{\mu}\\
     0 & 0 & 0 & 1\\
\end{array} \right] \left[ \begin{array}{c}
     u\\
     v\\
     1\\
     1
\end{array} \right] = \left[ \begin{array}{cc}
     \mathbf{\Sigma} & \mathbf{\mu}\\
     0 & 1\\
\end{array} \right]p'
\label{eq:important}
\end{equation}

where $p'=(u,v,1,1)^T$ is the homogeneous coordinate of $P$ in the space. And $\mathbf{\Sigma} \in \mathbb{R}^{3 \times 3}$ is represented as the geometry of the 2D plane, which can be decomposed into rotation transformation $\mathbf{R}$ and scaling transformation $\mathbf{S}$ as described by the following formulation. 
\begin{equation}
\mathbf{\Sigma} = \mathbf{RS}
\label{eq:geometry}
\end{equation}
The scaling transformation $S$ has a scaling factor of $0$ along the third dimension. Then, the value of the 2D Gaussian at point $p$ can be evaluated by the following formulation:
\begin{equation}
\mathcal{G}(p') = \exp\left( \frac{u^2+v^2}{2} \right)
\label{eq: Gaussian}    
\end{equation}
Following the representation method of 3DGS\cite{3DGS}, a scene can be represented as a set of 2D Gaussian primitives with geometry $\Sigma_i$, central points $\mu_i$, opacity $o_i$, and color $c_i$:
\begin{equation}
G = \{\mathcal{G}_i:(\Sigma_i, \mu_i, o_i, c_i) | i=1, ..., n\}
\label{eq:primitives}
\end{equation}
Given a ray sampled from the observation perspective and a set of 2D Gaussian primitives, the intersection $p'_i$ of the ray with $i$-th primitive can be solved. The color corresponding to the ray can be obtained by accumulating the Gaussian values at intersection points sequentially using the $\alpha$-blending technique, which can be formulated as:
\begin{align}
&\alpha_i = o_i \mathcal{G}_i(p'_i) 
& & w_i=\alpha_i \prod_{j=1}^{i-1}(1-\alpha_i) \label{eq: weight} \\
&I(\mathbf{r}) = \sum_{i=1}c_iw_i 
& & A(\mathbf{r}) = \sum_{i=1}w_i \label{eq: colormap} 
\end{align}
where $\mathbf{r}$ is the sampled ray corresponding to a pixel, $w_i$ is denoted the blending weight of the i-th Gaussian. We also render the accumulated opacity map $A$, which represents the sum of the blending weights along each ray. When the accumulated opacity approaches $1$, it indicates that the scene has been well-optimized.
\subsection{Surface-aware Depth Rendering}
\paragraph{Unbiased depth.}
In 2DGS\cite{2DGS}, depth rendering no longer relies on EWA splatting\cite{EWA-Splatting}, a method that approximates Gaussian primitives as ellipses on the projection plane. Instead, it directly computes the intersection depth (unbiased depth) between the ray and the 2D Gaussian primitives. Compared to the 3D Gaussian-based SLAM, our innovative use of 2D Gaussian primitives achieves an unbiased depth that more accurately represents the geometry of the Gaussian primitive.


\paragraph{Depth adjustment.}
In $\alpha$-blending process, when a ray passes through multiple surfaces, surfaces located behind, although having a small weight, may still interfere with the depth estimation of the foreground surfaces due to their significant depth differences. This interference is challenging to eliminate with a simple threshold. To address this, we propose a depth adjustment approach, which assigns a weight $\beta_i$ to $i$-th Gaussian in the $\alpha$-blending process. This weight is used to adjust the contribution of $i$-th Gaussian's unbiased depth $d_i$ to the overall depth composition. The surface-aware weighted depth $d_i'$ of $i$-th Gaussian is computed by the following equation:
\begin{equation}
d_{i}' = \beta_id_i+(1-\beta_i)d_{m}
\label{eq:fix_depth}
\end{equation}
The median depth $d_m$ is defined as the depth corresponding to the $m$-th Gaussian when the accumulated opacity first exceeds 0.5($\sum_{i=0}^mw_i > 0.5$) during ray traversal through the sequence of Gaussians. Specifically, if the ray has not reached the $m$-th Gaussian, we set $\beta_i=1$, as the depth of $i$-th Gaussian does not negatively impact the blending depth. Otherwise, $\beta_i$ is computed based on the distance between $d_i$ and $d_m$, as well as the variance of distances, as described by the following formula.
\begin{equation}
\sigma_i=\sqrt{\sum_{j=0}^{i-1}w_i(d_j'-d_{m})^2}
\label{eq:sa_sigma}
\end{equation}

\begin{equation}
\beta_i = Exp(-\frac{(d_i-d_{m})^2}{B\sigma_i^2}), i>m
\label{eq:sa_conf}
\end{equation}
where $B$ serves as a hyperparameter that controls the sensitivity of the weight with respect to both distance and variance. As shown in \cref{fig:overview}, when the distance between the Gaussian and the median depth increases, the weight $\beta_i$ decreases, leading to a reduced influence of the $i$-th Gaussian on the depth rendering. 

\paragraph{Depth Normalization.}
During the $\alpha$-blending process, small differences in the accumulated weights along the rays from different views can lead to significant underestimation of the rendered depth. To address this, we normalize the weights of all Gaussian depths during depth map rendering, which can be formulated as:
\begin{equation}
D(\mathbf{r}) = \frac{\sum_{i=1}^nd_i'w_i}{A(\mathbf{r})}
\label{eq:depth_rend}
\end{equation}
\subsection{Camera Tracking}
GauS-SLAM employs a frame-to-model tracking approach. Specifically, given a set of 2D Gaussian primitives $
G^w=\{\mathcal{G}_i:(\Sigma_i, \mu_i, o_i, c_i) | i=1, ..., n\}
$ in the scene and the initial pose $
\{\mathbf{R},\mathbf{t}\}$ of the tracking frame, the camera pose is iteratively optimized by refining $\mathbf{R}$ and $\mathbf{t}$ to minimize the discrepancy between the rendered and real images. Similar to Gaussian-SLAM \cite{Gaussian-SLAM}, we treat the optimization of the camera pose as an equivalent optimization of the relative pose of the Gaussian primitives within the scene under a fixed camera viewpoint. Specifically, the set of Gaussian primitives is transformed into the camera coordinate system, which can be formalized as: 
\begin{equation}
G^{c}= \{\mathcal{G}_i:(\mathbf{R}\Sigma_i, \mathbf{R}\mu_i+\mathbf{t}, o_i, c_i)| i=1, ..., n\}
\label{eq: gaussians}
\end{equation}
The following loss function will be used to jointly optimize the $\{\mathbf{R},\mathbf{t}\}$, such that the rendered depth and color maps are aligned with the ground truth.
\begin{equation}
    \mathcal{L}_{track}=(A>0.9)\left(
    \mathcal{L}_1(D,\hat{D}) +
    \lambda_1 \mathcal{L}_1(I,\hat{I})\right)
     \label{eq:track_loss}
\end{equation}
To mitigate the influence of new observed scene and outliers, we apply the loss function to pixels with accumulated opacity exceeding 0.9.

\subsection{Incremental Mapping}

In mapping process, the densification strategy based on cloning and splitting in 3DGS\cite{3DGS} typically requires substantial iteration counts and extensive multi-view  constraints to achieve satisfactory scene coverage. To accelerate reconstruction efficiency, we propose a Surfel Attachment initialization approach inspired by \cite{SplaTAM}. Specifically, the 2D Gaussian surfels will be directly positioned at the unprojection of pixels where the accumulated opacity is less than 0.6. And the initial scale $\mathbf{S}$ and orientation $R$ are determined by the following equations:
\begin{equation}
\mathbf{S}=diag(\frac{d_{gt}}{f}, \frac{d_{gt}}{f}, 0) 
\end{equation}
\begin{equation}
\mathbf{R}=(\mathbf{e_1},\mathbf{e_2}, \mathbf{n_{gt}}) \quad \mathbf{e_1} \perp \mathbf{e_2} \perp \mathbf{n_{gt}}
\end{equation}
\begin{equation}
\mathbf{n}_{gt}=\frac{\nabla_x \mathbf{D}_{gt}\times \nabla_y \mathbf{D}_{gt}}{|\nabla_x \mathbf{D}_{gt}\times \nabla_y \mathbf{D}_{gt}|}
\end{equation}
where $\mathbf{D}_{gt}$, $\mathbf{n}_{gt}$ is represented as the ground-truth depth and normal of the pixel. 
In regions lacking ground-truth depth, Gaussians are placed in areas where the accumulated opacity falls between 0.4 and 0.6. Since these regions have been partially reconstructed, the rendered depth can be used as the ground-truth depth to initialization Gaussians. This process facilitates the expansion of the Gaussian model along object boundaries, which we refer to as Edge Growth.

During the mapping process, the front-end randomly selects a frame from the local map and utilizes the following loss function to optimize the Gaussian model for local reconstruction.
\begin{equation}
    \mathcal{L}_{map}= \mathcal{L}_1(D,\hat{D}) +
    \lambda_1 \mathcal{L}_1(I,\hat{I}) + 
    \lambda_2\mathcal{L}_{reg}
    \label{eq: map_loss}
\end{equation}
The term $\mathcal{L}_{reg}$ reduces the depth uncertainty along the rays by minimizing the weighted MSE between all fixed depth of each Gaussian on the ray and the median depth $d_{m}$.
\begin{equation}
  \mathcal{L}_{reg}=\sum_{r} \sum_{i=1}w_i(d_i'-d_{m})^2
  \label{eq: reg_loss}
\end{equation}
\subsection{GauS-SLAM System}
\paragraph{Front-end.}
In the front-end, all optimization processes are performed within a local map. The first frame of the local map serves as the reference keyframe (RKF). When processing a new frame, the front-end first performs camera tracking to estimate its pose relative to the RKF $\mathbf{T}_{RKF}^{F}\{\mathbf{R}, \mathbf{t}\}$ . It then evaluates whether the frame qualifies as a keyframe (KF) based on the proportion of newly observed scene exceeding a predefined threshold $\tau_{k}$. The incremental mapping is performed on KFs. If the number of Gaussian primitives in the local map exceeds a specified threshold $\tau_l$, the front-end send the frames and local Gaussian map to the back-end and a new local map is reinitialized to continue tracking and mapping. At this point, the current frame is marked as a new RKF within the new local map. 

\paragraph{Back-end.}

The back-end of the system is primarily responsible for merging local maps and optimizing the global map. Upon receiving a local map, the backend stores the frames from the local map as sub-maps in the database and integrates the local Gaussian map into the global map. Specifically, the Gaussian primitives in the local map are first reset to 0.01 in terms of opacity and then added to the global map according to their RKF poses. Subsequently, the current submap and its co-visible submaps are jointly selected for local mapping. To determine co-visibility between submaps, we utilize the visual features extracted from the first and last frames of each submap using NetVLAD\cite{NetVLAD}. After the mapping process, Gaussian primitives with opacity below 0.05 are pruned. This step effectively removes overlapping parts between the local and global maps, thereby preventing the continuous accumulation of Gaussian primitives.

To reduce the accumulation of trajectory errors, Bundle Adjustment(BA) will be applied to optimize the poses of the RKFs $\mathbf{T}_w^{RKF}\{\mathbf{R},\mathbf{t}\}$ involved in the co-visible submaps, as well as the global map. During the BA process, frames are randomly selected from co-visible submaps and the optimization is performed through the minimization of the following formula.
\begin{equation}
    \mathcal{L}_{ba}= \mathcal{L}_1(D,\hat{D}) 
 + \lambda_1 \mathcal{L}_1(I,\hat{I}) 
    \label{eq:ba}
\end{equation}

When the backend is not busy, a frame is stochastically selected from the submaps in the database to refine the global map, which we refer to as the Random Optimization. This process effectively mitigates the forgetting issue and enhances the global consistency of the Gaussian scene. After the reconstruction is completed by the front-end and back-end, Random Optimization continues to run for an additional period to reduce floating Gaussians, ensuring that the global map is evenly optimized. We refer to this process as Final Refinement, and experiments demonstrate that it significantly improves the rendering quality.

\section{Experiments}
\definecolor{top1}{HTML}{BFE8C1}
\definecolor{top2}{HTML}{FFF5B3}
\definecolor{top3}{HTML}{FFD9B3}
\definecolor{gray}{HTML}{EEEEEE}
\subsection{Experiment Setup}
\begin{figure*}[ht]
    \centering
    \setlength{\tabcolsep}{1pt}
    \renewcommand{\arraystretch}{0.5}
    \begin{tabular}{cccccc}
        \multirow{2}{*}[4ex]{\rotatebox{90}{\texttt{Office 2}}} & 
        \includegraphics[width=0.19\linewidth]{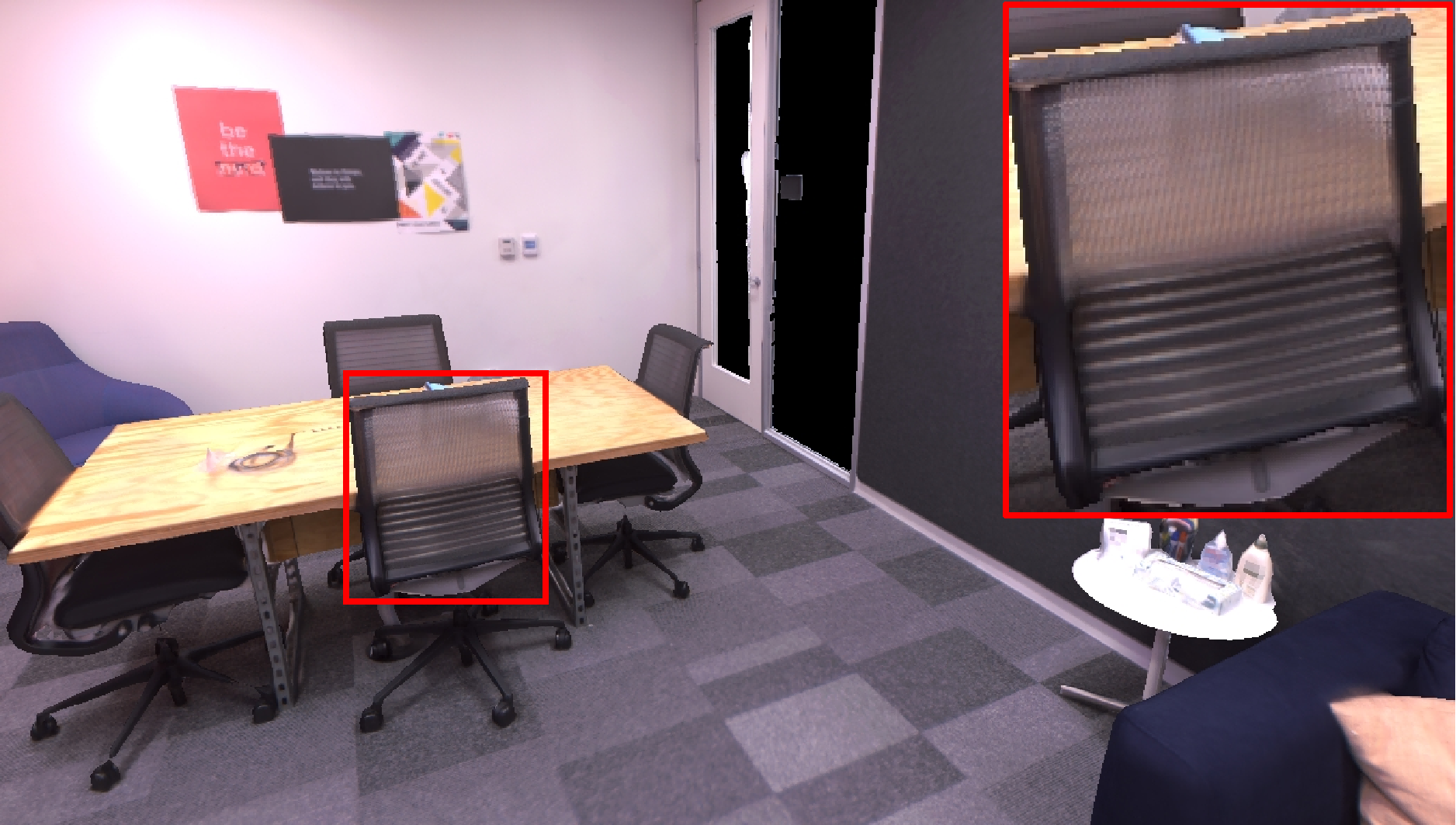} &
        \includegraphics[width=0.19\linewidth]{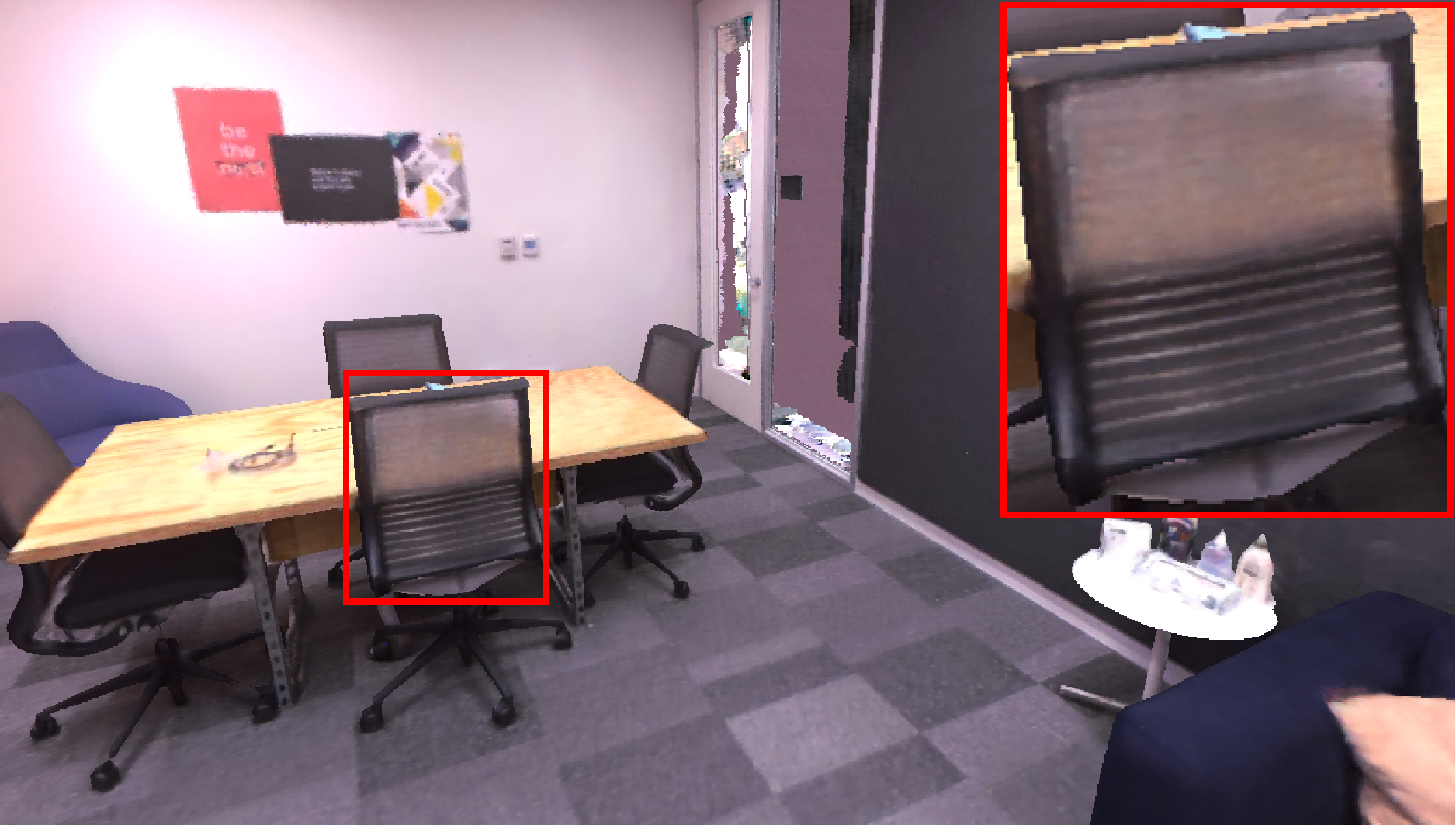} &
        \includegraphics[width=0.19\linewidth]{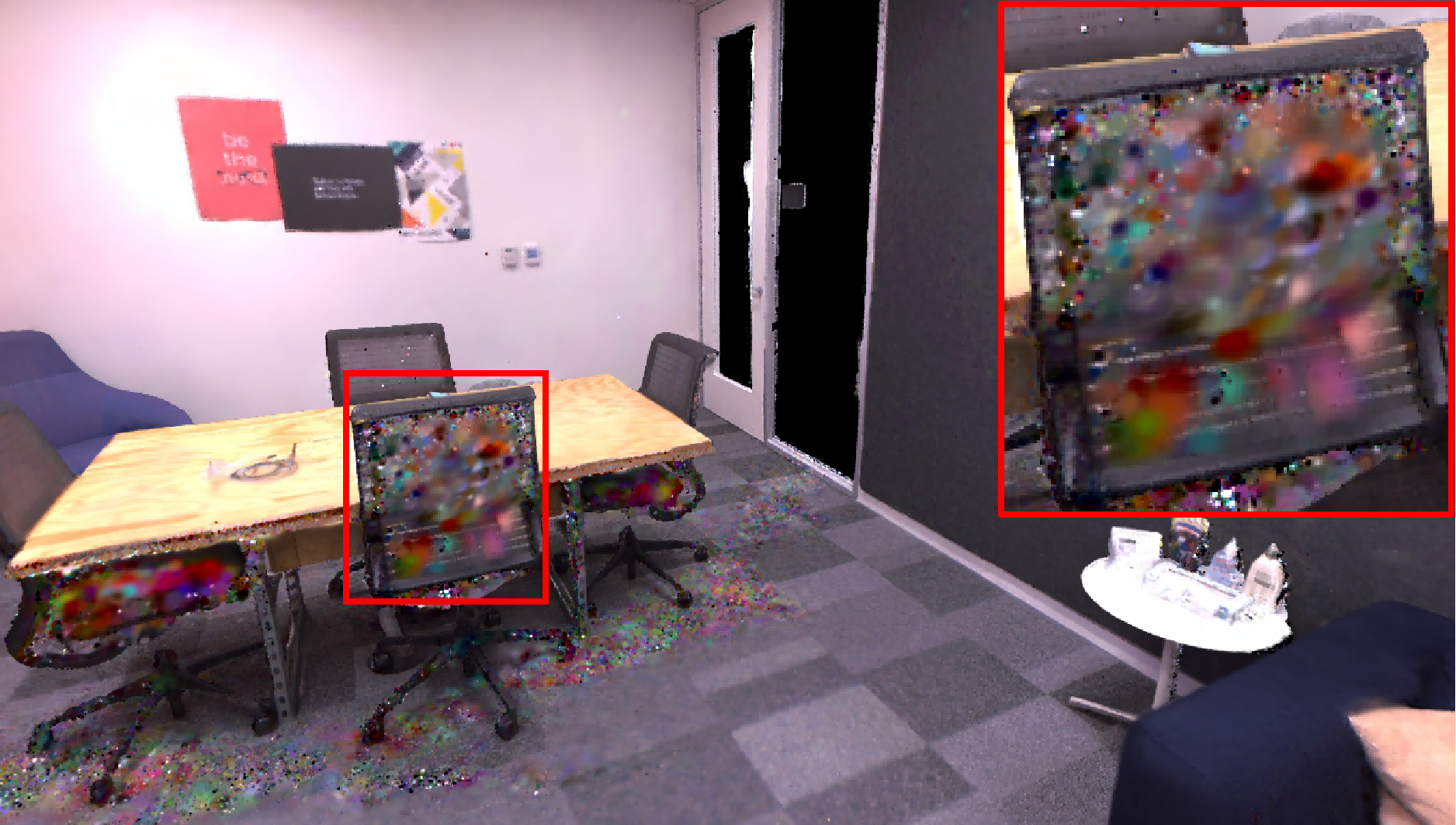} &
        \includegraphics[width=0.19\linewidth]{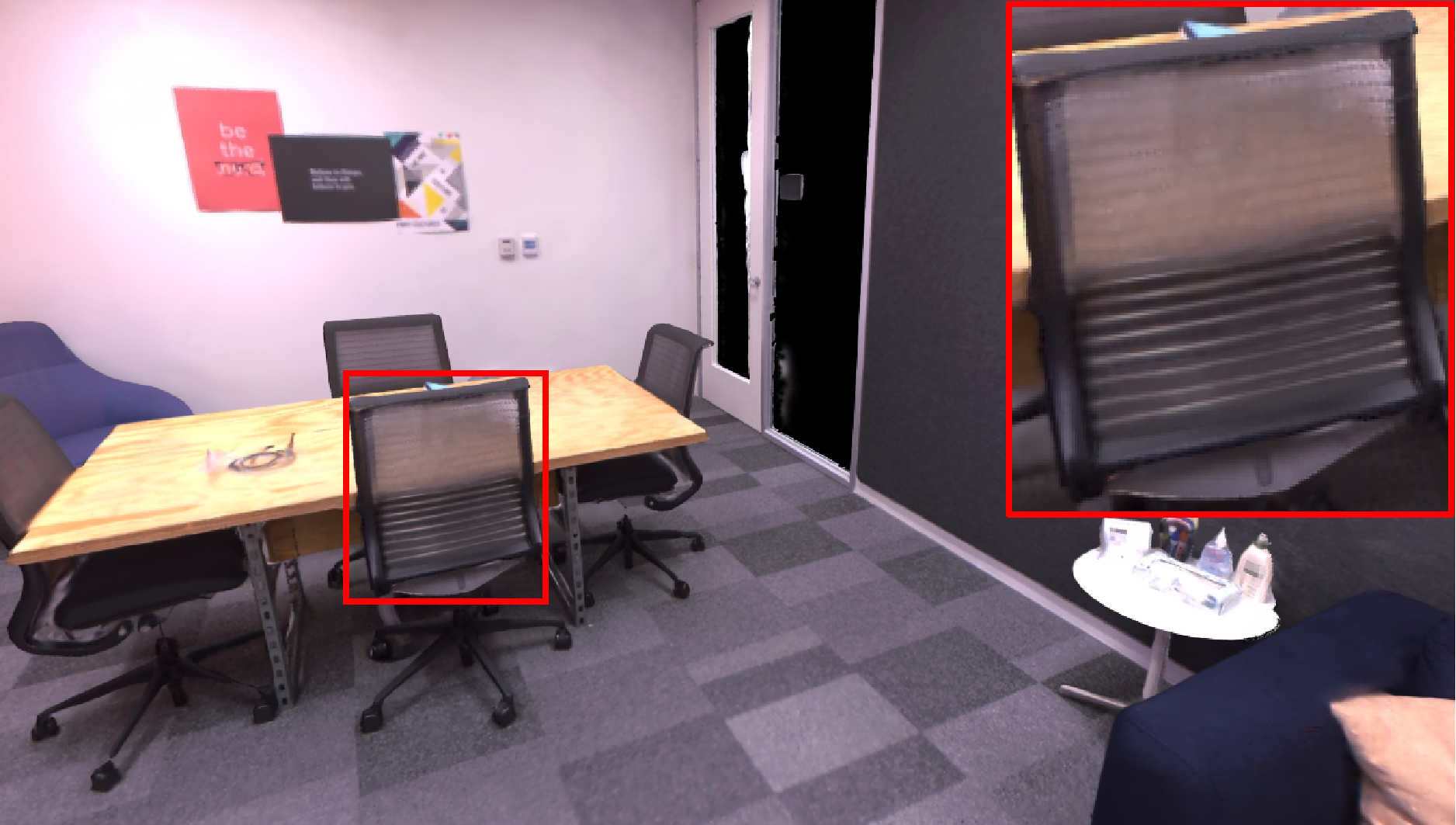} &
        \includegraphics[width=0.19\linewidth]{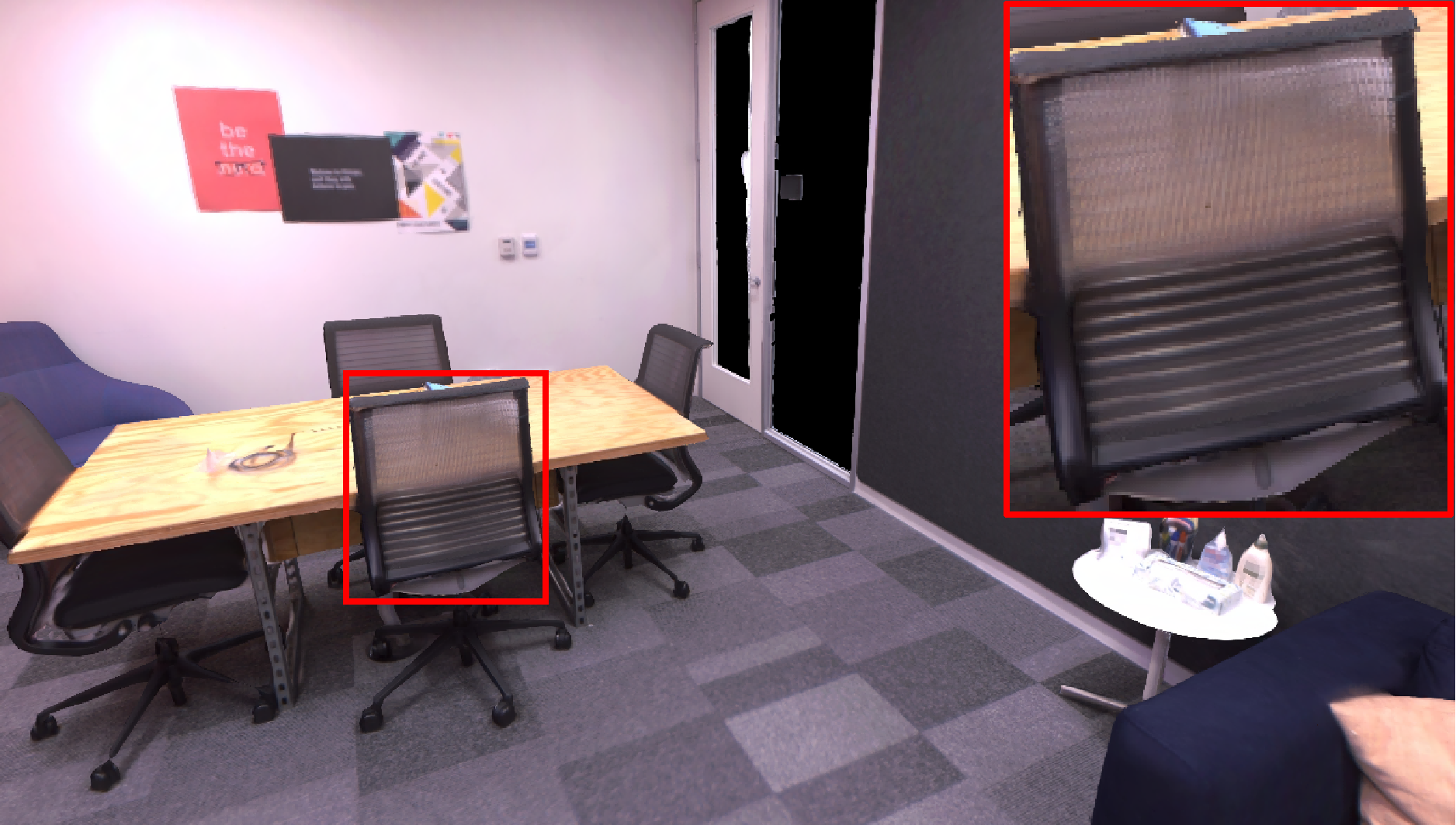} \\
         &
        \includegraphics[width=0.19\linewidth]{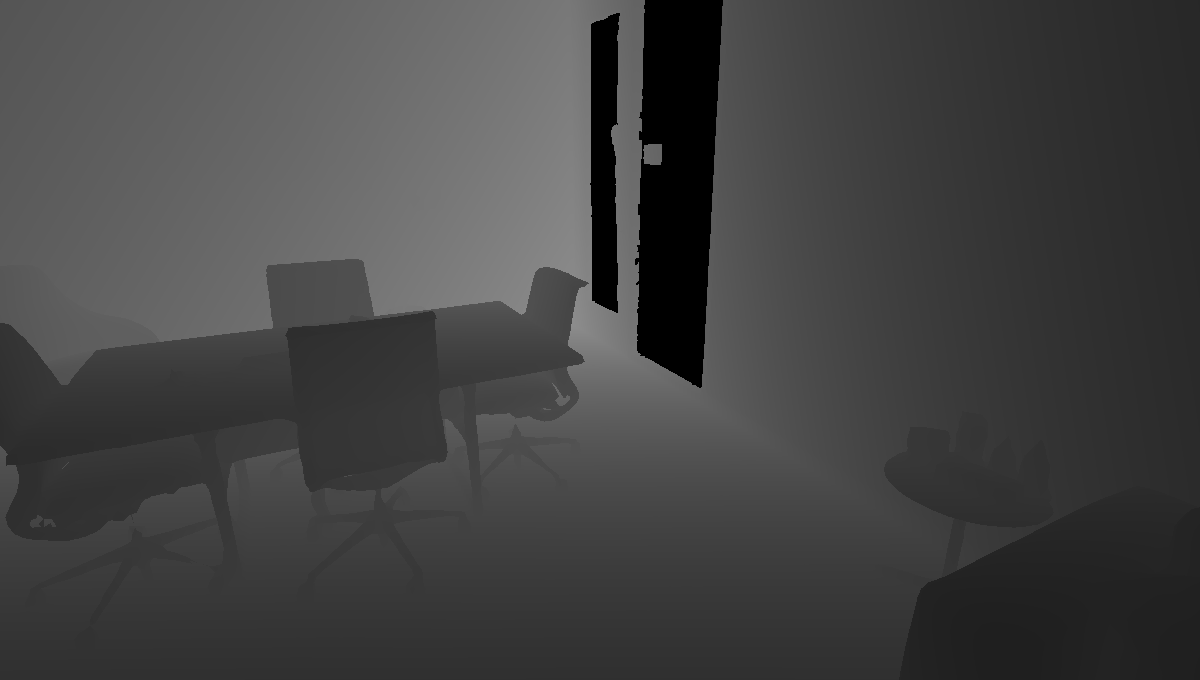} &
        \includegraphics[width=0.19\linewidth]{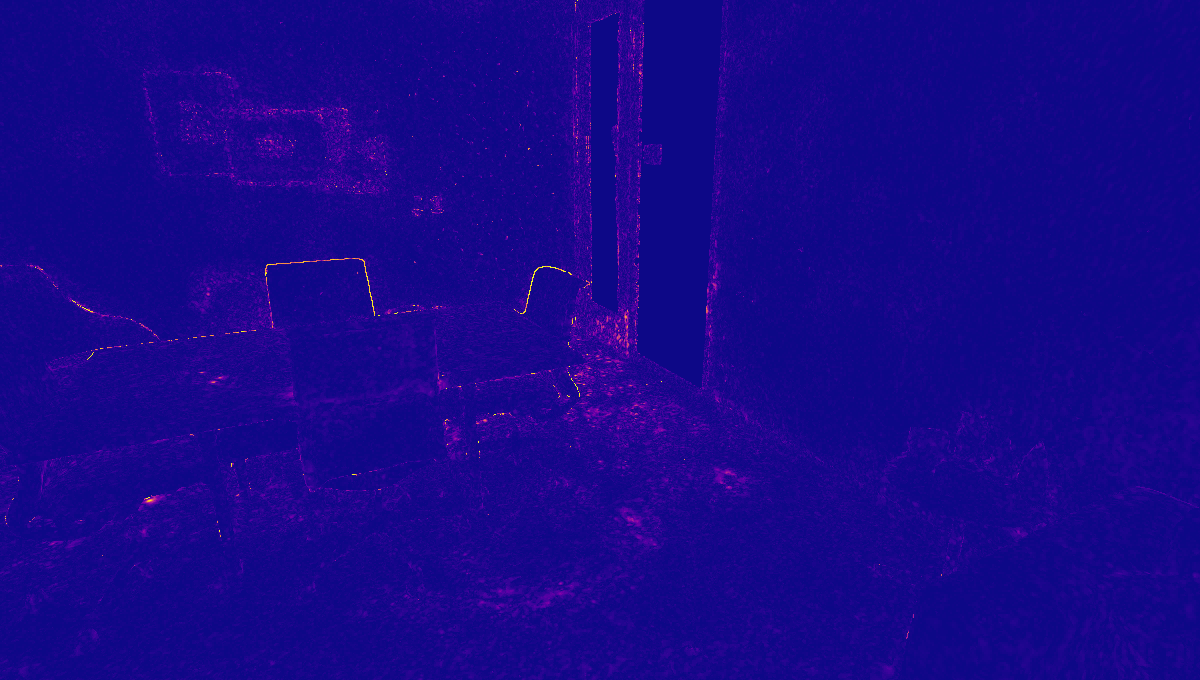} &
        \includegraphics[width=0.19\linewidth]{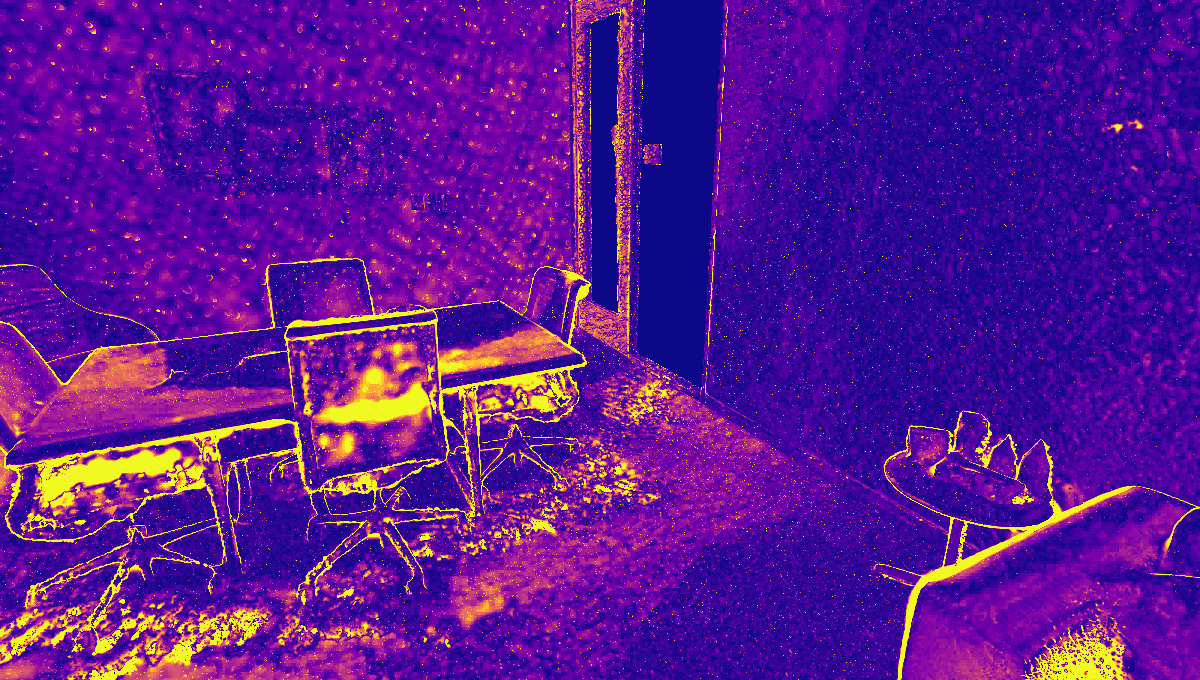} &
        \includegraphics[width=0.19\linewidth]{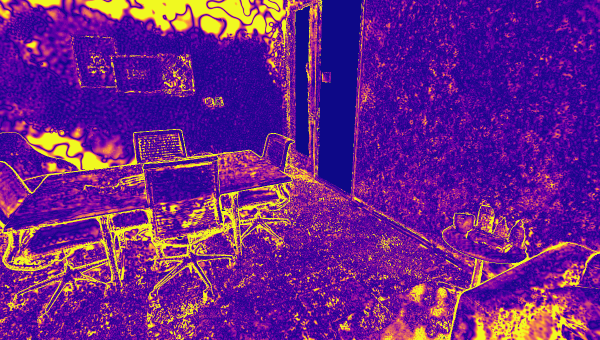} &
        \includegraphics[width=0.19\linewidth]{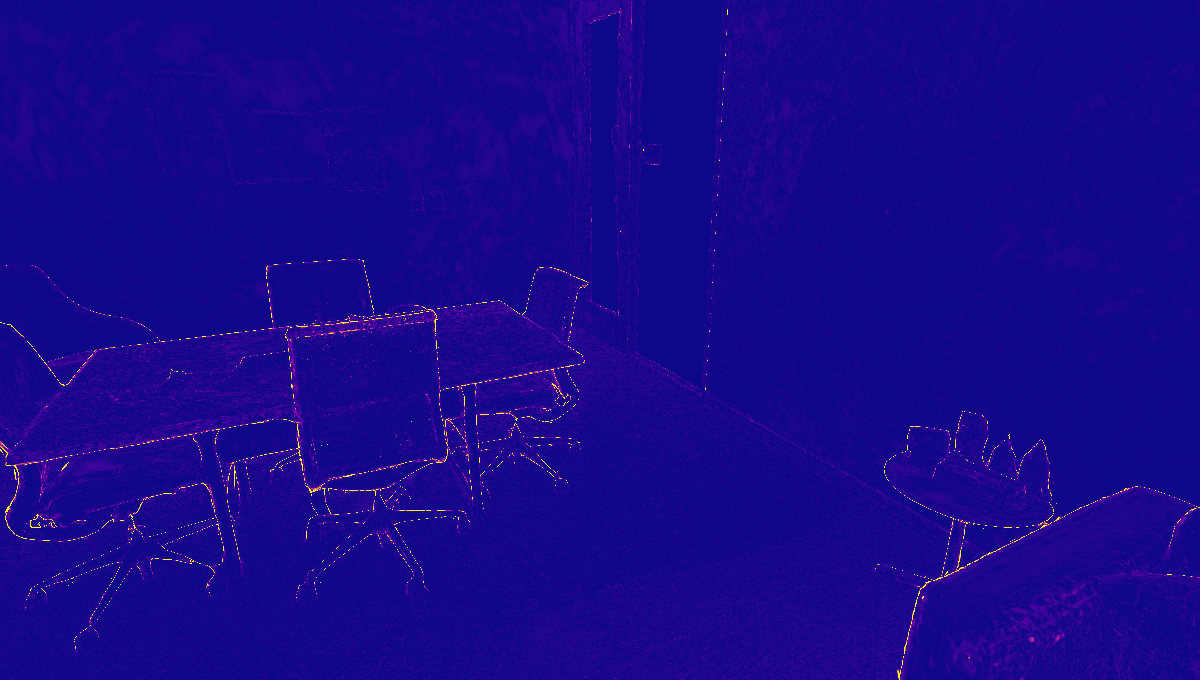} \\
        
        \multirow{2}{*}[4ex]{\rotatebox{90}{\texttt{Office 4}}} & 
        \includegraphics[width=0.19\linewidth]{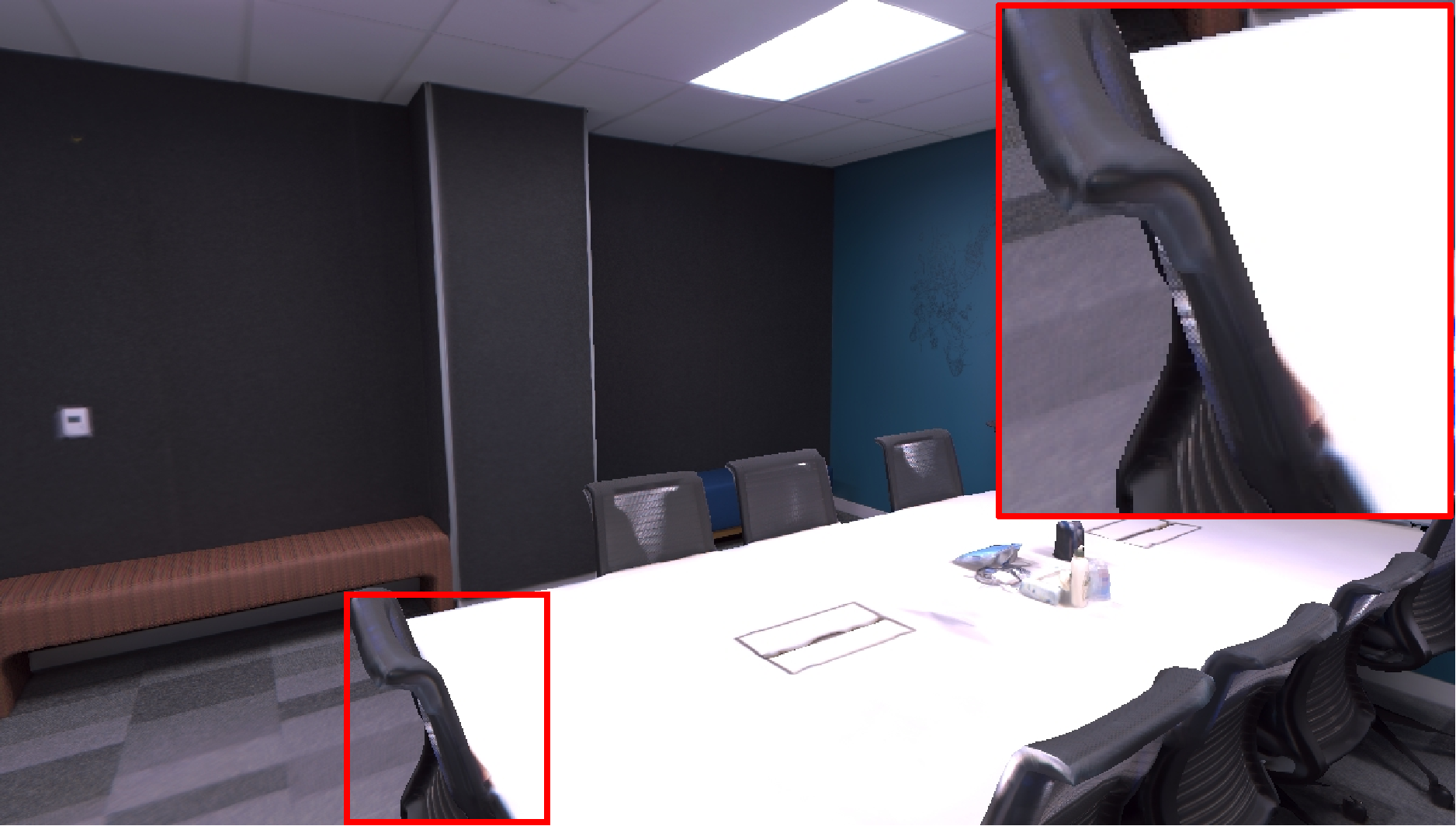} &
        \includegraphics[width=0.19\linewidth]{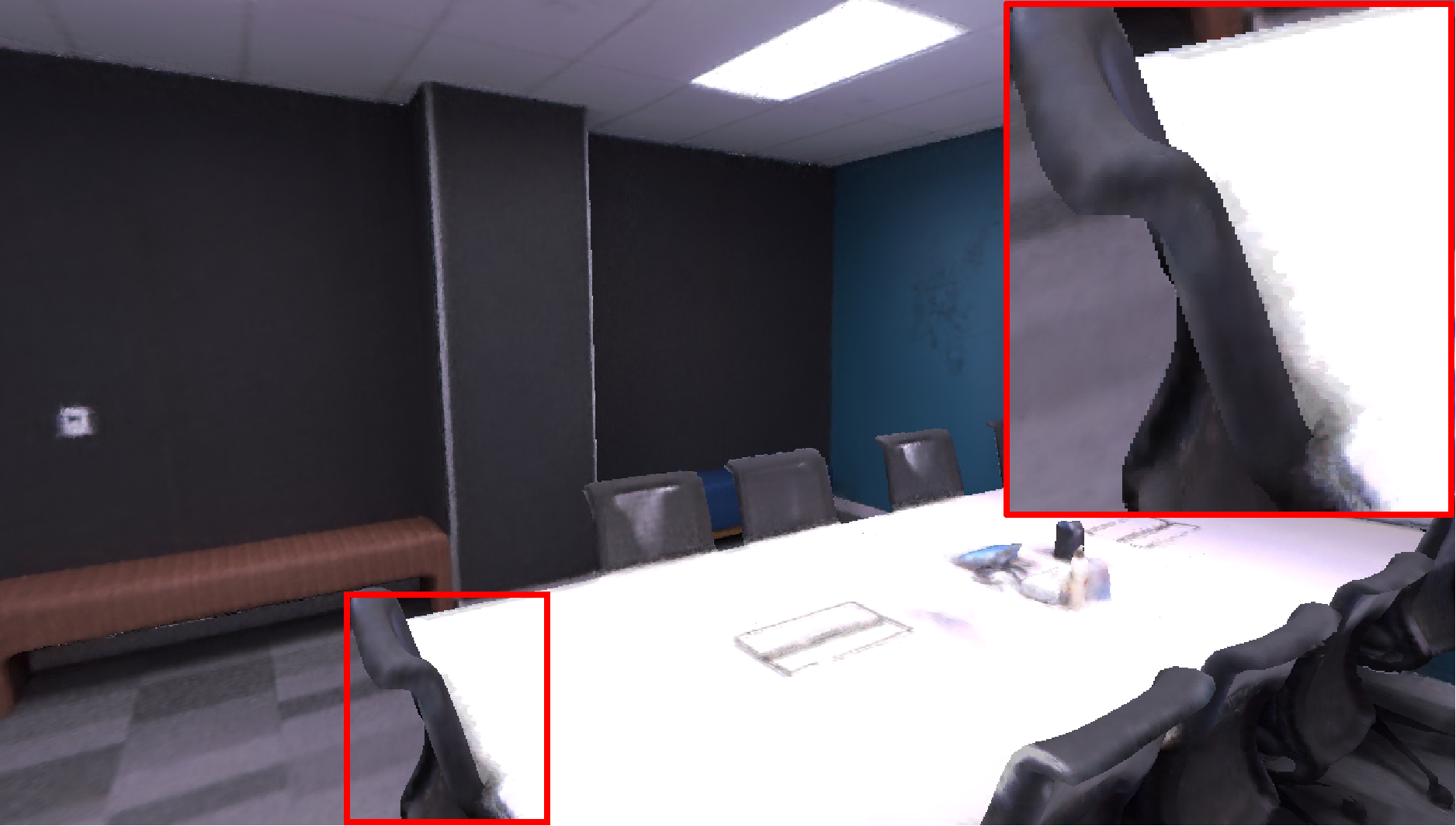} &
        \includegraphics[width=0.19\linewidth]{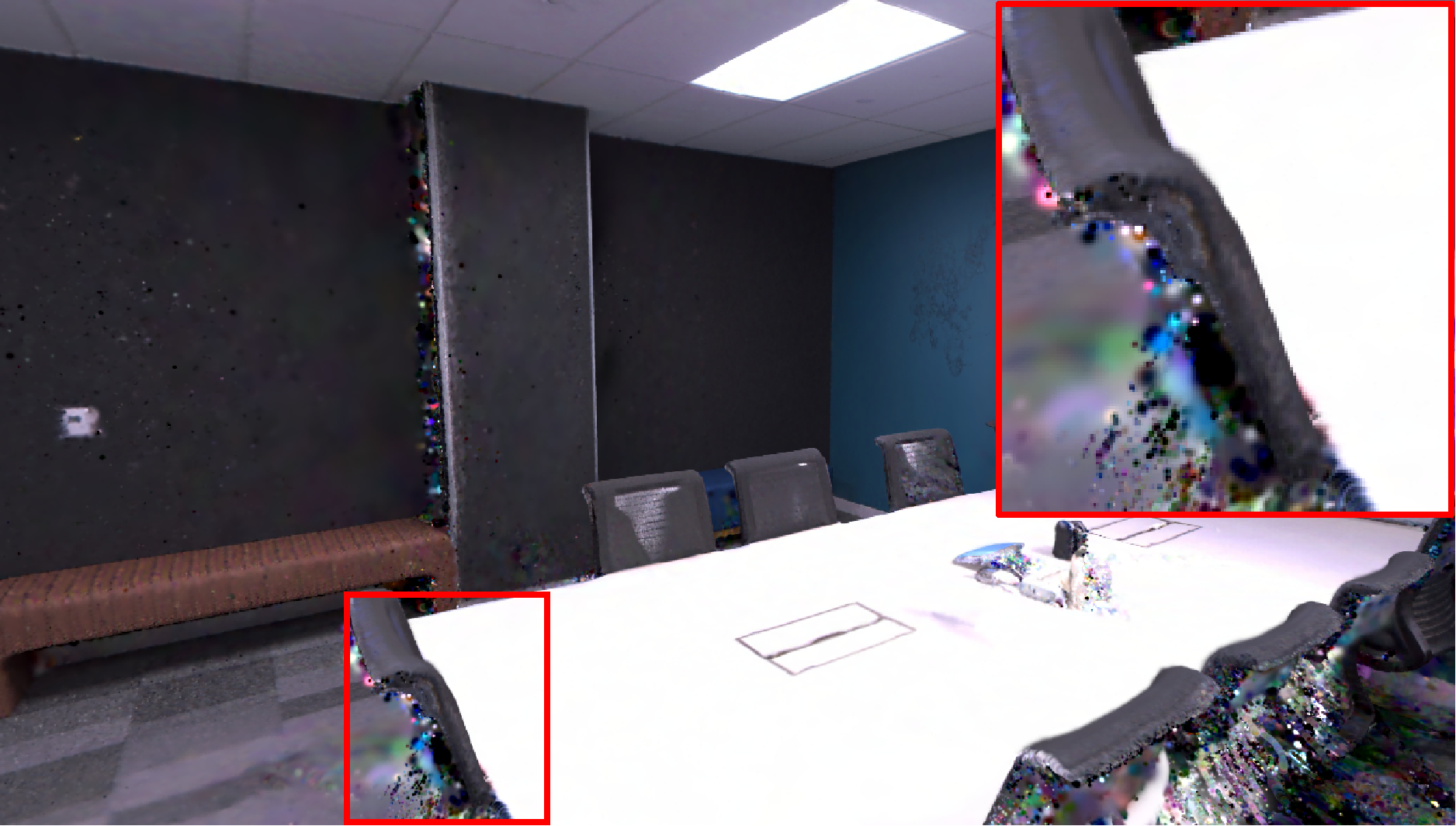} &
        \includegraphics[width=0.19\linewidth]{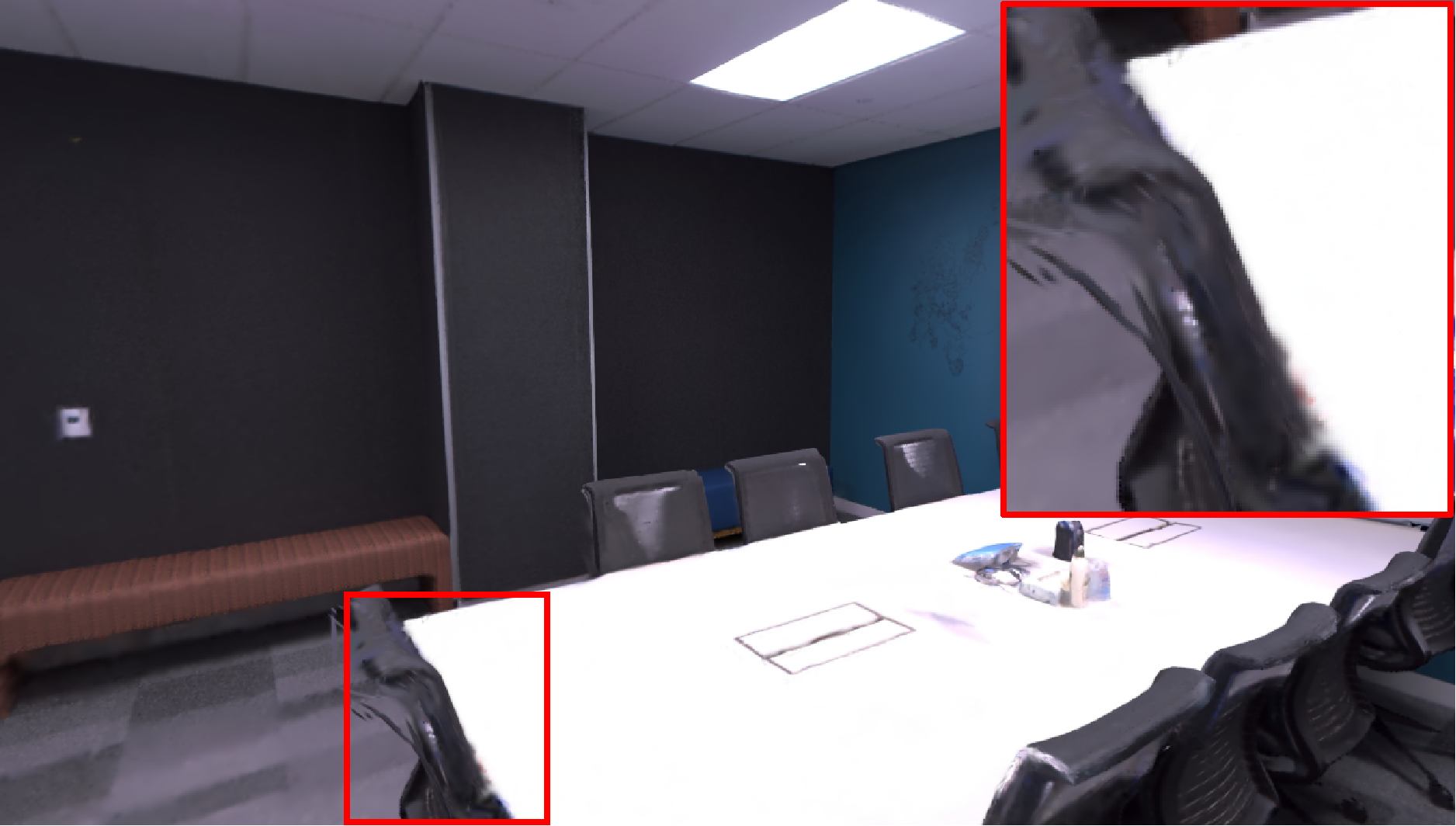} &
        \includegraphics[width=0.19\linewidth]{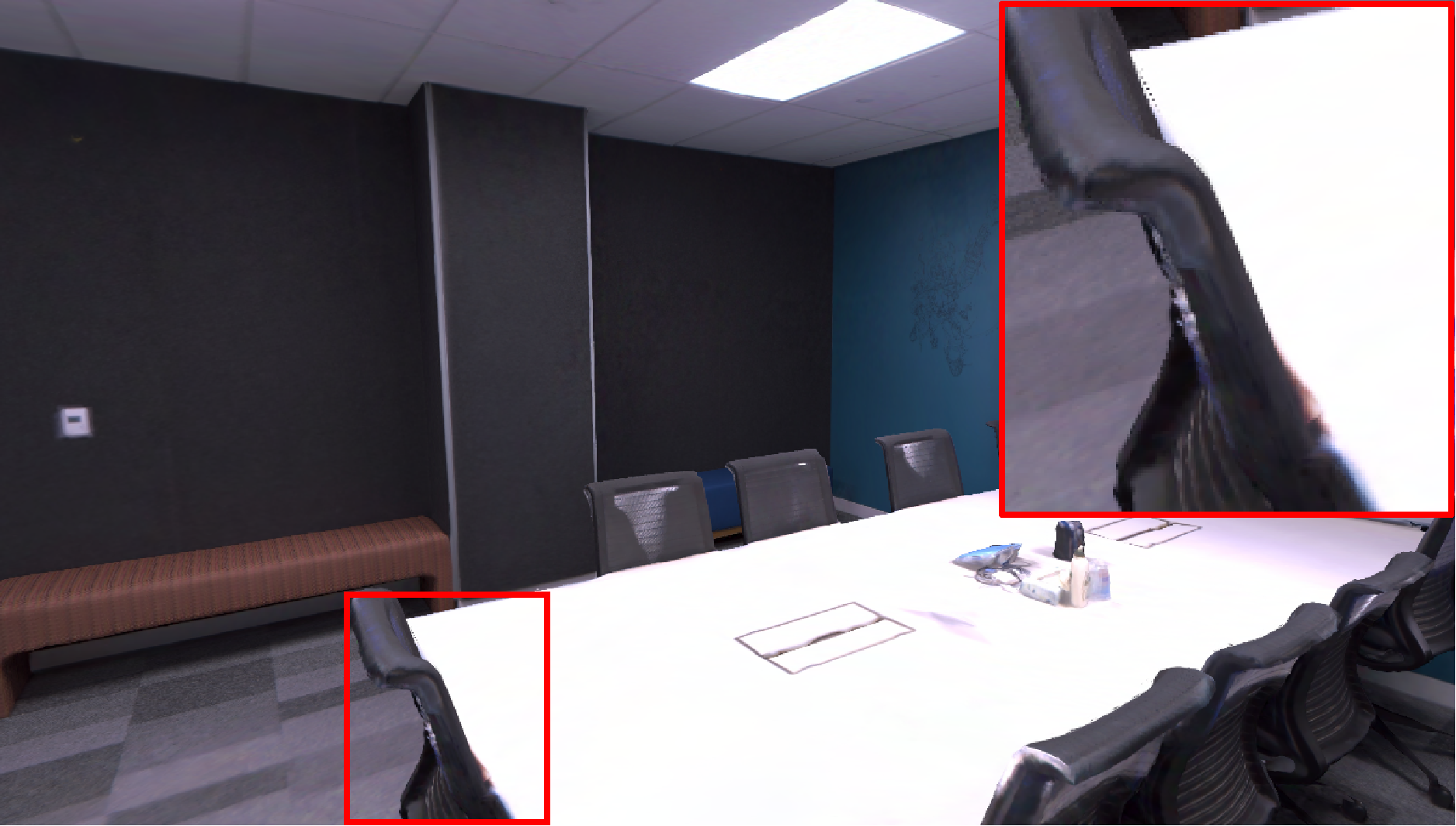} \\
         &
        \includegraphics[width=0.19\linewidth]{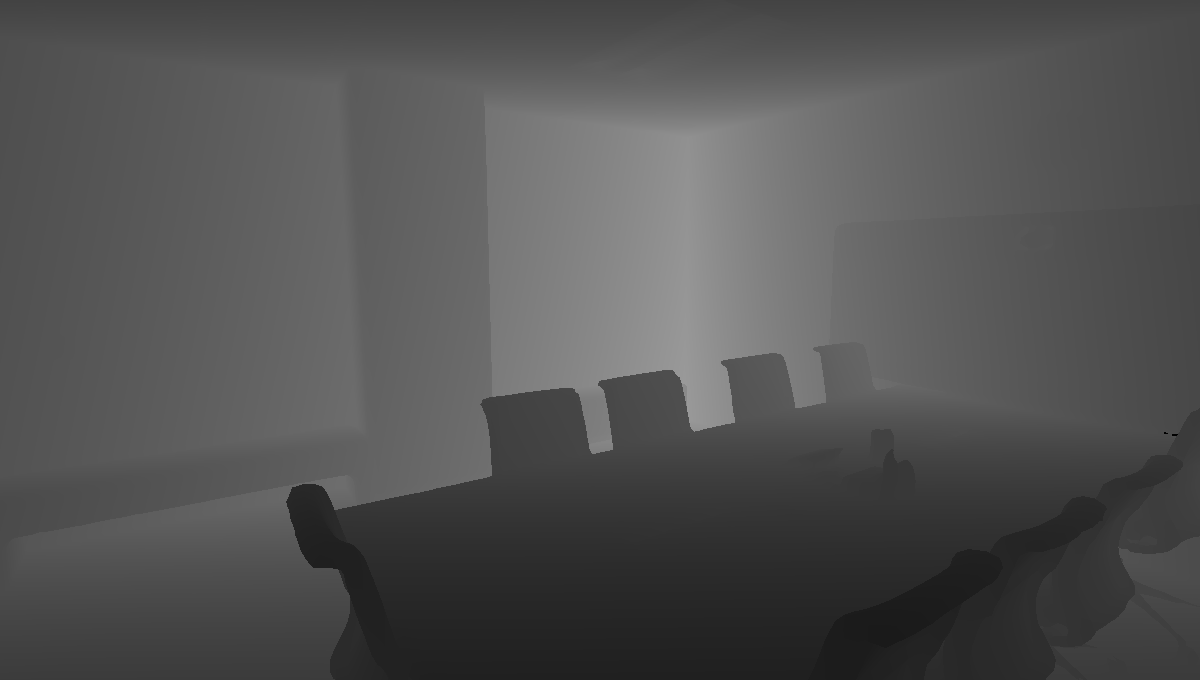} &
        \includegraphics[width=0.19\linewidth]{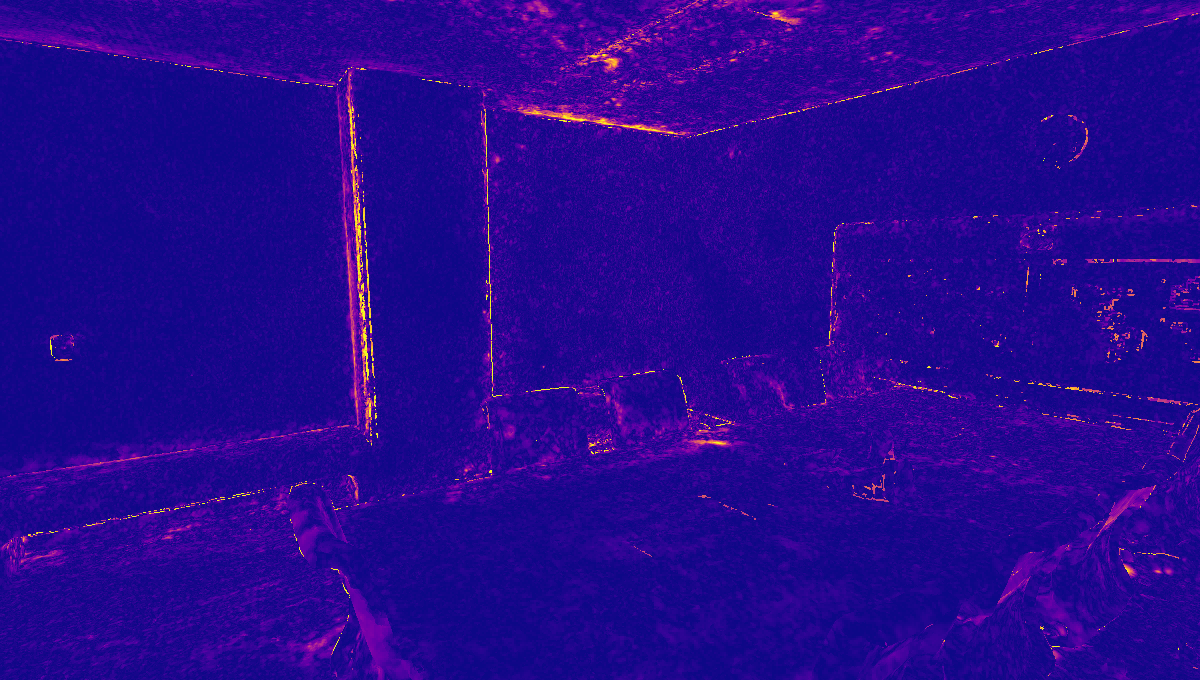} &
        \includegraphics[width=0.19\linewidth]{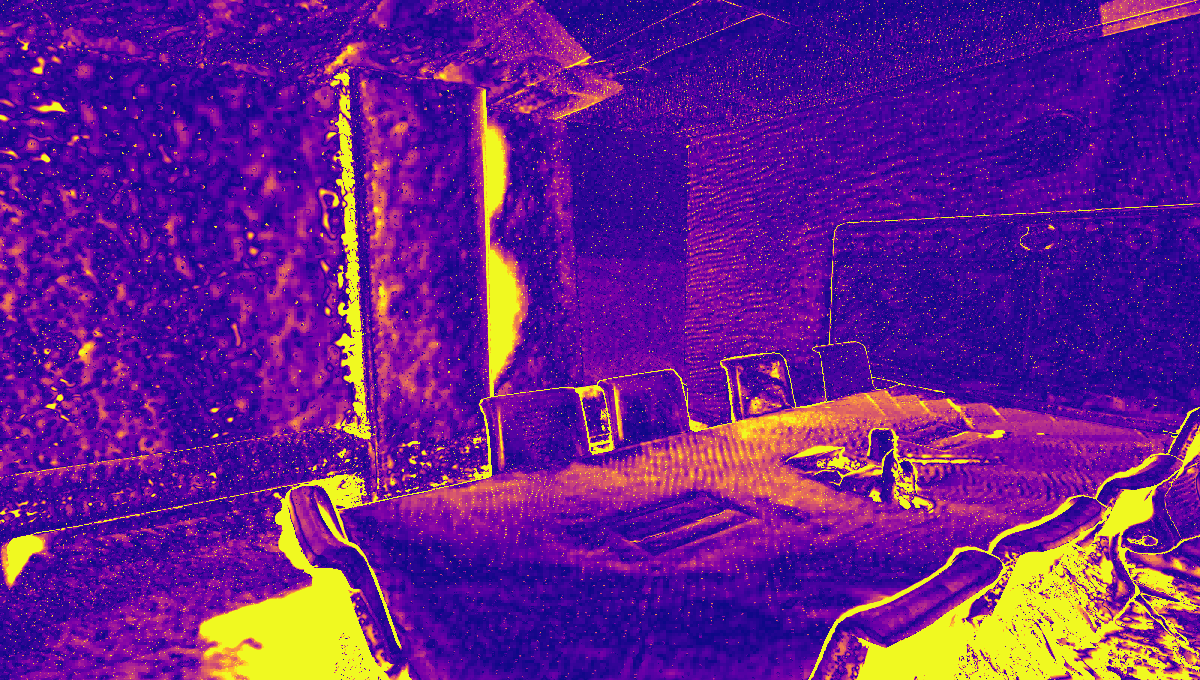} &
        \includegraphics[width=0.19\linewidth]{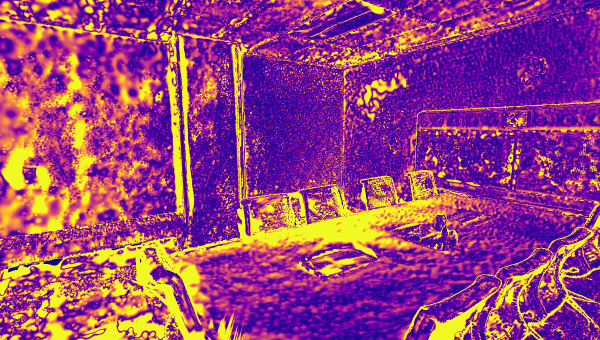} &
        \includegraphics[width=0.19\linewidth]{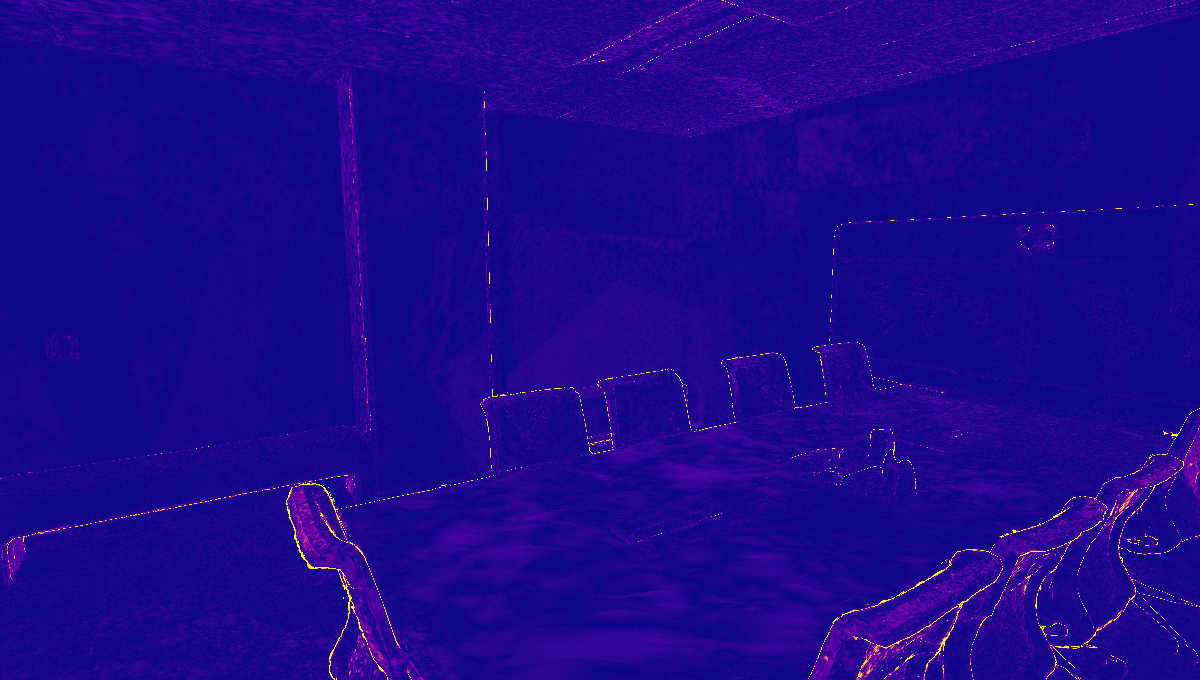} \\

        & Ground-truth & Point-SLAM\cite{Point-SLAM} & SplaTAM\cite{Splat-SLAM} & MonoGS\cite{MonoGS} & \textbf{GauS-SLAM(Ours)} \\
    \end{tabular}
    
    \caption{\textbf{The comparison of Rendering performance on Replica\cite{Replica}.} We present rendered color maps and depth error maps from 2 viewpoints to comparatively evaluate the rendering quality and geometry accuracy of different approaches.}
    \label{fig:replica_rendering_comparison}
\end{figure*}
\paragraph{Implementation Details.} The parameter $B$ in the depth adjustment is set to 4. To accelerate the convergence of pose estimation during tracking, the parameters $\beta_1$ and $\beta_2$ of the Adam optimizer are set to 0.7 and 0.99, respectively. In the frontend, a keyframe is generated when the proportion of newly observed scene exceeds $\tau_k=1\%$. And, a new local map is created when the number of Gaussian primitives in the local map exceeds $\tau_l=1.5\cdot HW$. In the loss function, the parameters $\lambda_1$ and $\lambda_2$ are set to $0.5$ and $0.1$, respectively. To mesh scenes, we employed TSDF Fusion \cite{TSDF-Fusion} with voxel size 1 cm to integrate the color maps and depth maps, which is similar to \cite{Gaussian-SLAM, Point-SLAM}. All experiments in this paper are conducted on the Intel Core i9-14900K processor along with an NVIDIA GeForce A6000 GPU.

\paragraph{Datasets \& Evaluation Metrics.} We evaluate our approach on both synthetic and real-world datasets, including the Replica\cite{Replica} synthetic dataset, and the TUM-RGBD\cite{TUM-RGBD}, ScanNet\cite{ScanNet}, and ScanNet++\cite{scannetpp} real-world datasets. Two challenging sequences \texttt{S1}(\texttt{b20a261fdf}), \texttt{S2}(\texttt{8b5caf3398}) are sampled from ScanNet++\cite{scannetpp} for evaluation. EVO\cite{evo} toolkit is employed to calculate the ATE-RMSE\cite{ATE-RMSE} metric to measure the trajectory accuracy. PSNR, SSIM\cite{SSIM} and LPIPS\cite{LPIPS} are utilized for evaluating rendering quality. Meanwhile, we also provide the Depth L1 as in \cite{Nice-SLAM} and the F1-Score to evaluate the geometry quality of reconstruction results. 

\subsection{Comparison with SOTA Baselines}
\paragraph{Tracking Performance.}
The tracking performance comparison of selected sequences on the four datasets is presented in \cref{table: replica} and \cref{table:tracking_performance}. Our proposed GauS-SLAM achieves millimeter-level localization accuracy, establishing new SOTA performance on both Replica\cite{Replica} and ScanNet++\cite{scannetpp} datasets. Specifically, on the Replica\cite{Replica} dataset, our method demonstrates superior performance with an ATE-RMSE\cite{ATE-RMSE} of $0.06$ cm, representing a $62.5\%$ improvement over the previous SOTA method GS-ICP\cite{GS-ICP-SLAM} and an $83\%$ enhancement compared to our baseline SplaTAM\cite{SplaTAM}. Despite the presence of challenging factors such as exposure variations and motion blur in the TUM-RGBD\cite{TUM-RGBD} and ScanNet\cite{ScanNet} datasets, GauS-SLAM maintains competitive performance. Notably, it even outperforms SLAM methods with loop closure correction on some sequences in ScanNet\cite{scannetpp}.

\begin{table}[ht]
\centering

\resizebox{0.5\textwidth}{!}{
\begin{tabular}{lcccccc}
\toprule
\multirow{2}{*}{\textbf{Method}} & \textbf{PSNR} & \textbf{SSIM} & \textbf{LPIPS} & \textbf{ATE} & \textbf{Depth L1} & \textbf{F1-Score}\\ 
& [db]$\uparrow$ & $\uparrow$& $\downarrow$& [cm]$\downarrow$ & [cm]$\downarrow$ & [\%]$\uparrow$ \\
\midrule
ESLAM\cite{ESLAM}  & 27.80 & 0.921  & 0.245  & 0.63 & 2.08 & 78.2 \\
Point-SLAM\cite{Point-SLAM} & 35.17 & 0.975 & 0.124 & 0.52 & \cellcolor{top2}0.44 & \cellcolor{top2}89.7\\
MonoGS\cite{MonoGS} & 37.50 & 0.960 & 0.070 & 0.58 & 0.95 & 78.6 \\
SplaTAM\cite{SplaTAM} & 34.11 & \cellcolor{top3} 0.978 & 0.104 & 0.36 & 0.72 & 86.1 \\
Gaussian SLAM\cite{Gaussian-SLAM} & \cellcolor{top1} 42.08 & \cellcolor{top1} 0.996 & \cellcolor{top1} 0.018 & \cellcolor{top3}0.31 & \cellcolor{top3}0.68 & \cellcolor{top3}88.9\\
GS-ICP\cite{GS-ICP-SLAM} & \cellcolor{top3}38.83 & 0.975 & \cellcolor{top3}0.041 & \cellcolor{top2}0.16 & - & -\\
\textbf{GauS-SLAM(Ours)}  & \cellcolor{top2}40.25 & \cellcolor{top2}0.991 & \cellcolor{top2}0.027 & \cellcolor{top1}0.06 & \cellcolor{top1}0.43 & \cellcolor{top1} 90.5\\
\bottomrule
\end{tabular}
}
\caption{\textbf{Comparison of Tracking and Reconstruction Performance on the Replica Dataset.} The best results are highlighted as \colorbox{top1}{first}, \colorbox{top2}{second}, and \colorbox{top3}{third}. Same methods we employed to extract meshes and compute metrics for both MonoGS\cite{MonoGS} and SplaTAM\cite{SplaTAM}. Other results are from the respective papers.}
\label{table: replica}
\end{table}
\begin{table}
\centering
 
\resizebox{0.49\textwidth}{!}{
\begin{tabular}{lcccccc}
\toprule

\multirow{2}{*}{\textbf{Dataset}} & \multicolumn{2}{c}{\underline{TUM-RGBD\cite{TUM-RGBD}}} & \multicolumn{2}{c}{\underline{ScanNet\cite{ScanNet}}} & \multicolumn{2}{c}{\underline{ScanNet++\cite{scannetpp}}}\\
& \texttt{fr2} & \texttt{fr3} &  \texttt{0059} & \texttt{0169} & \texttt{S1} & \texttt{S2} \\
\midrule
ORB-SLAM2\cite{ORB-SLAM2}
 & \cellcolor{top1} 0.40 & \cellcolor{top1}1.00 & 14.25 &\cellcolor{top2}8.72 & \textcolor{red}{\ding{55}}& \textcolor{red}{\ding{55}}\\ 

Point-SLAM\cite{Point-SLAM} &
1.31 & 3.48 & \cellcolor{top3}7.81 & 22.16 & \textcolor{red}{\ding{55}} & \textcolor{red}{\ding{55}} \\  
MonoGS*\cite{MonoGS} &
1.77 & \cellcolor{top3}1.49 & 32.10 & 10.70 & 7.00 & 3.66 \\
SplaTAM\cite{SplaTAM} & 
\cellcolor{top2}1.24 & 5.16 & 10.10 & 12.10  & 1.91 & \cellcolor{top2}0.61 \\
Gaussian-SLAM*\cite{Gaussian-SLAM} &
1.39 & 5.31 & 12.80 &16.30  &\cellcolor{top3}1.37 &\cellcolor{top3}2.28\\

LoopSplat*\cite{LoopSplat} &
\cellcolor{top3}1.30 & 3.53 &\cellcolor{top1}7.10 & \cellcolor{top3}10.60& \cellcolor{top2}1.14  &3.16 \\

\textbf{GauS-SLAM(Ours)} &
1.34 & \cellcolor{top2}1.46 & \cellcolor{top2}7.14 & \cellcolor{top1}7.45& \cellcolor{top1}0.42 &\cellcolor{top1}0.47  \\
\bottomrule
\multicolumn{7}{r}{  \textcolor{red}{\ding{55}} indicates a large trajectory error, even in the first 250 frames}\\
\end{tabular}
}
\caption{\textbf{Comparison of Tracking Performance(ATE-RMSE $\downarrow$[cm]).} The methods marked with an "*" were evaluated only on the first 250 frames of the ScanNet++\cite{scannetpp} dataset, as the large camera disparity in later frames often led to tracking failures. }
\label{table:tracking_performance}
\end{table}

\begin{figure}[H]
    \centering
    \setlength{\tabcolsep}{1pt}
    \renewcommand{\arraystretch}{0.5}
    \small
    \begin{tabular}{cccc}
        \footnotesize \rotatebox{90}{\texttt{Room 0}} & 
    	\includegraphics[width=0.32\linewidth]{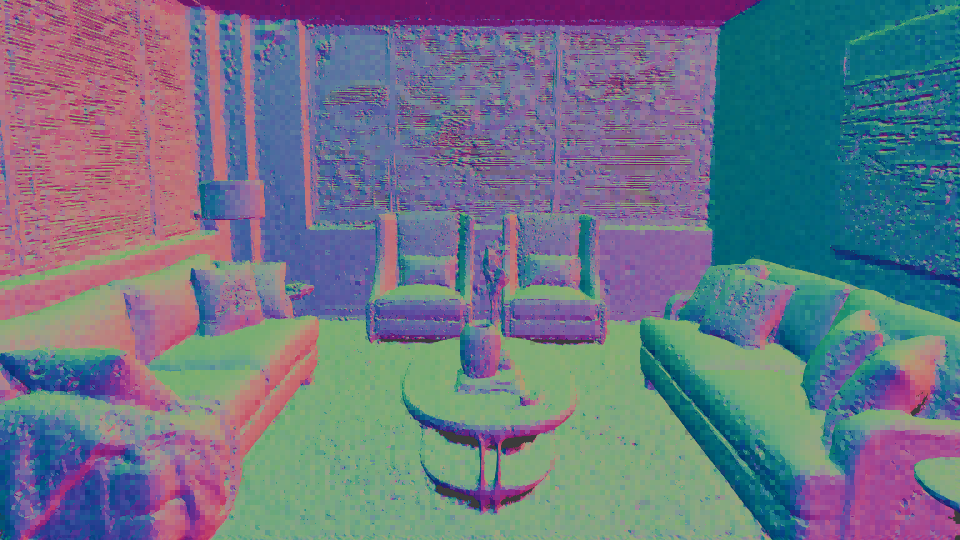} &
    	\includegraphics[width=0.32\linewidth]{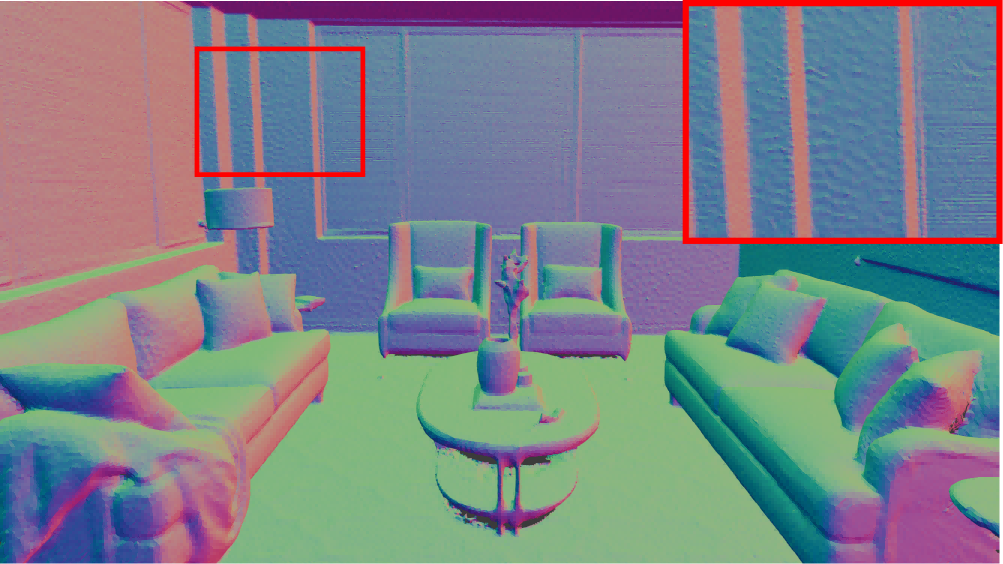} & 
    	\includegraphics[width=0.32\linewidth]{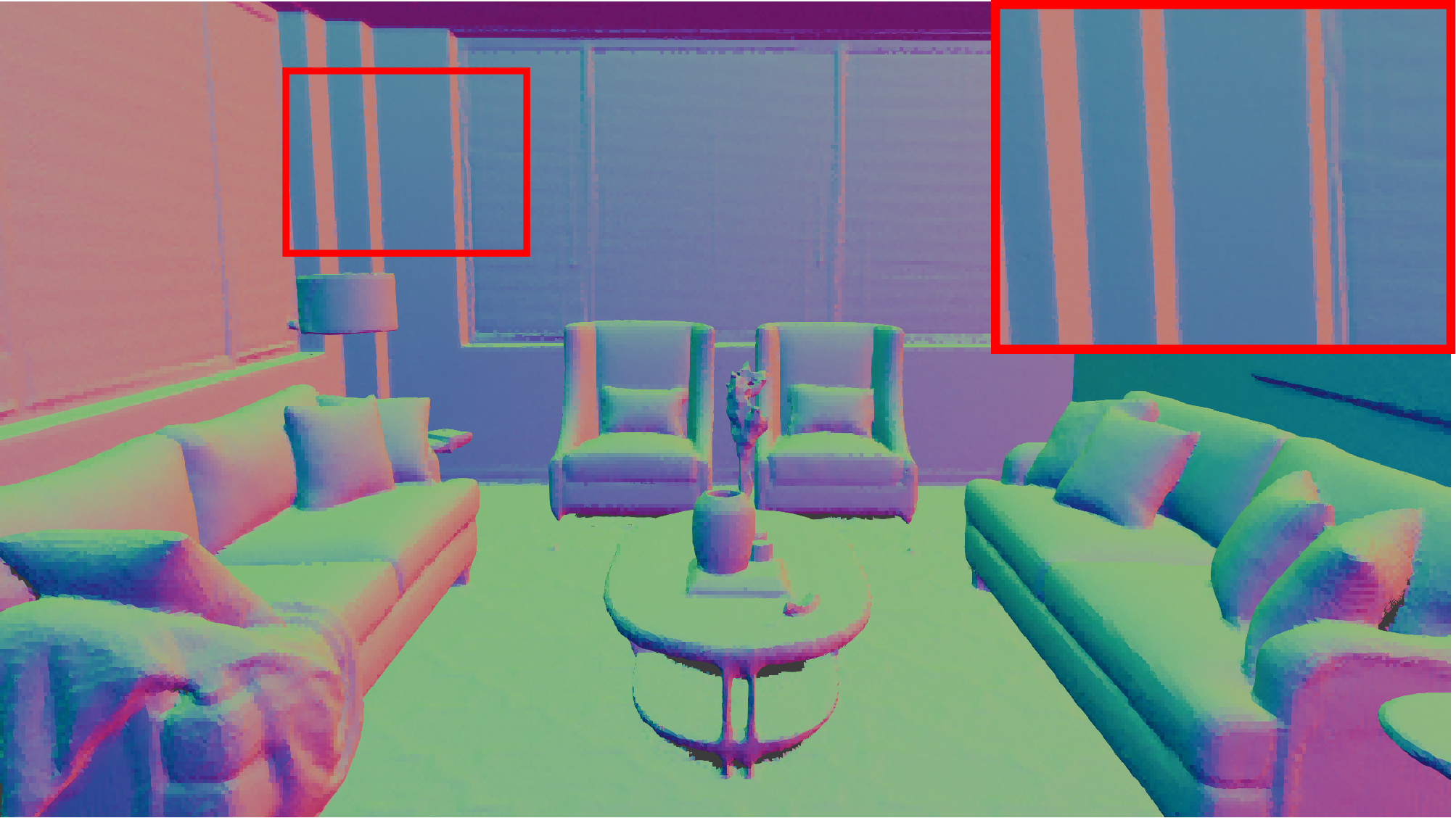} \\
        \footnotesize \rotatebox{90}{\texttt{Room 2}} & 
    	\includegraphics[width=0.32\linewidth]{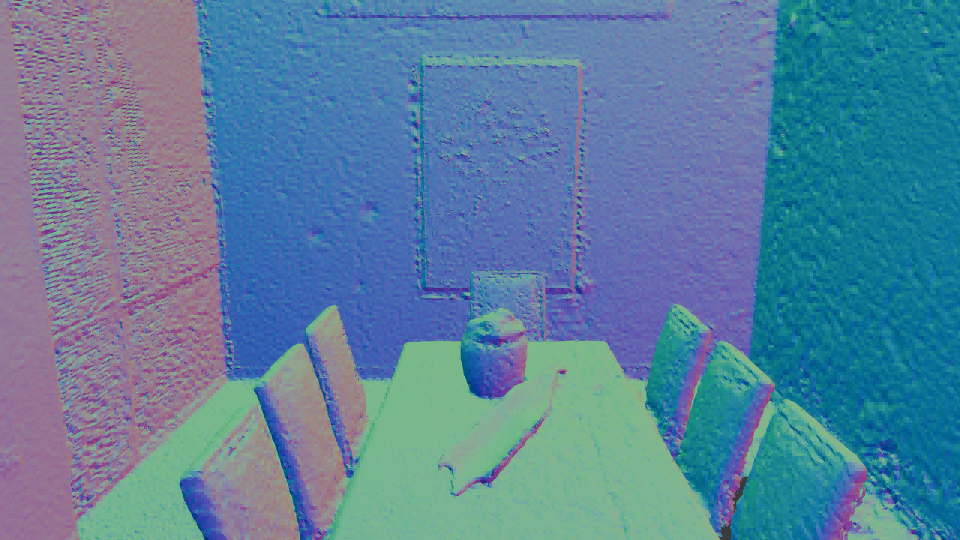} & 
    	\includegraphics[width=0.32\linewidth]{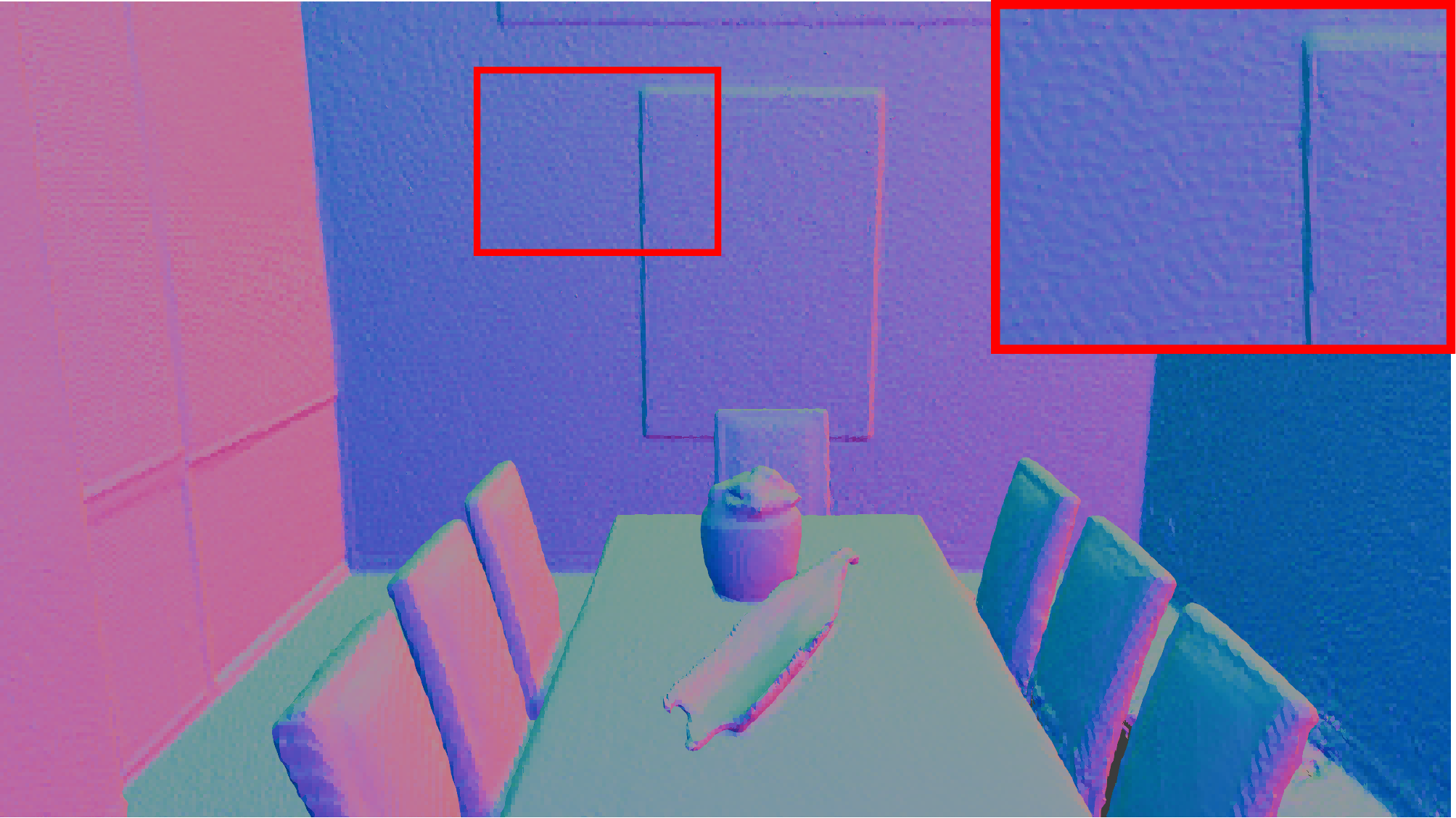} & 
    	\includegraphics[width=0.32\linewidth]{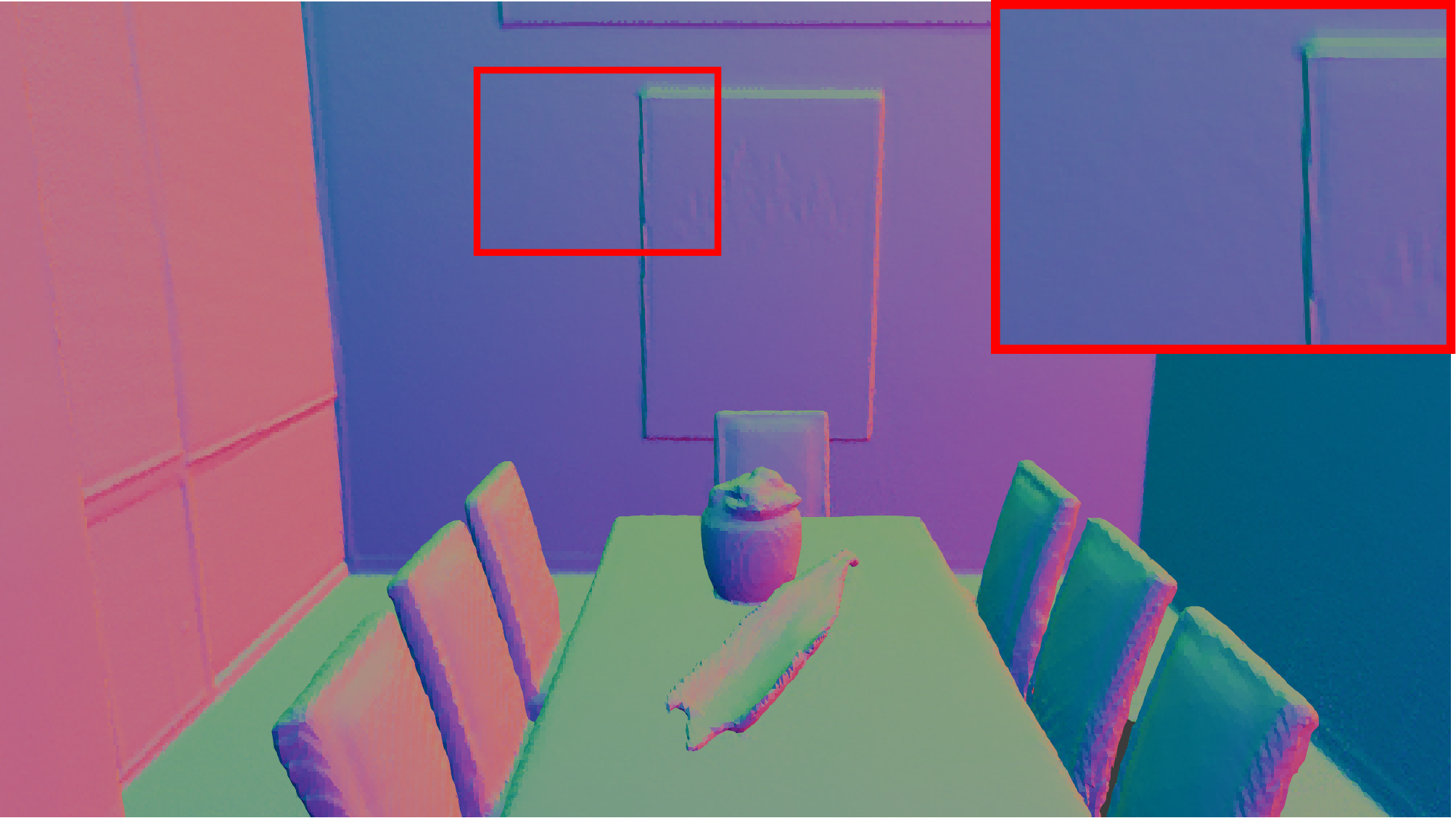} \\ 
        \footnotesize \rotatebox{90}{\texttt{Office 2}} & 
    	\includegraphics[width=0.32\linewidth]{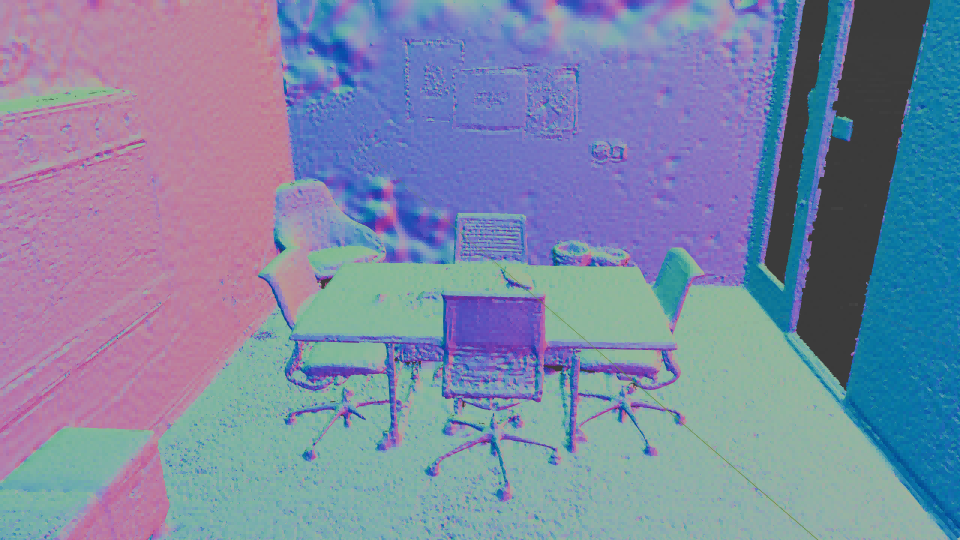} & 
    	\includegraphics[width=0.32\linewidth]{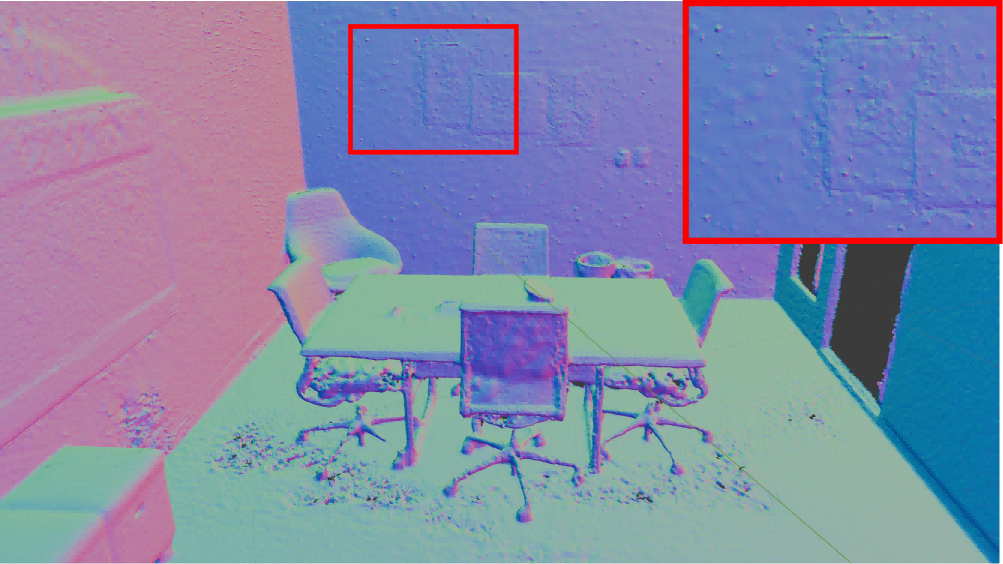} & 
    	\includegraphics[width=0.32\linewidth]{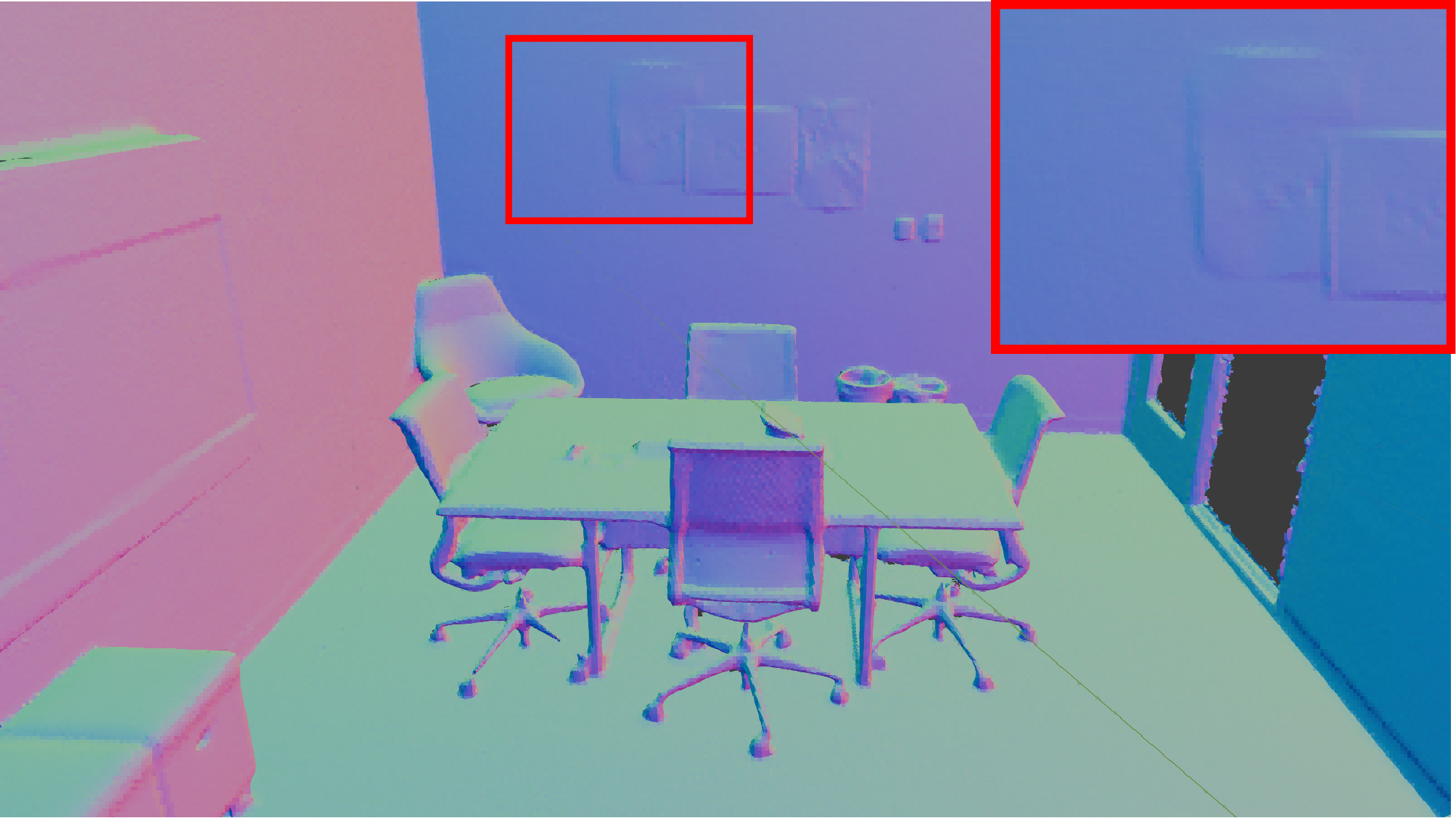} \\ 
        & MonoGS\cite{MonoGS} & SplaTAM\cite{SplaTAM} & \textbf{GauS-SLAM(Ours)} \\
    \end{tabular}
    \caption{\textbf{Comparison of mesh results on Replica\cite{Replica}.} Compared to isotropic 3D Gaussians, Gaussian surfels produce smoother mesh reconstructions.}
    \label{fig:mesh_eval}
\end{figure}
\paragraph{Rendering and Reconstruction Performance.}
In the \cref{table: replica}, we present the rendering and reconstruction performance of GauS-SLAM on the Replica \cite{Replica} dataset. Although the rendering quality of 2DGS \cite{2DGS} has been empirically demonstrated to be inferior to that of 3DGS \cite{3DGS}, GauS-SLAM surpasses the majority of 3D Gaussian-based methods. Notably, it outperforms our baseline algorithm, SplaTAM \cite{SplaTAM}, by 6 dB in PSNR. This improvement is attributed to our novel local map-based design, which enables more accurate initialization of Gaussian primitives. A more detailed comparison is presented in \cref{fig:replica_rendering_comparison}.
\begin{figure}
    \centering
    \setlength{\tabcolsep}{1pt}
    \renewcommand{\arraystretch}{0.5}
    \small
    \begin{tabular}{cccc}
        \footnotesize \rotatebox{90}{\texttt{Frame 40}} & 
    	\includegraphics[width=0.32\linewidth]{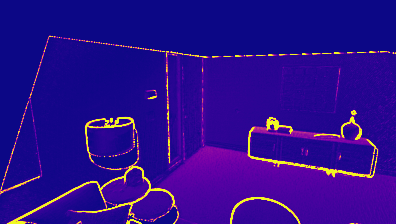} &
    	\includegraphics[width=0.32\linewidth]{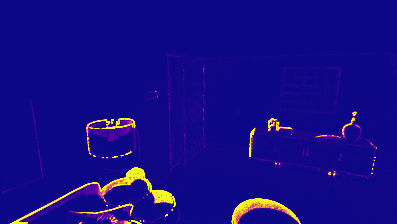} & 
    	\includegraphics[width=0.32\linewidth]{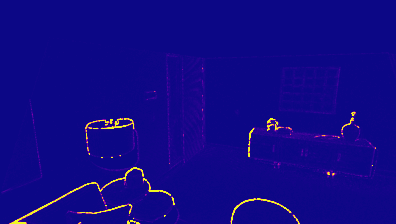} \\
        \footnotesize \rotatebox{90}{\texttt{Frame 55}} & 
    	\includegraphics[width=0.32\linewidth]{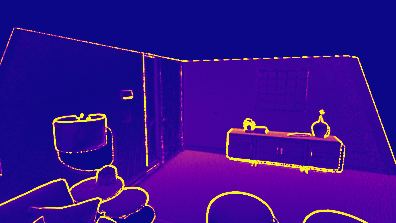} & 
    	\includegraphics[width=0.32\linewidth]{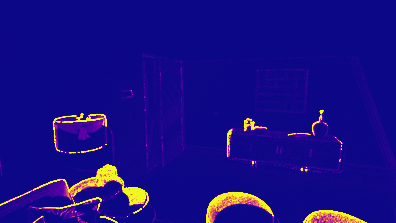} & 
    	\includegraphics[width=0.32\linewidth]{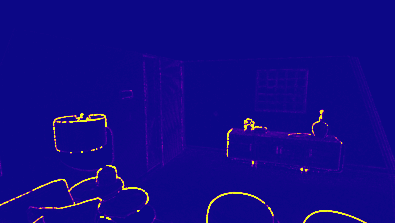} \\ 
        & 3DGS\cite{3DGS} & 2DGS\cite{2DGS} & \textbf{GauS-SLAM(Ours)} \\
    \end{tabular}
    \caption{In the geometry consistency experiment, the error maps of the depth rendering results on the \texttt{Frame 40} and \texttt{Frame 55} of \texttt{Room 0} in the Replica dataset.}
    \label{fig:better_multiview_depth_rendering}
\end{figure}
By employing the Surface-aware Depth Rendering approach, our method achieves superior performance in both Depth L1 and F1-Score compared to other Gaussian-based algorithms. It is noteworthy that the isotropic Gaussian primitives \cite{SplaTAM,MonoGS} tend to produce uneven mesh surfaces, as depicted in \cref{fig:mesh_eval}, whereas the 2D Gaussian surfels yield significantly smoother results.

\begin{table}
\centering
\small

\label{tab:results}
\begin{tabular}{ccccc}
\toprule
\multirow{2}{*}{\textbf{Methods}}  
& \textbf{Tracking}  & \textbf{Mapping} & \textbf{ATE} & \textbf{PSNR} \\
& \textbf{/Frame} & \textbf{/Frame}  &[cm]$\downarrow$ & [dB]$\uparrow$\\
\midrule
Point-SLAM & \cellcolor{top2}0.47 & \cellcolor{top3}2.57 & 0.49 & \cellcolor{top3}33.10\\
SplaTAM & 2.07 & 3.73 & \cellcolor{top3}0.31 & 32.56\\
\textbf{GauS-SLAM} & \cellcolor{top3}1.04 & \cellcolor{top2}0.65 & \cellcolor{top1}0.05 & \cellcolor{top1}38.20\\
\textbf{GauS-SLAM-S} & \cellcolor{top1}0.42 & \cellcolor{top1}0.16 & \cellcolor{top2}0.10 & \cellcolor{top2}37.01\\
\bottomrule
\end{tabular}
\caption{\textbf{Runtime on Replica/Room0 using a A6000.} The average per-frame time for tracking and mapping are computed by dividing the total processing time of the tracking and mapping procedures by the total number of frames.}
\label{tab: runtime}
\end{table} 
\paragraph{Geometry consistency}
To evaluate the geometry consistency of the rendering method, we design the following experiment. First, we fully train the model on the first four frames of the Replica \texttt{Room0} dataset using ground-truth poses. We then compute the average L1 error of the rendered depth maps across the first 60 viewpoints as the metric for geometry consistency assessment, which is utilized in ablation studies.
We present the error maps for \texttt{Frame 40} and \texttt{Frame 55} in \cref{fig:better_multiview_depth_rendering}. While 2DGS \cite{2DGS} demonstrates higher view consistency, significant depth errors are also observed at object boundary regions. This phenomenon occurs because depth values from different surfaces collectively influence the final rendering depth. Our proposed surface-aware depth rendering strategy effectively mitigates the impact of occluded surfaces on the rendering results, thereby enhancing geometry consistency.

\paragraph{Runtime Comparison}
\cref{tab: runtime} presents the average per-frame time for tracking and mapping process of GauS on the \texttt{Room 0} sequence(resolution $1200\times680$). Compared to our baseline SplaTAM \cite{SplaTAM}, GauS-SLAM
 We demonstrate significant improvements not only in rendering quality and tracking accuracy but also achieve a more than threefold enhancement in time efficiency. Specifically, we developed a more efficient model, GauS-SLAM-S, which reduces the number of tracking iterations from 40 to 25 and mapping iterations from 40 to 30, while decreasing the keyframe threshold $\tau_k$ to 5\%. 

\subsection{Ablation Study}
In this section, we conduct ablation studies on the depth rendering and SLAM components of our proposed method. By systematically removing components, we demonstrate the superior rendering quality and tracking accuracy.

\paragraph{Depth rendering ablation}
We conducted ablation studies on improvements to depth rendering, including unbiased depth, depth adjustment, depth normalization, and regularization loss. The effectiveness of these strategies was evaluated using three metrics: geometry consistency, trajectory error, and rendering quality. The results are presented in \cref{tab:ablation_study_on_render_model}. Among these, unbiased depth and depth normalization had the most significant impact on the results. Removing them led to notable degradation. Depth adjustment effectively reduces the mutual influence among different surfaces, thereby improving tracking performance.

\paragraph{Ablation on SLAM components}
We conducted ablation studies on the SLAM system to analyze the impact of key components, including keyframes, local mapping, random optimization, and final refinement. We employed ATE-RMSE \cite{ATE-RMSE}, PSNR, and the average processing time per frame as metrics to evaluate the impact of these modules on system accuracy and efficiency. Notably, we utilized \texttt{Room0} from Replica \cite{Replica} and \texttt{fr3/office} from TUM \cite{TUM-RGBD}, the latter of which represents a typical scenario where the camera moves around an object. The experimental results are presented in \cref{tab:ablation_study_on_vs_slam}. 

In Experiment E, we ablate keyframes and performed mapping for each frame. This approach reduces tracking errors to some extent, but at the cost of significantly increased computational time. In Experiment F, we replace the front-end and the back-end based on local maps with the tracking and mapping framework proposed by SplaTAM \cite{SplaTAM}. The results indicate that this modification greatly reduced system efficiency, with a noticeable decline in accuracy in the \texttt{fr3/office} sequence. This suggests that the design of the local map plays a crucial role in tracking with cameras that perform object-circling movements. Experiments G and H conducted ablation studies on the system's back-end, specifically on random optimization and final refinement. The results show that random optimization helps improve tracking accuracy, while final optimization enhances rendering quality.

\begin{table}
\centering
\resizebox{0.49\textwidth}{!}{
\begin{tabular}{lccccc}
\toprule
\multirow{2}{*}{\textbf{Methods}} & \textbf{Geo. Con} & \textbf{ATE} &  \textbf{PSNR} \\ 
& [mm]$\downarrow$ &  [mm]$\downarrow$  &  [dB]$\uparrow$ \\

\midrule
A. w/o Unbiased Depth & 1.94 & 2.10 & 36.06 \\
B. w/o Depth Adjustment & 1.75 & 0.85 & 38.10 \\
C. w/o Depth Norm. & 2.51 & 1.92 & 35.98 \\
D. w/o Regulation Loss  & \textbf{1.01} & 0.63 & \textbf{38.25} \\
Full Model & \textbf{1.01}  & \textbf{0.60} & 38.04 \\
\bottomrule
\end{tabular}
}
\caption{\textbf{Ablation Study on Depth Rendering.} In Experiment A, we exclusively substituted 2D Gaussian surfels in rendering model with 3D Gaussian primitives.} 
\label{tab:ablation_study_on_render_model}
\end{table}
\begin{table}
\centering
 
\small
\resizebox{0.49\textwidth}{!}{
\begin{tabular}{llccc}
\toprule
\multirow{2}{*}{\textbf{Methods}} & \multirow{2}{*}{\textbf{Sequence}}  & \textbf{ATE} &  \textbf{PSNR} & \textbf{Time}\\ 
& & [mm]$\downarrow$  &  [dB]$\uparrow$ & [s]$\downarrow$\\
\midrule
\multirow{2}{*}{\makecell[l]{E. w/o \\ Keyframe}} 
& \texttt{Room 0} & 0.52 & \textbf{38.28} & 2.13\\
& \texttt{fr3/office} & 14.53 & 25.03 & 1.72\\
\midrule

\multirow{2}{*}{\makecell[l]{F. w/o \\ LocalMap}} 
& \texttt{Room 0}& \textbf{0.49} & 38.25 & 6.77 \\
& \texttt{fr3/office} & 52.91 & 24.16 & 5.58 \\
\midrule

\multirow{2}{*}{\makecell[l]{G. w/o Random \\ Optimization}} 
& \texttt{Room 0}& 0.70 & 37.78 & \textbf{1.63} \\
& \texttt{fr3/office} & 14.37 & 25.03 & 1.62 \\
\midrule

\multirow{2}{*}{\makecell[l]{H. w/o Final  \\ Refinement}} 
& \texttt{Room 0} & 0.54 & 37.48 & 1.73\\
& \texttt{fr3/office} & 14.30 & 24.34 & 1.63\\
\midrule

\multirow{2}{*}{Full Model} 
& \texttt{Room 0} & 0.60 & 38.04 & 1.73 \\
& \texttt{fr3/office} & \textbf{14.29} & \textbf{25.06} & \textbf{1.62} \\
\bottomrule
\end{tabular}
}
\caption{\textbf{Ablation Study on GauS-SLAM System.} In Experiment F, we implemented an alternating execution paradigm for tracking and mapping following the SplaTAM\cite{SplaTAM} framework }
\label{tab:ablation_study_on_vs_slam}
\end{table}

\section{Conclusion}
In this paper, we address two critical challenges in camera tracking within Gaussian representation: geometric distortions in multi-view scenarios and outlier rejection during the frame-to-model alignment process. To address these issues, we propose a Surface-aware Depth Rendering strategy based on 2DGS\cite{2DGS} and design a SLAM system that integrates keyframes and local maps. Our experimental results demonstrate that the proposed GauS-SLAM outperforms the baseline methods in both tracking and rendering on four benchmark datasets. In particular, it achieves SOTA tracking performance in the Replica\cite{Replica} and ScanNet++\cite{scannetpp} datasets, underscoring the efficacy of 2D Gaussian in camera tracking tasks. 
\paragraph{Limitation and Future Work}
GauS-SLAM, similar to other Gaussian-based algorithms, exhibits significant sensitivity to motion blur and exposure variations. This limitation leads to suboptimal tracking performance on benchmark datasets such as TUM-RGBD\cite{TUM-RGBD} and ScanNet \cite{ScanNet}. For future work, we will focus on enhancing the robustness to these factors causing multi-view inconsistency.
{
    \small
    \bibliographystyle{ieeenat_fullname}
    \bibliography{main}
}

\clearpage
\setcounter{page}{1}
\maketitlesupplementary

\begin{abstract}
This supplementary material encompasses an evaluation video of GauS-SLAM on the S2(\textit{8b5caf3398}) sequence of ScanNet++. Furthermore, we present comprehensive experimental results and visualizations across multiple benchmark datasets, including Replica, ScanNet, TUM-RGBD, and ScanNet++. In addition, we provide an extended analysis and discussion on the ablation studies on the local map and the runtime evaluation.
\end{abstract}

\section{Video}
We submit a video \texttt{GauS\_SLAM\_S2.mp4}. This video demonstrates the experimental results of GauS-SLAM in a challenging real-world scenario \texttt{S2}. It records the back-end operation of GauS-SLAM, where the deep blue lines represent the camera trajectory, the blue frustums denote keyframes, and the larger orange frustums indicate the reference keyframes of each submap. Throughout the video, GauS-SLAM continuously integrates local maps received from the front-end, optimizing the reconstruction and removing redundant Gaussian surfels after each fusion. Despite the large disparity between adjacent frames, GauS-SLAM maintains stable tracking and reconstruction. Additionally, the video highlights the effectiveness of Edge Growth in reconstructing depth-missing regions, such as the window area. 

\section{Implementation details}
\paragraph{Hyperparameters}
\cref{table:hyperparamters} presents the principal hyperparameters used in our experiments on Replica\cite{Replica}, TUM-RGBD\cite{TUM-RGBD}, ScanNet\cite{ScanNet} and ScanNet++\cite{scannetpp}. The learning rates for camera rotation $l_r$ and translation $l_t$ during tracking are dynamically adjusted using an exponential decay strategy, which is consistently applied in both front-end and back-end optimization processes. Notably, these learning rates are adaptively modulated based on the camera's motion velocity, resulting in dataset-specific variations. Furthermore, the table enumerates the number of iterations for tracking($iter_t$) and mapping($iter_m$) across different datasets. In scenarios with large disparity between adjacent frames, an increased number of tracking iterations is implemented to ensure the convergence of pose optimization.
\begin{table}[h]
\centering
\resizebox{0.5\textwidth}{!}{
\begin{tabular}{lcccc}
\toprule
\textbf{Params} & Replica & TUM-RGBD & ScanNet & ScanNet++ \\ 
\midrule
$l_r$ &  0.0004 & 0.0008 & 0.0008 & 0.01\\
$l_t$ &  0.002 & 0.004 & 0.004 & 0.04\\
$iter_t$ & 40 & 120 & 100 & 120 \\
$iter_m$ & 40 & 40  & 40 & 60 \\
\bottomrule
\end{tabular}
}
\caption{\textbf{Per-dataset Hyperparameters.}}
\label{table:hyperparamters}
\end{table}

\paragraph{Pose Initialization} 
Similar to the majority of SLAM algorithms, the constant velocity model is employed for initializing the pose of each frame. Specifically, the initial pose $T_i'$ of the $i$-th frame can be derived using the following formulation.
\begin{equation}
\mathbf{T}_i' = (\mathbf{T}_{i-1}\mathbf{T}_{i-2}^{-1})\cdot \mathbf{T}_{i-1}
\end{equation}
where $T_{i-1}$ denotes the predicted pose of the preceding frame.
\paragraph{Exposure compensation} 
Exposure variation poses a significant challenge in 3D reconstruction using Gaussian Splatting, particularly in the incremental reconstruction task. To address this issue, we employ a simple linear compensation method by introducing learnable compensation coefficients $a$, $b$ for each frame. During rendering, the color map is corrected according to the following formulation:
\begin{equation}
\mathbf{I'}=a\mathbf{I}+b
\end{equation}
\paragraph{Re-tracking} In the front-end tracking process, significant disparity can result in the loss of tracking for specific frames in the local map. This phenomenon is particularly observed in the \texttt{S1} sequence of the ScanNet++\cite{scannetpp} dataset. To address this issue, we propose a re-tracking strategy. When the rendered depth error significantly exceeds the average level, the front-end will flag it as a lost frame. Subsequently, the local map is reset, and the lost frame is designated as the reference keyframe for a new local map, enabling the continuation of the front-end. Upon receiving a lost frame, our tracking approach will extend beyond local map by leveraging a broader set of submaps, thereby enhancing the robustness of the tracking process
\begin{table}[ht]
\centering
\resizebox{0.5\textwidth}{!}{
\begin{tabular}{lcccc|ccc}
\toprule
\multirow{2}{*}{\textbf{Methods}} & \multirow{2}{*}{\textbf{Metrics}} & \multicolumn{3}{c}{\textbf{Novel View}} & \multicolumn{3}{c}{\textbf{Training View}} \\ 
                  &                   & S1    & S2 & \textbf{Avg.}    & S1    & S2   & \textbf{Avg.}  \\ \midrule
\multirow{3}{*}{Point-SLAM\cite{Point-SLAM}}   & PSNR [dB] $\uparrow$ & 12.10 & 11.73 & 11.91  & 14.62 & 14.30 & 14.46\\ 
                  & SSIM $\uparrow$      & 0.31  & 0.26 & 0.28  & 0.35  & 0.41  & 0.38  \\
                  & LPIPS $\downarrow$   & 0.62  & 0.74  & 0.68  & 0.68  & 0.62  & 0.65  \\ \midrule
\multirow{4}{*}{SplaTAM\cite{SplaTAM}}  & PSNR [dB] $\uparrow$ & 23.99 & 24.84  & 24.41& 27.82 & 28.14  &27.98  \\ 
                  & SSIM $\uparrow$    & 0.88  & 0.87   & 0.88  & 0.94  & 0.94  & 0.94   \\
                  & LPIPS $\downarrow$ & 0.21  & 0.26   & 0.24  & 0.12  & 0.13 & 0.12   \\
                  & Depth L1 [cm] $\downarrow$   & 1.91  & 2.23 & 2.07  & 0.93  & 1.64  & 1.28 \\ \midrule
\multirow{4}{*}{\textbf{GauS-SLAM(Ours)}}  & PSNR [dB] $\uparrow$ & \cellcolor{top1}25.48 & \cellcolor{top1}25.76 & \cellcolor{top1}25.62 & \cellcolor{top1}29.21 & \cellcolor{top1}30.07 & \cellcolor{top1}29.64  \\ 
                  & SSIM $\uparrow$      & \cellcolor{top1}0.89  & \cellcolor{top1}0.87 & \cellcolor{top1}0.88 & \cellcolor{top1}0.95  & \cellcolor{top1}0.95 & \cellcolor{top1}0.95  \\
                  & LPIPS $\downarrow$   & \cellcolor{top2}0.25  & \cellcolor{top1}0.23 & \cellcolor{top1}0.24   & \cellcolor{top2}0.16  & \cellcolor{top1}0.11  & \cellcolor{top2}0.13  \\
                  & Depth L1 [cm] $\downarrow$& \cellcolor{top1}1.15  & \cellcolor{top1}2.23 & \cellcolor{top1}1.69 & \cellcolor{top1}0.52  & \cellcolor{top1}0.62 & \cellcolor{top1}0.57 \\ \bottomrule
\end{tabular}
}
\caption{\textbf{Novel \& Train View Rendering Performance on Scan-Net++\cite{scannetpp}.}}
\label{tab:nvs}
\end{table}
\begin{figure*}[t]
    \centering
    \setlength{\tabcolsep}{1pt}
    \renewcommand{\arraystretch}{0.5}
    \begin{tabular}{ccccp{0.1cm}ccc}
    
        &\multicolumn{3}{c}{\textbf{Training view}} & & \multicolumn{3}{c}{\textbf{Novel view}} \\

                \footnotesize \multirow{2}{*}{\rotatebox{90}{\texttt{S1}}} &
        \multirow{2}{*}[4.5ex]{
        \includegraphics[width=0.16\linewidth]{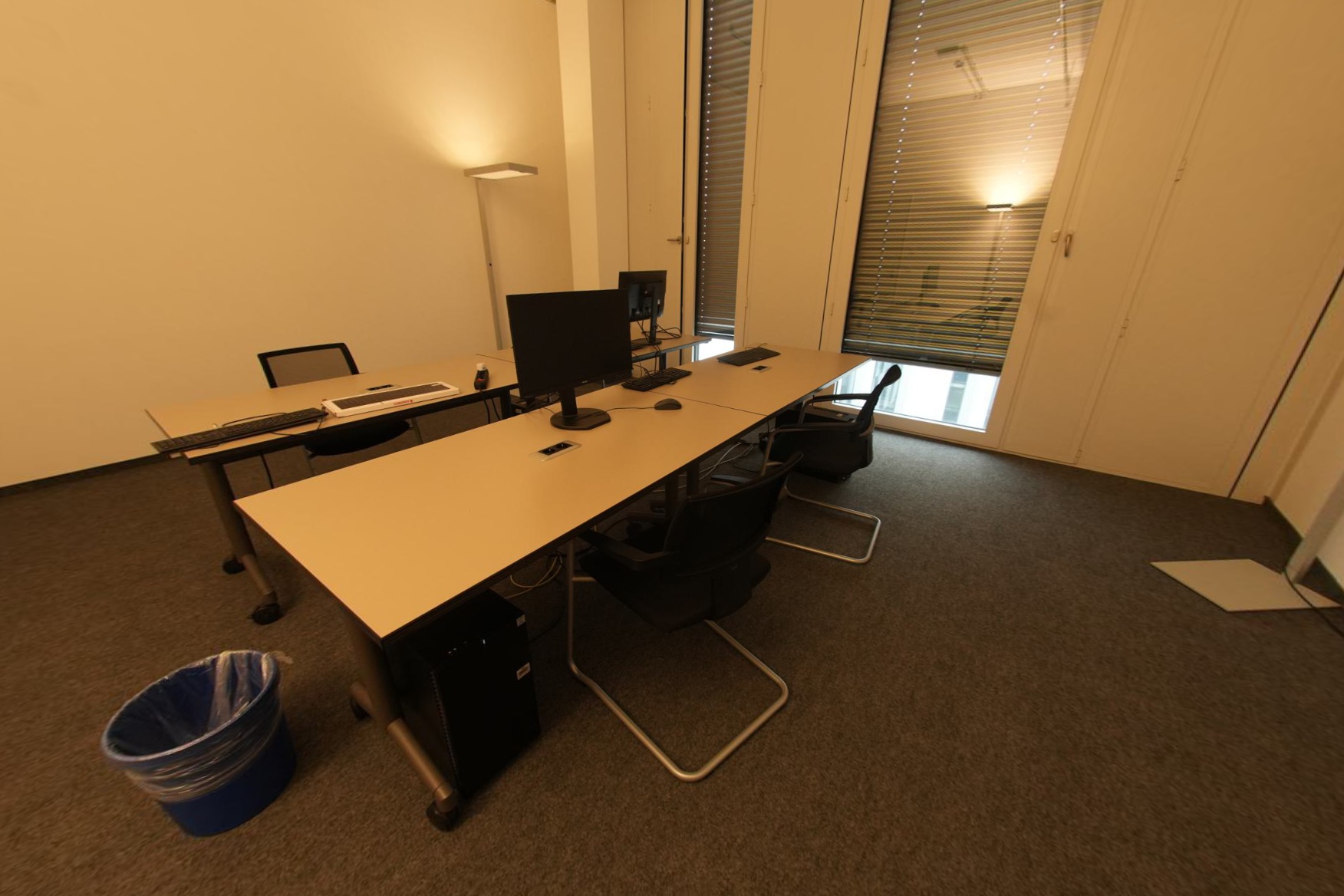} }&
        \includegraphics[width=0.16\linewidth]{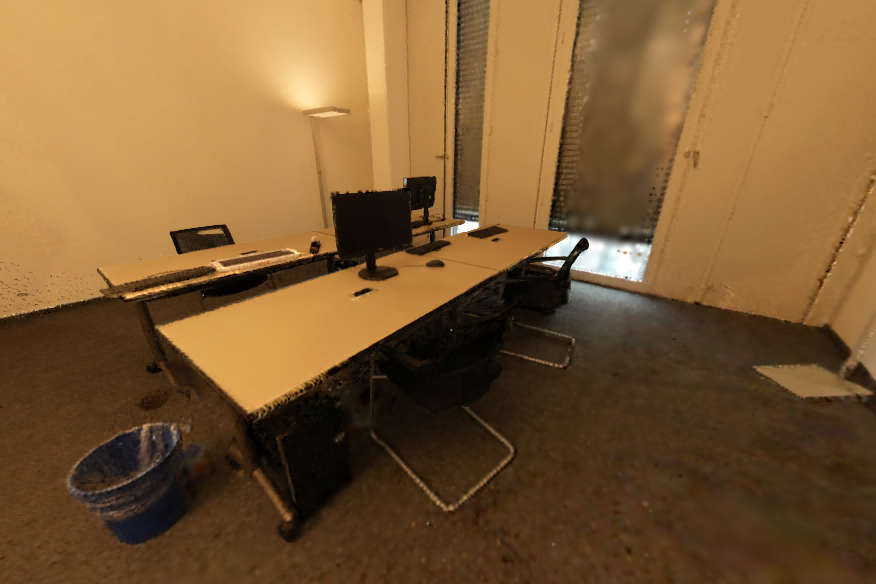} &
        \includegraphics[width=0.16\linewidth]{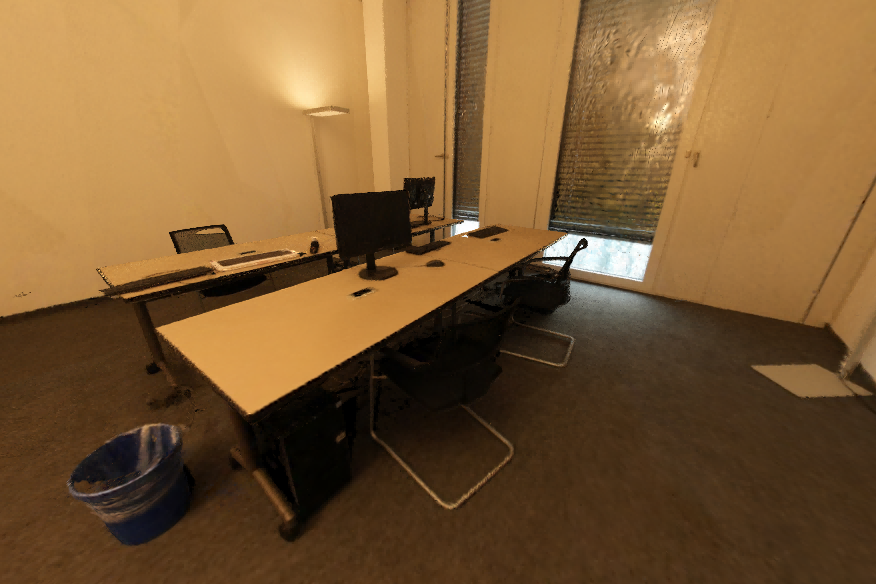} & & 
        \multirow{2}{*}[4.5ex]{
        \includegraphics[width=0.16\linewidth]{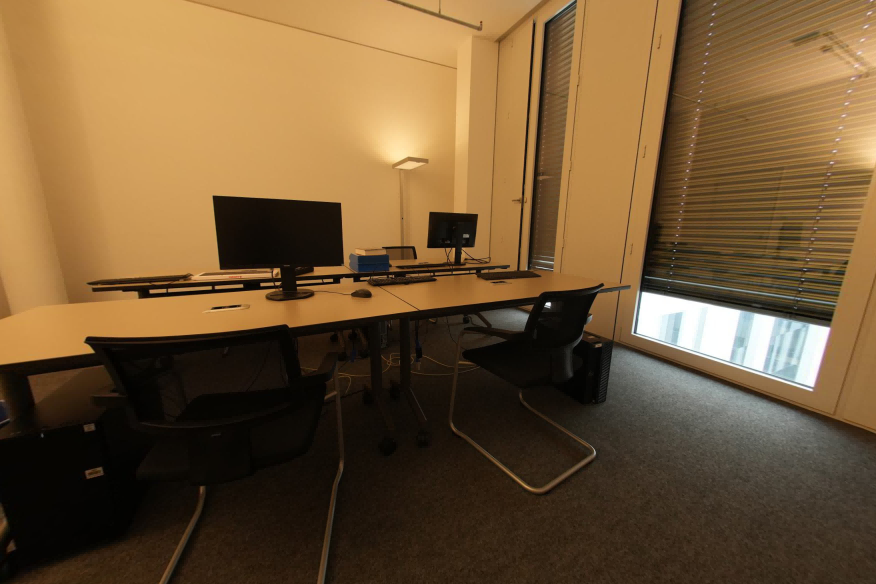} }&
        \includegraphics[width=0.16\linewidth]{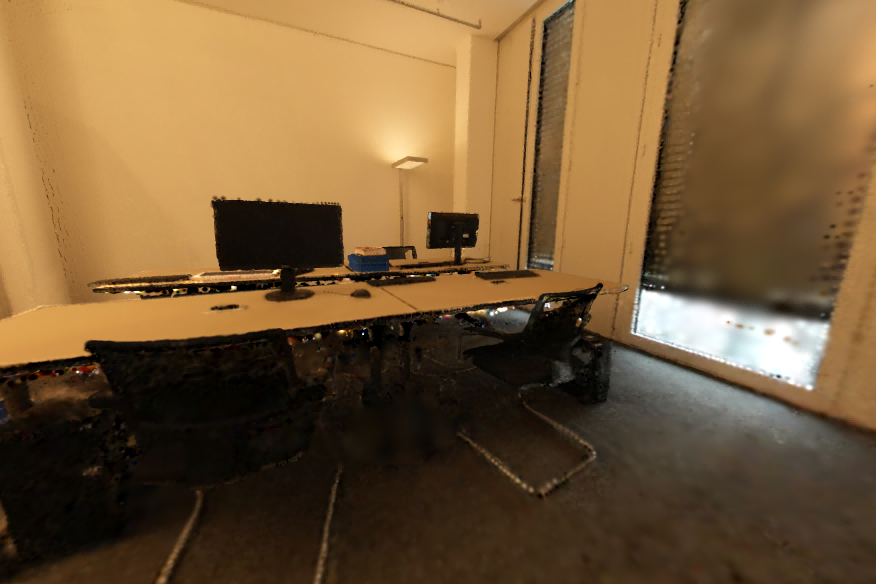} &
        \includegraphics[width=0.16\linewidth]{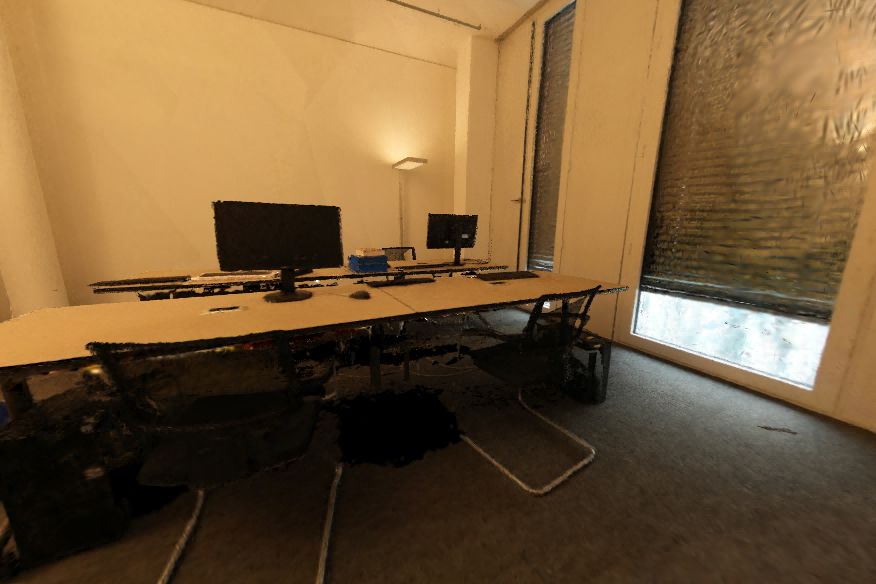}
        \\
        
        & &
        \includegraphics[width=0.16\linewidth]{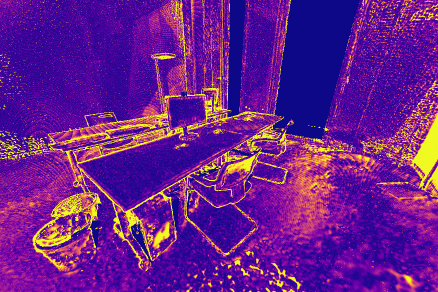} &
        \includegraphics[width=0.16\linewidth]{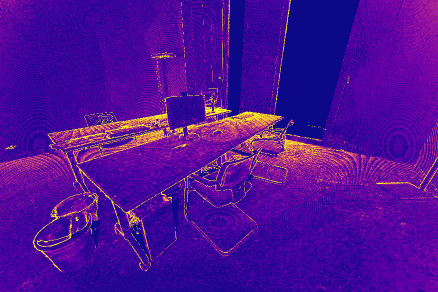} & & 
        &
        \includegraphics[width=0.16\linewidth]{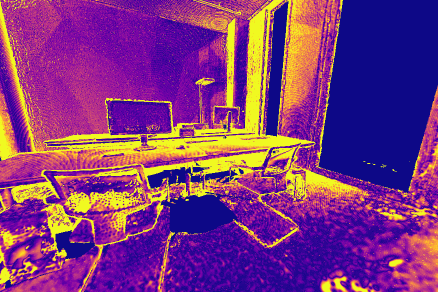} & 
        \includegraphics[width=0.16\linewidth]{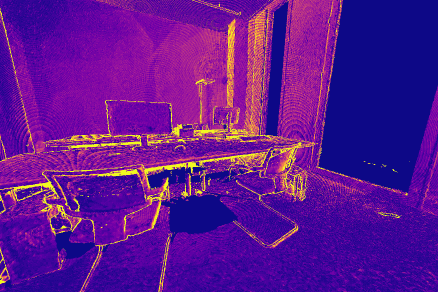} \\
        
        \footnotesize \multirow{2}{*}{\rotatebox{90}{\texttt{S2}}} &
        \multirow{2}{*}[4.5ex]{
        \includegraphics[width=0.16\linewidth]{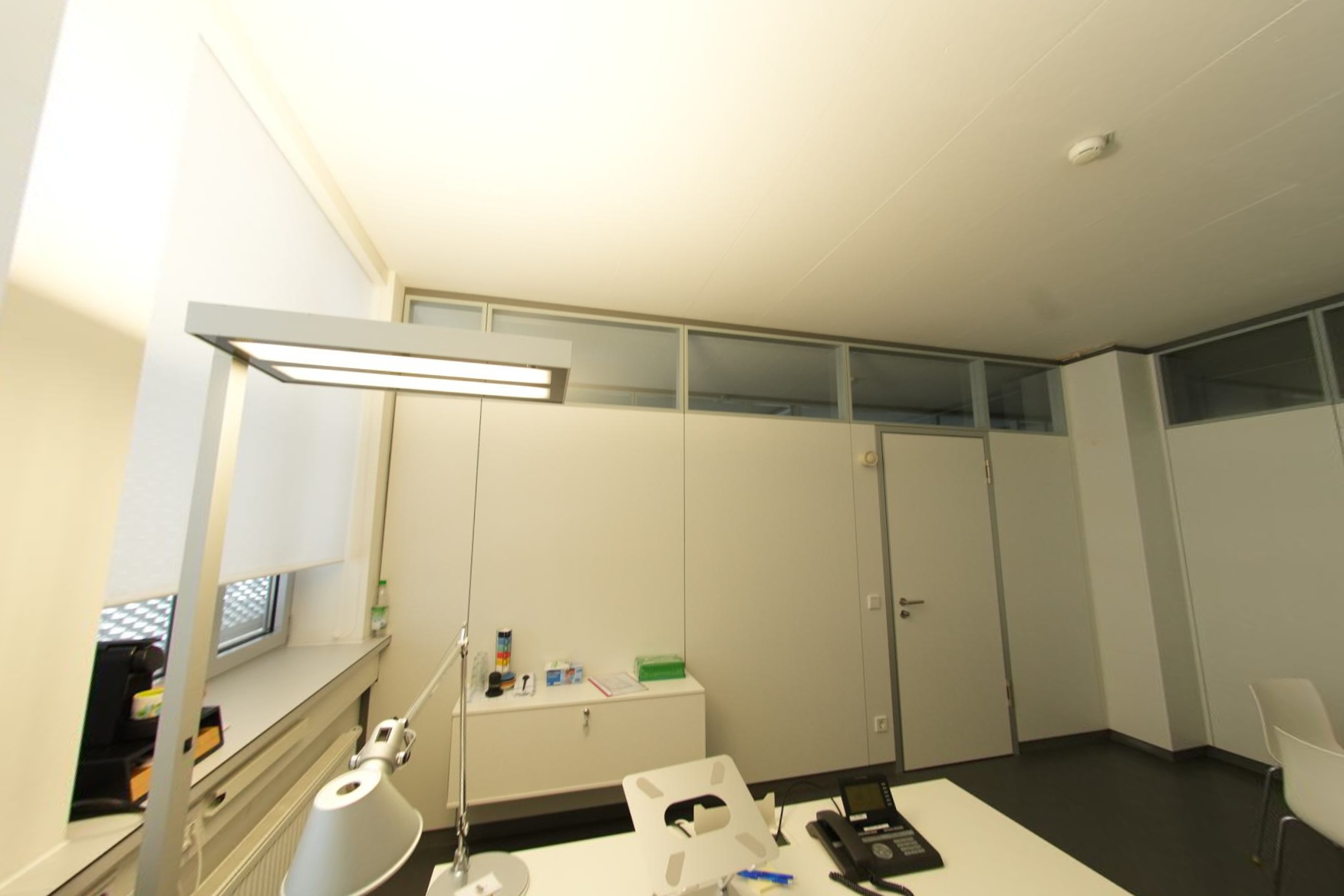}} &
        \includegraphics[width=0.16\linewidth]{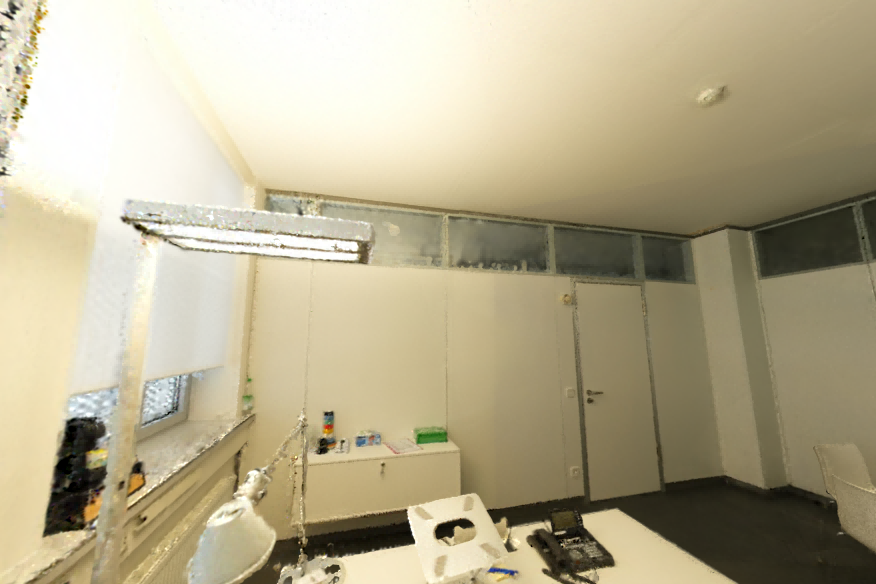} &
        \includegraphics[width=0.16\linewidth]{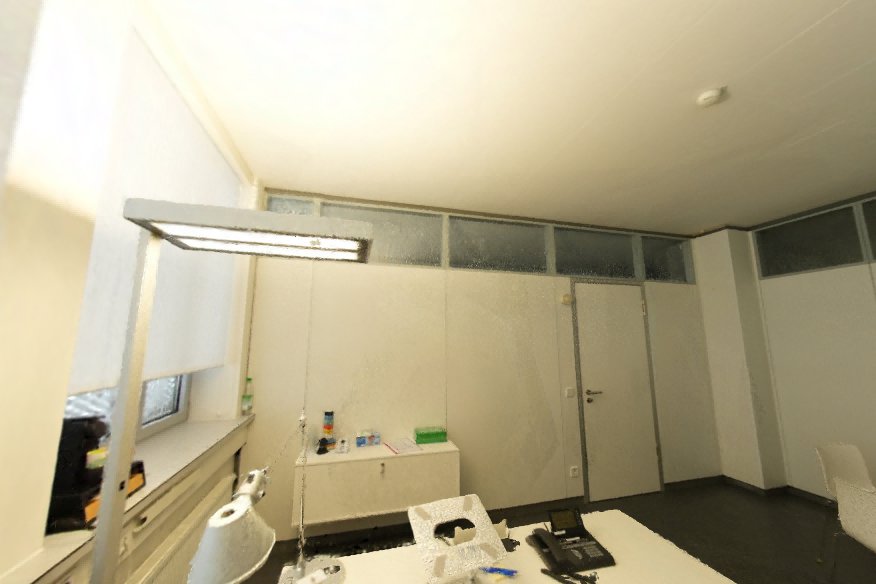} & &
        \multirow{2}{*}[4.5ex]{
        \includegraphics[width=0.16\linewidth]{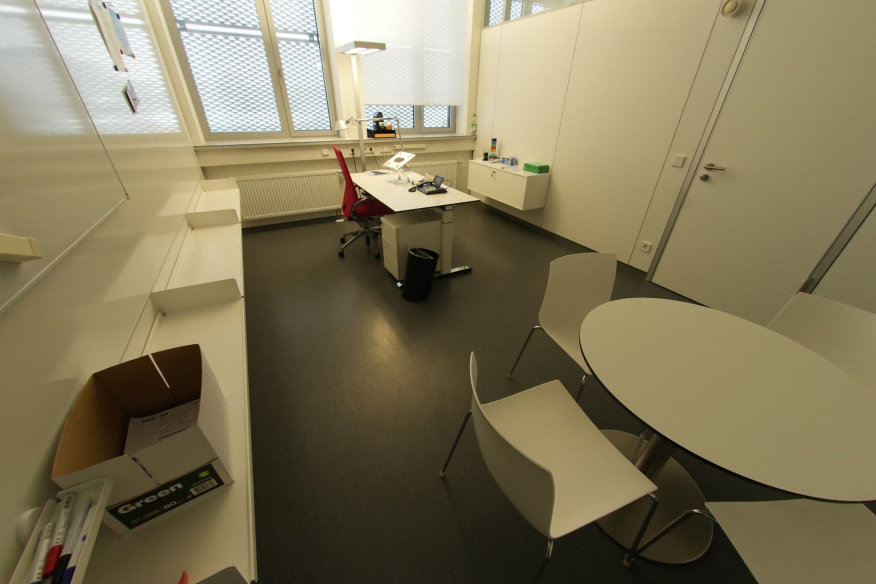} }&
        \includegraphics[width=0.16\linewidth]{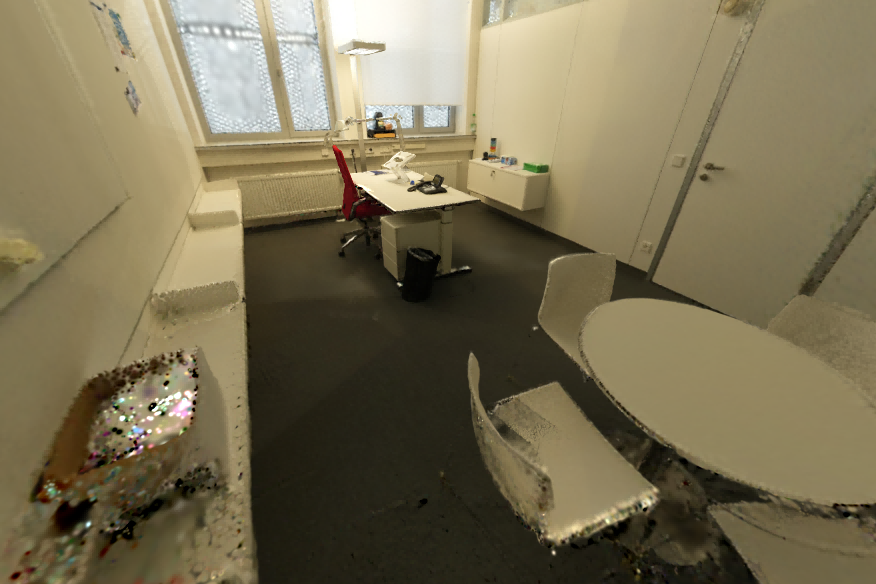} & 
        \includegraphics[width=0.16\linewidth]{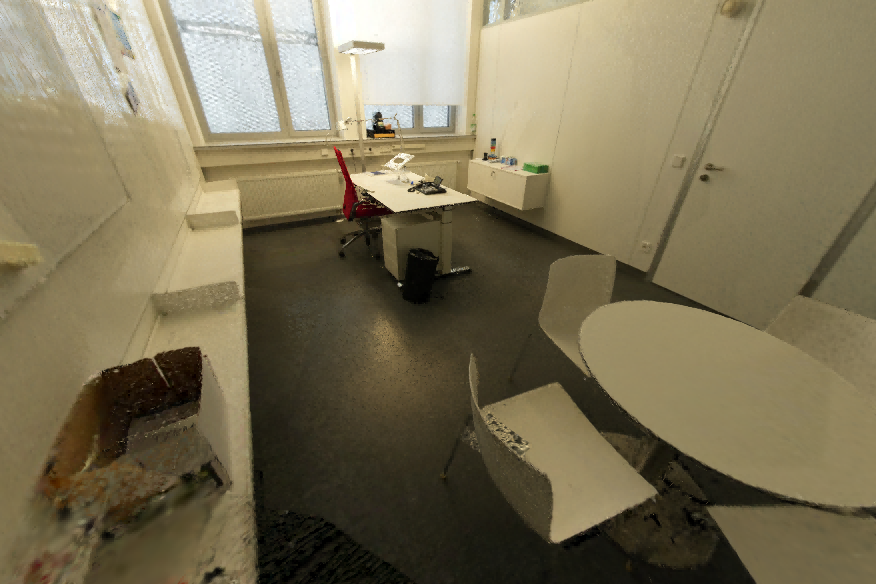} 
        \\
        
        & &
        \includegraphics[width=0.16\linewidth]{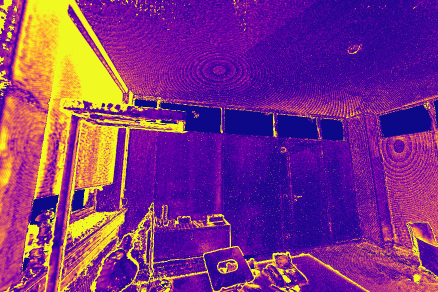} &
        \includegraphics[width=0.16\linewidth]{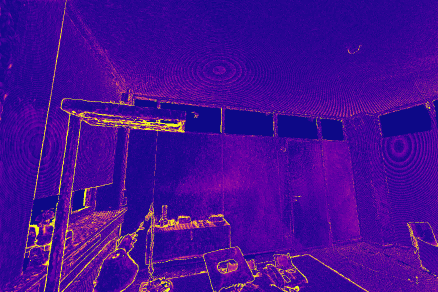} & &
        & \includegraphics[width=0.16\linewidth]{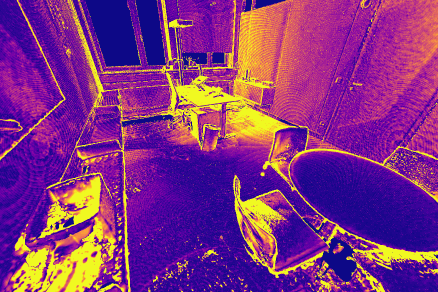} & \includegraphics[width=0.16\linewidth]{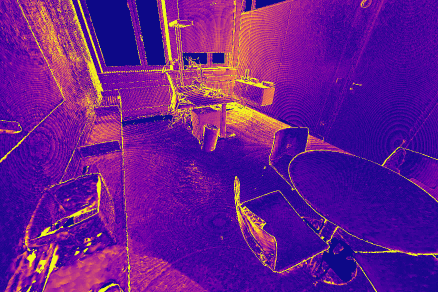}\\

        & Ground-truth & SplaTAM\cite{SplaTAM} & \textbf{GauS-SLAM(Ours)} & & Ground-truth & SplaTAM\cite{SplaTAM} & \textbf{GauS-SLAM(Ours)} \\
    \end{tabular}
    \caption{\textbf{NVS on ScanNet++\cite{scannetpp}.}}
    \label{fig:nvs}
    
\end{figure*}

\section{Novel View Synthesis}
\begin{figure*}[t]
    \centering
    \setlength{\tabcolsep}{1pt}
    \renewcommand{\arraystretch}{0.5}
    \small
    \begin{tabular}{ccc}
        
        \includegraphics[width=0.33\linewidth]{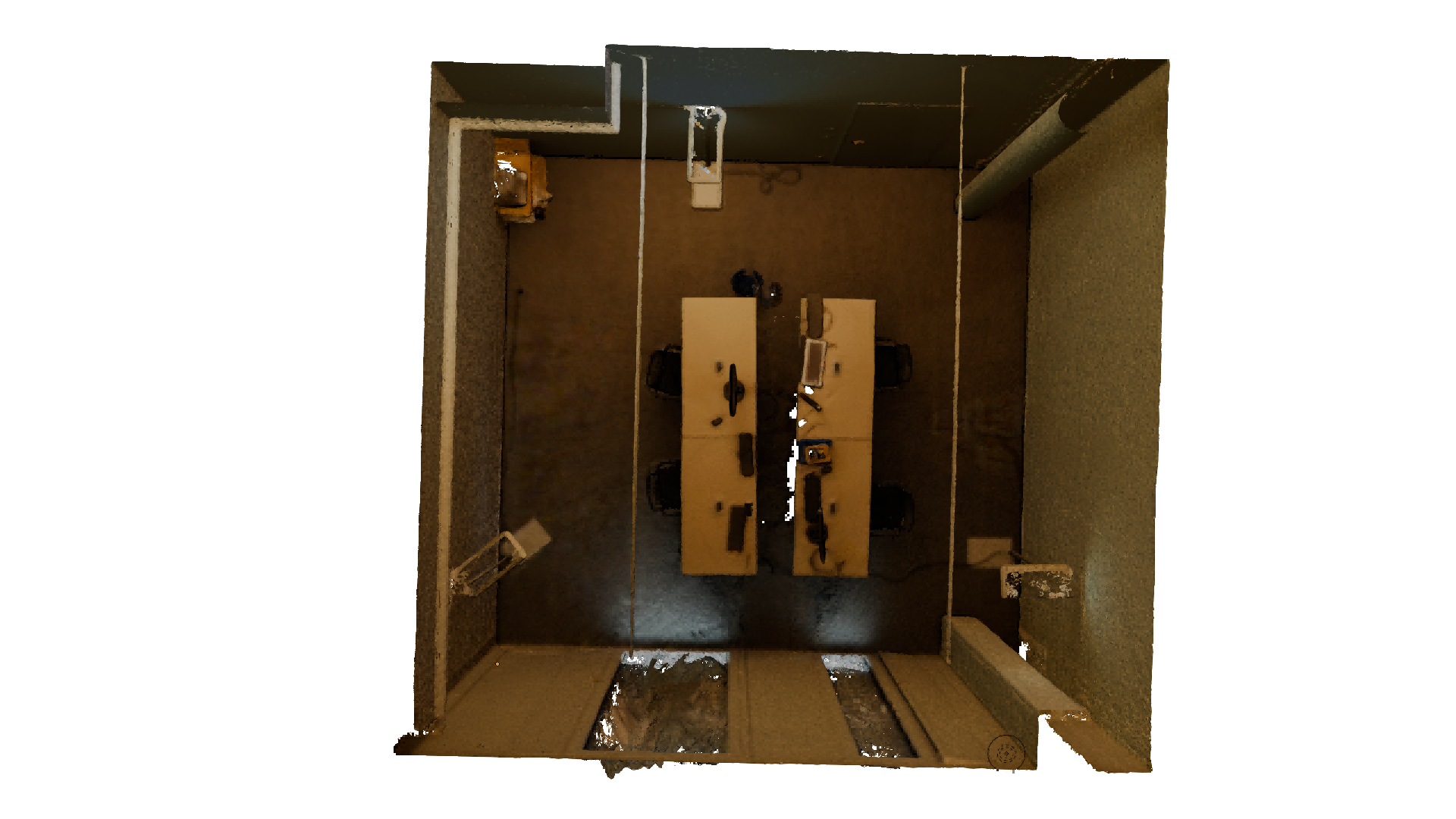} &
        \includegraphics[width=0.33\linewidth]{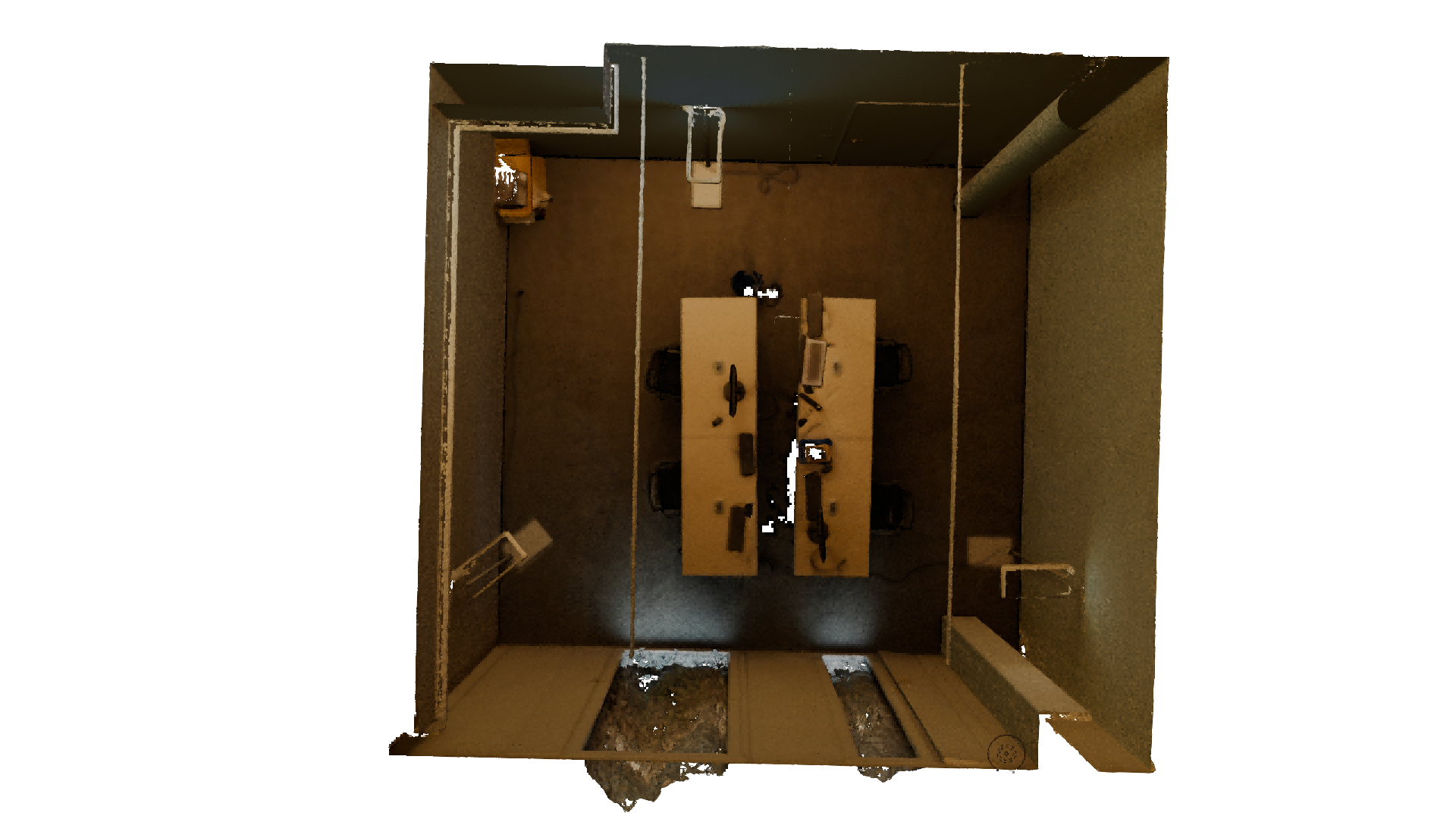} &  
        \includegraphics[width=0.33\linewidth]{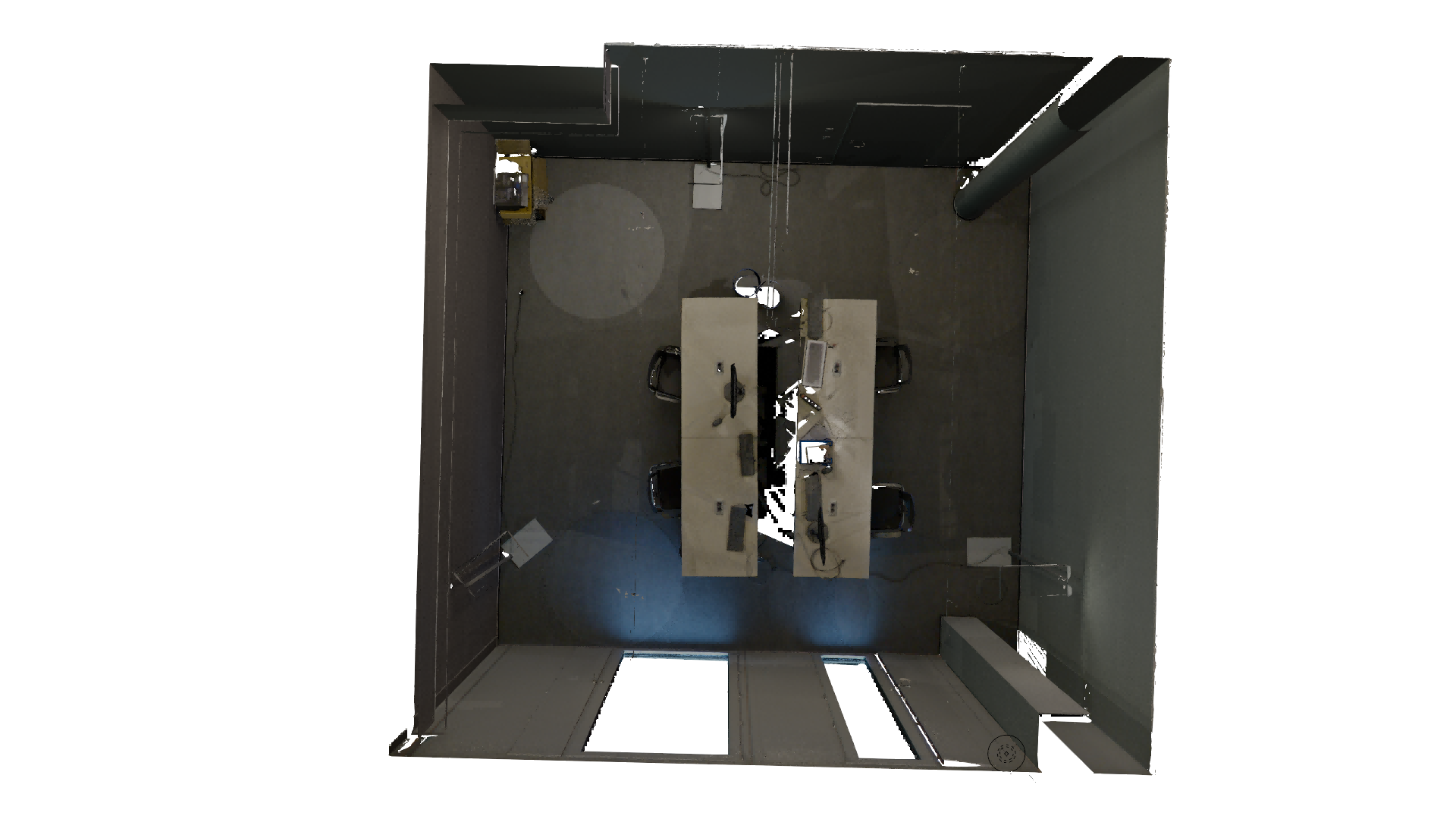} \\
        \includegraphics[width=0.33\linewidth]{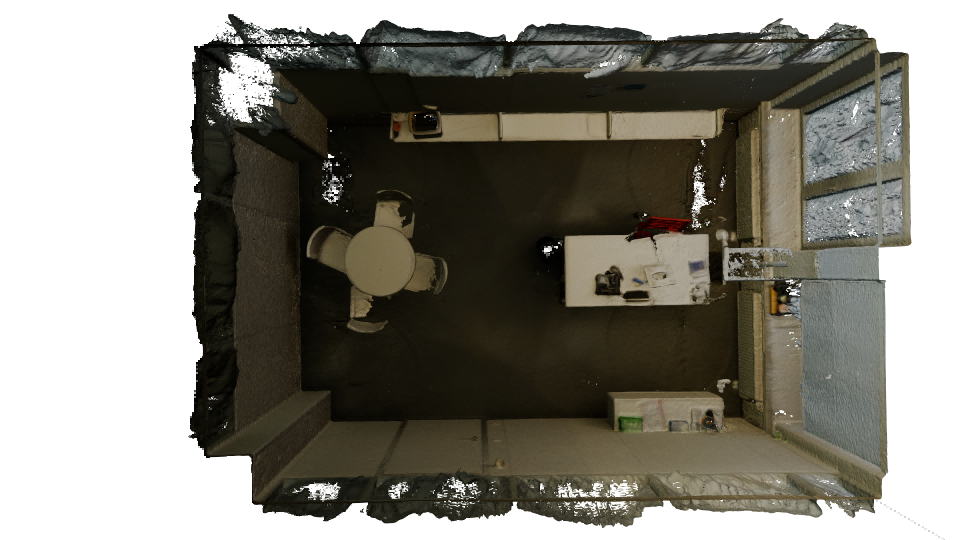} &
        \includegraphics[width=0.33\linewidth]{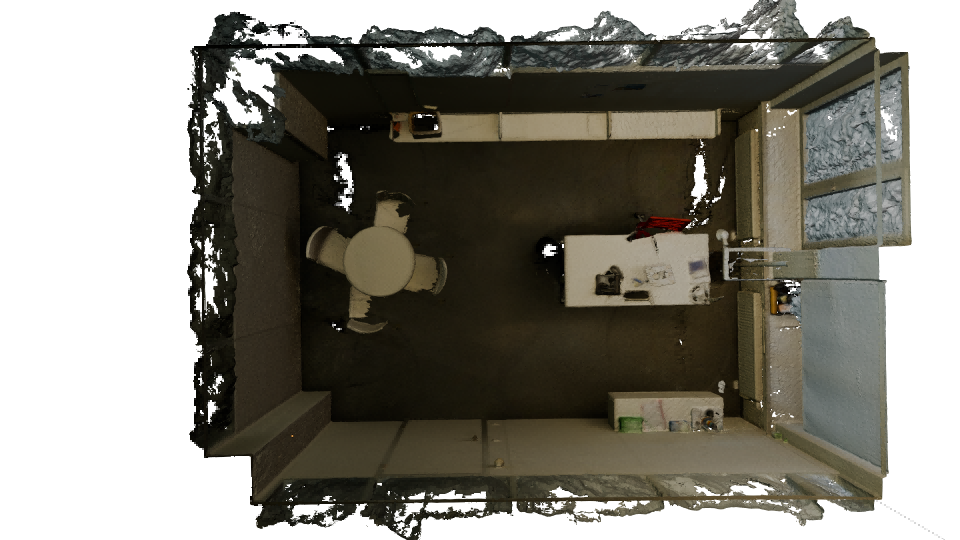} &  
        \includegraphics[width=0.33\linewidth]{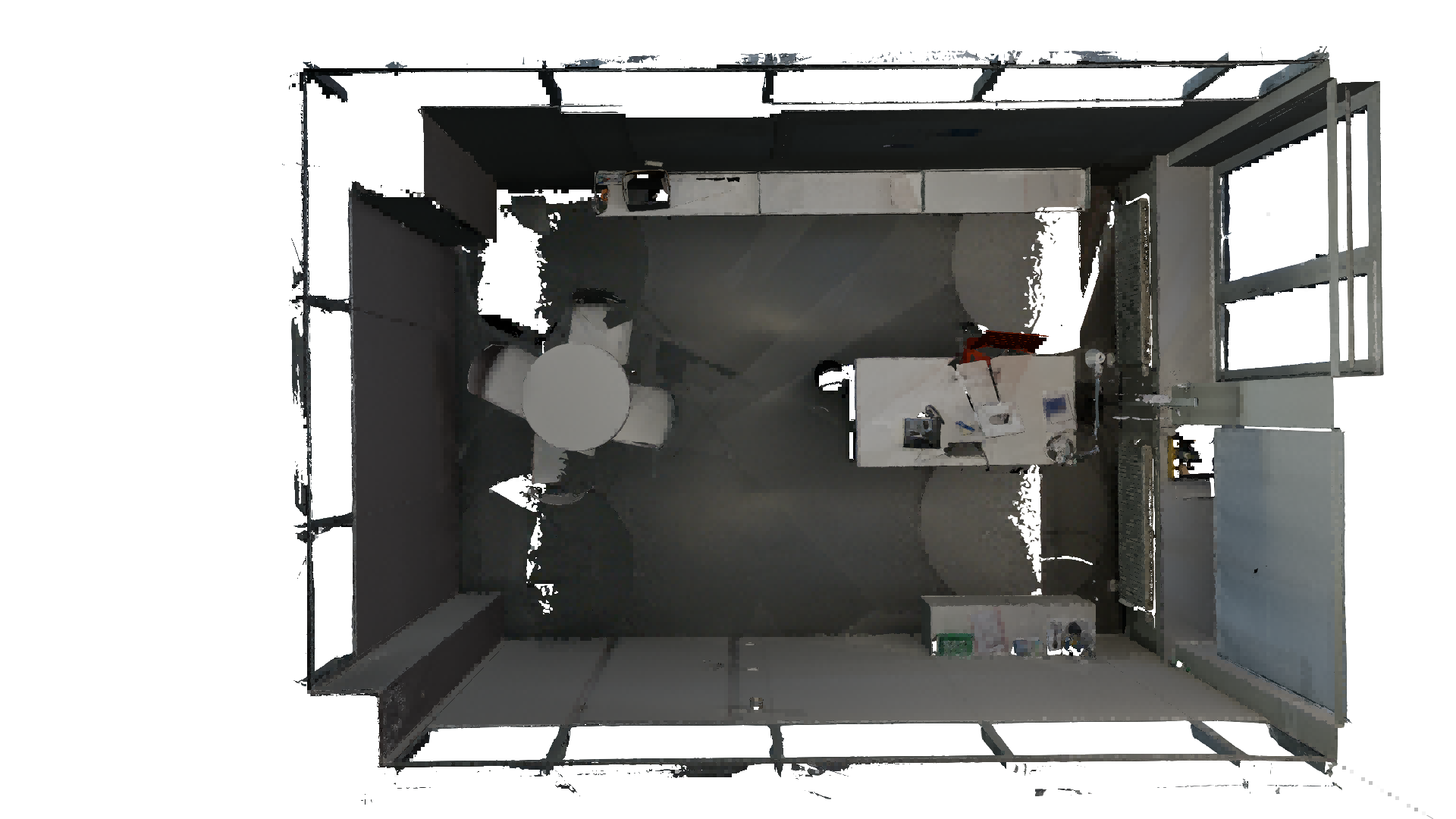} \\
        SplaTAM\cite{SplaTAM} & \textbf{GauS-SLAM(Ours)} & Ground-truth \\
    \end{tabular}
    \caption{\textbf{The comparison of mesh results in ScanNet++\cite{scannetpp}.}}
    \label{fig:scannet_mesh}
    
\end{figure*}

We evaluate novel view synthesis (NVS) performance on sequences \texttt{S1} and \texttt{S2} of the ScanNet++\cite{scannetpp} dataset, with results presented in \cref{tab:nvs}. Our method demonstrates superior performance over  SplaTAM\cite{SplaTAM} method in both rendering quality and depth map accuracy, which can be attributed to our improved pose estimation and depth rendering method. \cref{fig:nvs} provides a comparative visualization of rendering results from training and novel views, where our proposed Edge Growth strategy effectively handles regions with missing ground truth depth, particularly in window areas, demonstrating enhanced reconstruction capability. Owing to the enhanced accuracy of our depth maps, the extracted mesh exhibits superior quality, particularly in planar regions such as floors and walls, as shown in \cref{fig:scannet_mesh}.
\section{Additional experiments}
\paragraph{Experiments on Replica\cite{Replica}} \cref{tab: replica_tracking} and \cref{tab: replica_rendering} present comprehensive evaluations of tracking accuracy, rendering quality, and reconstruction performance across all sequences in the Replica dataset. Our proposed GauS-SLAM framework demonstrates state-of-the-art performance, achieving superior results on nearly every sequence compared to existing methods.
\begin{table}[!ht]
\centering
\resizebox{0.5\textwidth}{!}{
\begin{tabular}{lccccccc}
\toprule
\textbf{Methods} & 000 & 059 & 106 & 169 & 181 & 207 & \textbf{Avg.} \\ \midrule 
Point-SLAM\cite{Point-SLAM} & 10.2 & 7.8 & 8.6 & 22.1 & 14.7 & 9.5 & 12.1 \\
SplaTAM\cite{SplaTAM} & 12.8 & 10.1 & 17.7 & 12.1 & 11.1 & 7.5 & 11.8 \\
MonoGS \cite{MonoGS} & 9.8 & 32.1 & 8.9 & 10.7 & 21.8 & 7.9 & 15.2 \\
GLC-SLAM\cite{GLC-SLAM} & 12.9 & 7.9 & 6.3 & 10.5 & 11.0 & 6.3 & 9.2 \\ 
LoopSplat \cite{LoopSplat} & 6.2 & 7.1 & 7.4 & 10.6 & 8.5 & 6.6 & 7.7 \\

\textbf{GauS-SLAM(Ours)} & \cellcolor{top2}9.3 & \cellcolor{top1}7.1 & \cellcolor{top1}7.0 & \cellcolor{top1}7.4 & 17.5 & \cellcolor{top1}6.2 & \cellcolor{top3}11.5\\
\bottomrule
\end{tabular}
}
\caption{\textbf{Comparison of tracking performance on ScanNet\cite{ScanNet}.} Note that the GLC-SLAM\cite{GLC-SLAM} and LoopSplat\cite{LoopSplat} have loop closure correction. Results are taken from \cite{LoopSplat, GLC-SLAM}.}
\label{tab:scan-net-tracking-performance}
\end{table}

\begin{figure*}[ht]
    \centering
    \setlength{\tabcolsep}{1pt}
    \renewcommand{\arraystretch}{0.5}
    \small
    \begin{tabular}{cccc}
        
        \includegraphics[width=0.24\linewidth]{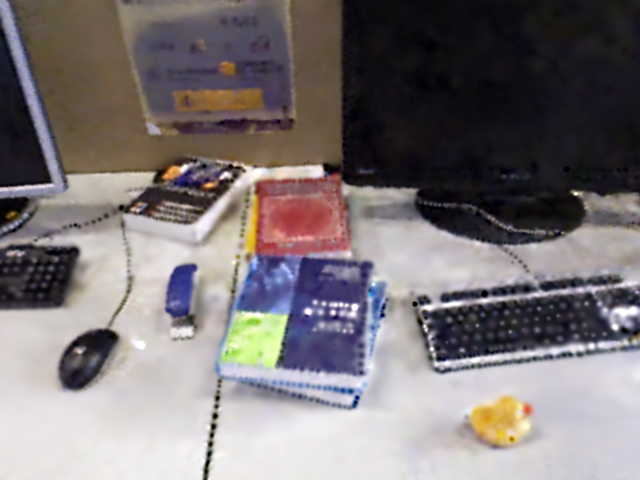} &
        \includegraphics[width=0.24\linewidth]{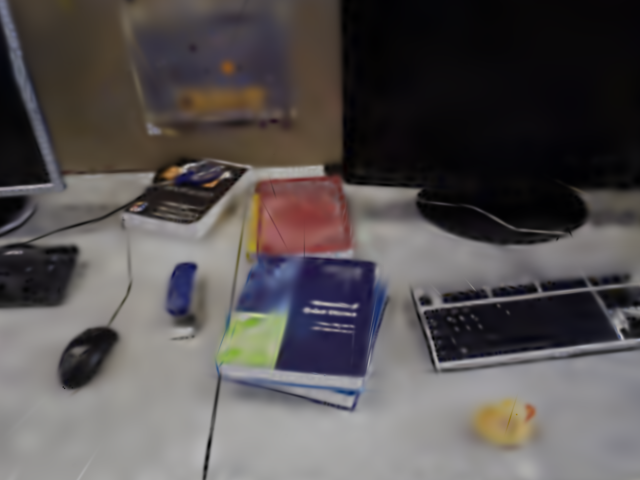} &  
        \includegraphics[width=0.24\linewidth]{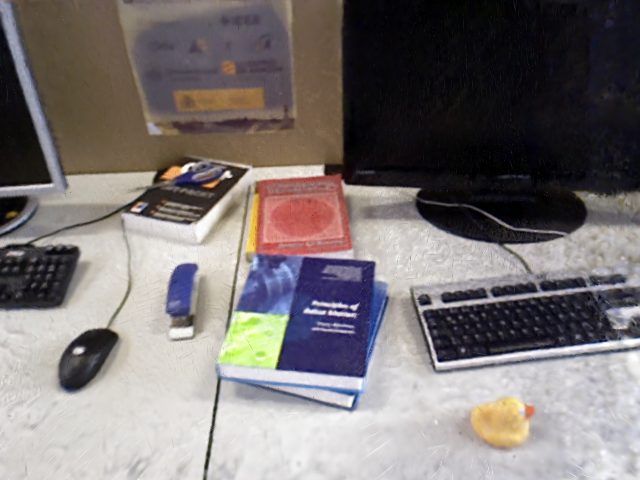} &
        \includegraphics[width=0.24\linewidth]{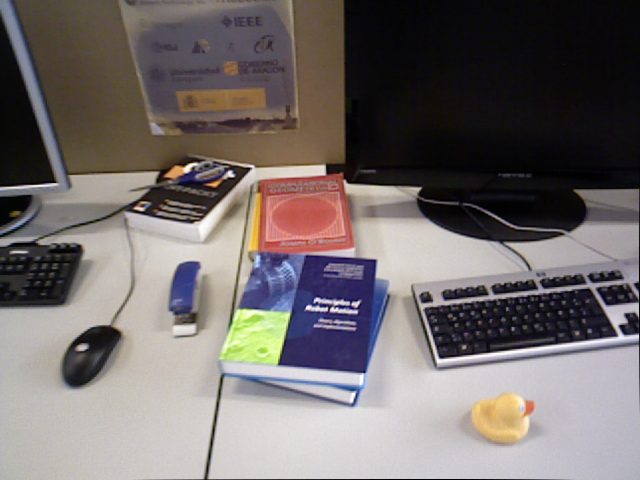} \\
        \includegraphics[width=0.24\linewidth]{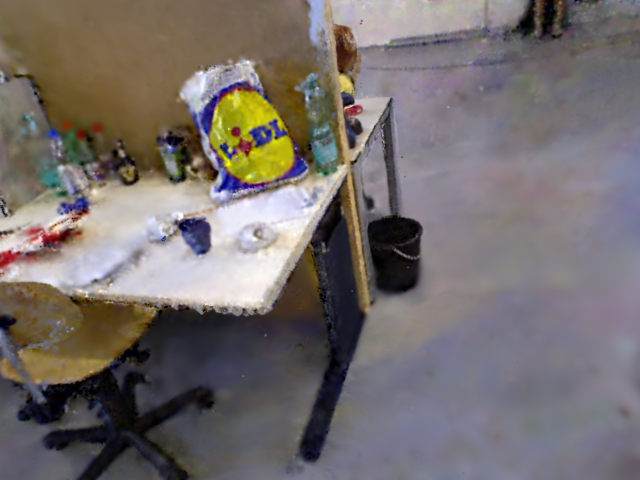} &
        \includegraphics[width=0.24\linewidth]{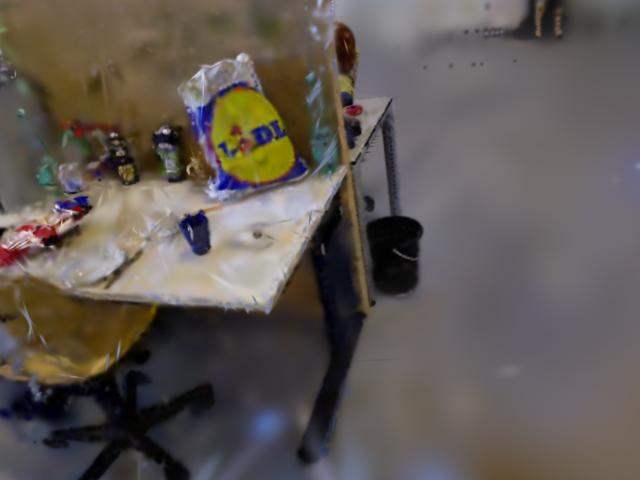} &  
        \includegraphics[width=0.24\linewidth]{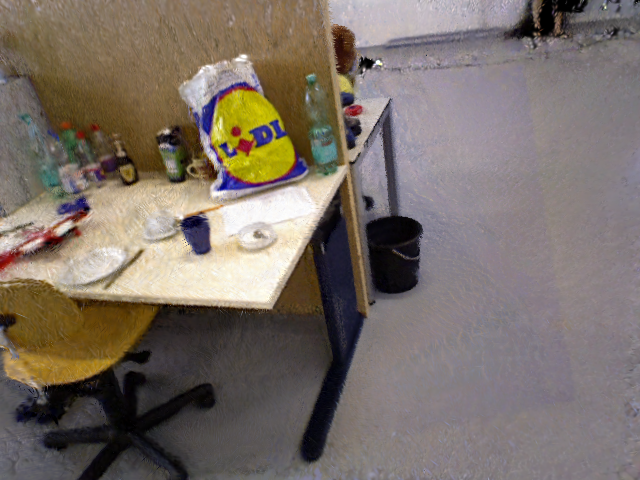} &
        \includegraphics[width=0.24\linewidth]{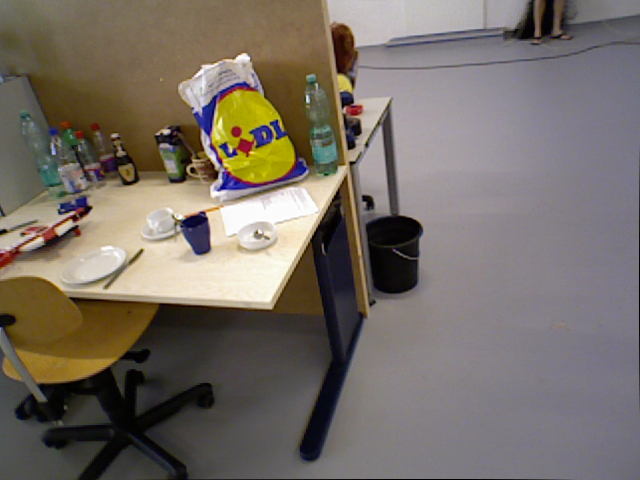} \\
        SplaTAM\cite{SplaTAM} & MonoGS\cite{MonoGS} & \textbf{GauS-SLAM(Ours)} & Ground-truth \\
    \end{tabular}
    \caption{\textbf{Rendering performance on TUM-RGBD\cite{TUM-RGBD}.}}
    \label{fig:tum_fr3}
\end{figure*}

\paragraph{Experiments on TUM-RGBD\cite{TUM-RGBD}.} \cref{tab:TUM_Rendering_performance} present an evaluation of tracking performance and rendering quality on TUM-RGBD\cite{TUM-RGBD}. The dataset's comparatively inferior depth map quality poses significant challenges for RGBD-based GauS-SLAM. While our proposed Edge Growth method can compensate for missing depth values, the absence of multiview constraints often leads to incorrect growth of Gaussians, directly impacting tracking performance. On the other hand, MonoGS\cite{MonoGS} and SplaTAM\cite{SplaTAM} employ isotropic 3D Gaussians, which help to maintain scene smoothness to some extent but result in the loss of fine details. As shown in \cref{fig:tum_fr3}, although our approach demonstrates a strong capability in modeling details, the reconstructed surfaces exhibit persistent high-frequency noise, making them appear visually cluttered.
\begin{table}[ht]
\centering
 
\resizebox{0.45\textwidth}{!}{
\begin{tabular}{lccccc}
\toprule
\textbf{Dataset} & \textbf{Metric} & \texttt{fr1/desk} & \texttt{fr2/xyz} & \texttt{fr3/office} & \textbf{Avg.} \\
\midrule
\multirow{4}{*}{Point-SLAM\cite{Point-SLAM}} & 
  PSNR$\uparrow$ & 13.91 & 17.23 & 18.11 & 16.41 \\
& SSIM$\uparrow$ & 0.626 & 0.707 & 0.756 & 0.696 \\
& LPIPS$\downarrow$ & 0.540 & 0.589 & 0.445 & 0.525 \\ 
&ATE-RMSE[cm]$\downarrow$ & 4.34 & 1.31 & 3.48 & 3.04 \\
\midrule
\multirow{4}{*}{SplaTAM\cite{SplaTAM}} & 
  PSNR$\uparrow$ & 22.07 & 24.55 & 21.62 & 22.74 \\
& SSIM$\uparrow$ & 0.851 & 0.930 & 0.816 & 0.865 \\
& LPIPS$\downarrow$ & 0.231 & 0.102 & 0.205 & 0.180 \\ 
& ATE-RMSE[cm]$\downarrow$ & 3.35 & 1.24 & 5.16 & 3.25 \\
\midrule
\multirow{4}{*}{MonoGS\cite{MonoGS}} & 
  PSNR$\uparrow$ & 23.73 & 24.75 & 24.96 & 24.28 \\
& SSIM$\uparrow$ & 0.786 & 0.836 & 0.840 & 0.820 \\
& LPIPS$\downarrow$ & 0.240 & 0.209 & 0.207 & 0.218 \\ 
& ATE-RMSE[cm]$\downarrow$ & 1.47 & 1.77 & 1.49 & 1.57 \\ 
\midrule
\multirow{4}{*}{\textbf{GauS-SLAM(Ours)}} & 
  PSNR$\uparrow$ & \cellcolor{top1}23.73 & \cellcolor{top1}27.72 & \cellcolor{top2}24.92 & \cellcolor{top1}25.45 \\
& SSIM$\uparrow$ & \cellcolor{top1}0.902 & \cellcolor{top1}0.958 & \cellcolor{top1}0.908 & \cellcolor{top1}0.922 \\
& LPIPS$\downarrow$ & \cellcolor{top1}0.231 & \cellcolor{top1}0.082 & \cellcolor{top1}0.198 & \cellcolor{top1}0.170 \\ 
& ATE-RMSE[cm]$\downarrow$ & \cellcolor{top2}1.82 & \cellcolor{top3}1.34 & \cellcolor{top1}1.47 & \cellcolor{top1}1.54 \\ 

\bottomrule
\end{tabular}
}
\caption{\textbf{Comparison of tracking and rendering performance on TUM-RGBD\cite{TUM-RGBD}}.}
\label{tab:TUM_Rendering_performance}
\end{table}

\paragraph{Experiments on ScanNet\cite{ScanNet}.}  \cref{tab:scan-net-tracking-performance} present an evaluation of tracking performance on the ScanNet\cite{ScanNet}. The ScanNet\cite{ScanNet} dataset presents significant challenges due to prevalent motion blur and exposure variations, which can severely compromise view consistency and lead to trajectory drift in SLAM systems. Despite these inherent difficulties, GauS-SLAM framework demonstrates superior tracking accuracy on specific sequences, outperforming even state-of-the-art methods incorporating loop closure correction mechanisms.

\paragraph{Experiments on ScanNet++\cite{scannetpp}} To ensure equitable benchmarking, we evaluate the first 250 frames for sequences: (\texttt{S1}) \texttt{b20a261fdf}, (\texttt{S2}) \texttt{8b5caf3398}, (\texttt{S3}) \texttt{fb05e13ad1}, (\texttt{S4}) \texttt{2e74812d00}, (\texttt{S5}) \texttt{281bc17764}. With tracking results detailed in \cref{tab:scannetpp-tracking-performance}. Due to the slight motion blur and the high-precision depth measurements, GauS-SLAM achieves millimeter-level tracking accuracy, reducing the trajectory error by approximately 84\% compared to the state-of-the-art method, LoopSplat\cite{LoopSplat}.

\begin{table}
\centering
\resizebox{0.5\textwidth}{!}{
\begin{tabular}{lcccccc}
\toprule
\textbf{Sequence} & \texttt{S1} & \texttt{S2} & \texttt{S3} & \texttt{S4} & \texttt{S5} & \textbf{Avg.}\\ 
\midrule
SplaTAM\cite{SplaTAM} &  1.50 & 0.57 & 0.31 & 443.10 & 1.58 & 89.41\\
MonoGS\cite{MonoGS} &  7.00 & 3.66 & 6.37 & 3.28 & 44.09 & 12.88\\
Gaussian SLAM\cite{Gaussian-SLAM} & 1.37 &  2.82 & 6.80 & 3.51 & 0.88 & 3.08 \\
LoopSplat \cite{LoopSplat} & 1.14 & 3.16  & 3.16 & 1.68 & 0.91 & 2.05 \\
\textbf{GauS-SLAM(Ours)} & \cellcolor{top1}0.41& \cellcolor{top1}0.38& \cellcolor{top1}0.16 & \cellcolor{top1}0.28 & \cellcolor{top1}0.34 & \cellcolor{top1}0.31\\
\bottomrule
\end{tabular}
}
\caption{\textbf{Tracking performance on ScanNet++\cite{scannetpp}(ATE RMSE$\downarrow$[cm]).} Results are taken from \cite{LoopSplat}.}
\label{tab:scannetpp-tracking-performance}
\end{table}

\begin{figure}
    \centering
    \setlength{\tabcolsep}{1pt}
    \renewcommand{\arraystretch}{0.5}
    \includegraphics[width=\linewidth]{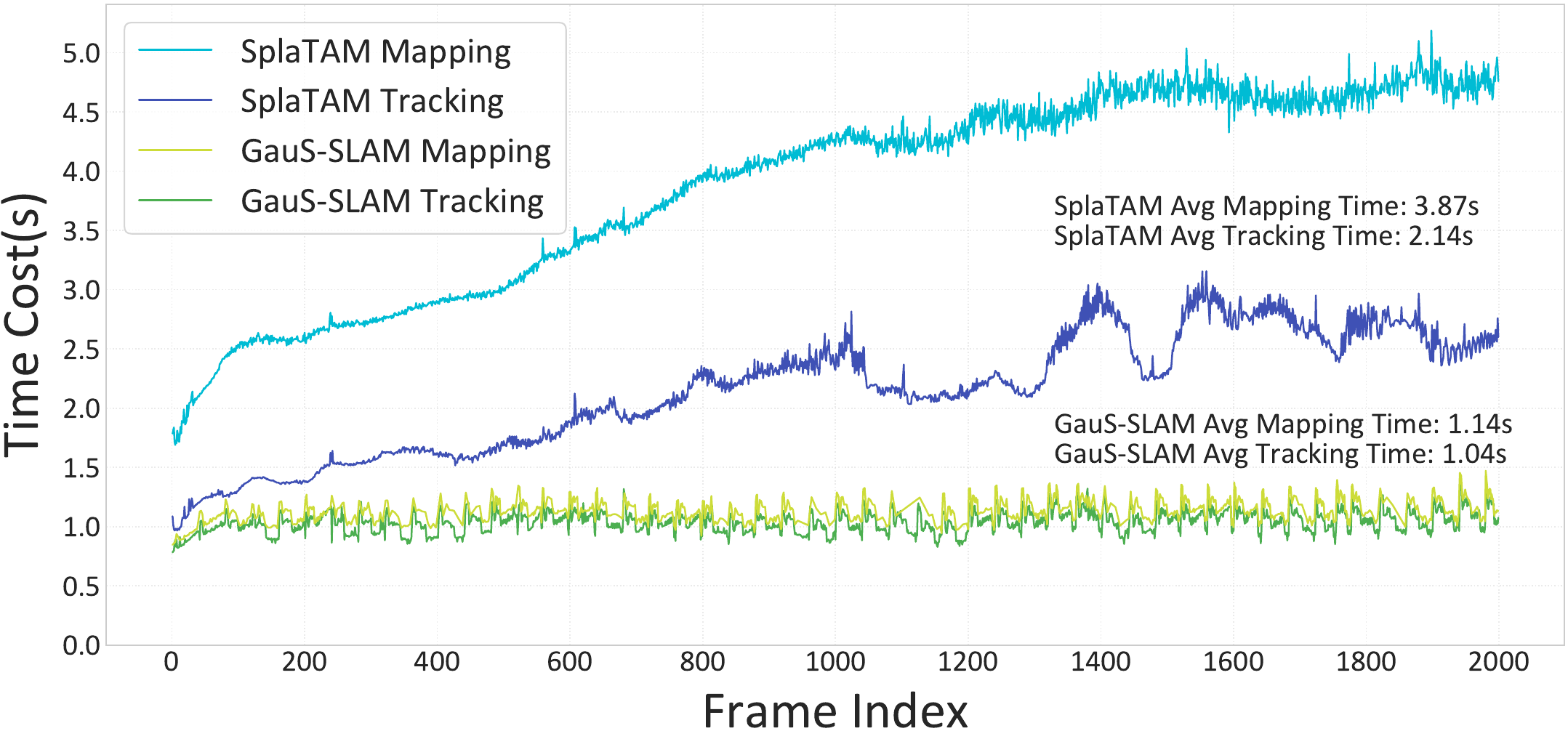} 
    \caption{\textbf{The comparison of system efficiency over time.} Notably, GauS-SLAM performs mapping operations exclusively on keyframes, and the mapping time of non-keyframe corresponds to the recent keyframe.}
    \label{fig:runtime_analysis}
\end{figure}

\paragraph{Runtime Analysis. } As shown in \cref{fig:runtime_analysis}, SplaTAM\cite{SplaTAM} experiences a decrease in the efficiency of tracking and mapping due to the continuous accumulation of Gaussian primitives in the global map. In contrast, GauS-SLAM periodically reset the local map, preventing degradation in tracking and mapping efficiency. 
\begin{table}
\centering

\label{tab:t1}
\resizebox{\linewidth}{!}{
\begin{tabular}{lcccccccccc}
\toprule
\textbf{Method} & \texttt{R0} & \texttt{R1} & \texttt{R2} & \texttt{O0} & \texttt{O1} & \texttt{O2} & \texttt{O3} & \texttt{O4} & \textbf{Avg.} \\ \midrule
\rowcolor{gray} \multicolumn{10}{l}{NeRF-based} \\
ESLAM\cite{ESLAM}
& 0.71 & 0.70 & 0.52 & 0.57 
& 0.55 & 0.58 & 0.72 & 0.63 & 0.63 \\
Point-SLAM\cite{Point-SLAM}
& 0.61 & 0.41 & 0.37 & 0.38 
& 0.48 & 0.54 & 0.69 & 0.72 & 0.52 \\
\midrule
\rowcolor{gray} \multicolumn{10}{l}{Gaussian-based (decoupled)} \\
RP-SLAM\cite{RP-SLAM}
& 0.43 & 0.38 & 0.53 & 0.36 
& 0.56 & 0.43 & 0.25 & 0.23 & 0.40 \\
DROID-Splat\cite{DROID-Splat}
& 0.34 & 0.13 & 0.27 & 0.25 
& 0.42 & 0.32 & 0.52 & 0.40 & 0.33 \\
HI-SLAM2\cite{HI-SLAM2}
& 0.23 & 0.22 & 0.19 & 0.23
& 0.27 & 0.25 & 0.37 & 0.33 & 0.26 \\
\midrule
\rowcolor{gray} \multicolumn{10}{l}{Gaussian-based (coupled)} \\
MonoGS\cite{MonoGS}
& 0.44 & 0.32 & 0.31 & 0.44 
& 0.52 & 0.23 & 0.17 & 2.25 & 0.58 \\
GS-SLAM\cite{GS-SLAM} 
& 0.48 & 0.53 & 0.33 & 0.52 
& 0.41 & 0.59 & 0.46 & 0.7 & 0.50 \\
SplaTAM\cite{SplaTAM}
& 0.31 & 0.40 & 0.29 & 0.47
& 0.27 & 0.29 & 0.32 & 0.55 & 0.36 \\ 
Gaussian-SLAM\cite{Gaussian-SLAM}
& 0.29 & 0.29 & 0.22 & 0.37
& 0.23 & 0.41 & 0.30 & 0.35 & 0.31 \\
GLC-SLAM\cite{GLC-SLAM}
& 0.20 & 0.19 & 0.13 & 0.31
& 0.13 & 0.32 & 0.21 & 0.33 & 0.23 \\
RTG-SLAM\cite{RTG-SLAM}
& 0.15 & 0.14 & 0.22 & 0.26 
& 0.25 & 0.21 & 0.19 & 0.12 & 0.19 \\
GS-ICP\cite{GS-ICP-SLAM}
& 0.15 & 0.16 & 0.11 & 0.18 
& 0.12 & 0.17 & 0.16 & 0.21 & 0.16 \\
\textbf{GauS-SLAM(Ours)} 
& \cellcolor{top1}0.06 & \cellcolor{top1}0.08 & \cellcolor{top1}0.08 & \cellcolor{top1}0.06 
& \cellcolor{top1}0.03 & \cellcolor{top1}0.09 & \cellcolor{top1}0.05 & \cellcolor{top1}0.05 & \cellcolor{top1}0.06\\ 
\bottomrule
\end{tabular}
}
\caption{\textbf{Comparison of tracking performance on Replica (ATE RMSE$\downarrow$[cm])}. Results are tacken from respective papers. } 
\label{tab: replica_tracking}
\end{table}

\begin{figure}
    \centering
    \includegraphics[width=\linewidth]{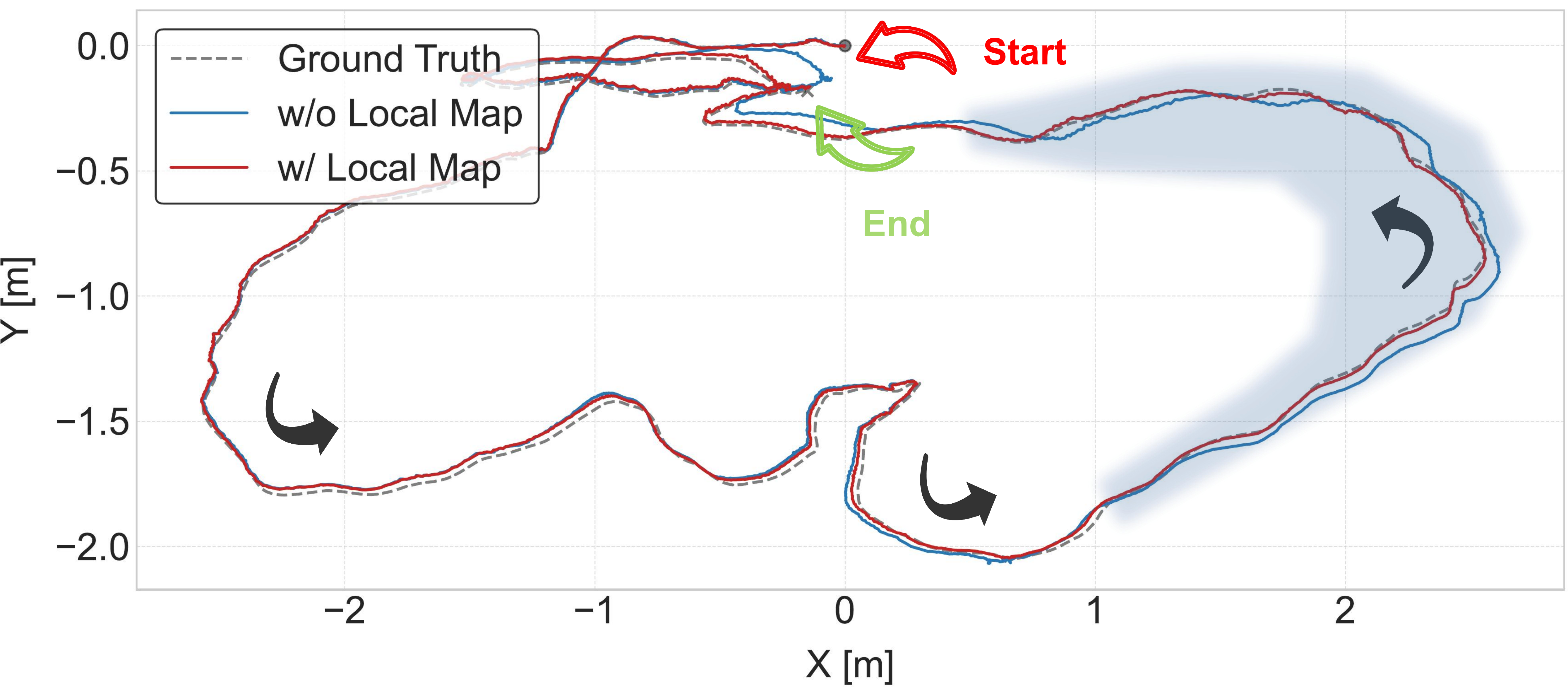}
    \caption{\textbf{The comparison of trajectories with and without local map.}. The local tracking method effectively avoids the influence of outlier regions in the blue area.}
    \label{fig:local_map_ablation}
\end{figure}

\paragraph{Ablation on Local map}
In the ablation study F presented in \cref{tab:ablation_study_on_vs_slam}, we investigate the impact of the structure of the scene without local mapping by conducting experiments on the \texttt{fr3/office} sequence in TUM-RGBD\cite{TUM-RGBD}. In this sequence, the camera follows a circular trajectory around a desk, as illustrated in \cref{fig:local_map_ablation}. The tracking error in the blue area is notably larger compared to the initial portion, which we attribute to the frequent occurrence of interference regions depicted in \cref{fig:mutliview_challenges}(b).



\begin{table*}[htbp]
    \centering
    \begin{tabular}{lcccccccccc}
    \toprule
    \textbf{Method} & \textbf{Metric} & \textbf{R0} & \textbf{R1} & \textbf{R2} & \textbf{O0} & \textbf{O1} & \textbf{O2} & \textbf{O3} & \textbf{O4} & \textbf{Avg.} \\ 
    \midrule
    \multirow{5}{*}{ESLAM\cite{ESLAM}} 
    & PSNR$\uparrow$& 25.25 & 25.31 & 28.09 & 30.33 & 27.04 & 27.99 & 29.27 & 29.15 & 27.80 \\
    & SSIM$\uparrow$ & 0.874 & 0.245 & 0.935 & 0.934 & 0.910 & 0.942 & 0.953 & 0.948 & 0.921 \\
    & LPIPS$\downarrow$ & 0.315 & 0.296 & 0.245 & 0.213 & 0.254 & 0.238 & 0.186 & 0.210 & 0.245 \\ 
    & Depth L1 [cm]$\downarrow$ & 0.970 & 1.070 & 1.280 & 0.860 & 1.260 & 1.710 & 1.430 & 1.060 & 1.180 \\
    & F1 [\%]$\uparrow$ & 81.00 & 82.20 & 83.90 & 78.40 & 75.50 & 77.10 & 75.50 & 79.10 & 79.10\\
    \midrule
    \multirow{3}{*}{Point-SLAM\cite{Point-SLAM}}
    & PSNR$\uparrow$ & 32.40 & 34.08 & 35.50 & 38.26 & 39.16 & 33.99 & 33.48 & 33.49 & 35.17 \\
    & SSIM$\uparrow$ & 0.974 & 0.977 & 0.982 & 0.983 & 0.986 & 0.960 & 0.960 & 0.979 & 0.975 \\
    & LPIPS$\downarrow$& 0.113 & 0.116 & 0.111 & 0.100 & 0.118 & 0.156 & 0.132  & 0.142 & 0.124 \\ 
    & Depth L1 [cm]$\downarrow$ & 0.530 & 0.220 & 0.460 & 0.300 & 0.570 & 0.490 & 0.510 & 0.460 & 0.440 \\
    & F1 [\%]$\uparrow$ & 86.90 & 92.31 & 90.78 & 93.77 & 91.62 & 88.98 & 88.22 & 85.55 & 89.77\\
    \midrule
    \multirow{5}{*}{MonoGS\cite{MonoGS}}
    & PSNR$\uparrow$    & 34.83 & 36.43 & 37.49 & 39.95 & 42.09 & 36.24 & 36.70 & 36.07 & 37.50\\
    & SSIM$\uparrow$    & 0.954 & 0.959 & 0.965 & 0.971 & 0.977 & 0.964 & 0.963 & 0.957 & 0.960\\
    & LPIPS$\downarrow$ & 0.068 & 0.076 & 0.075 & 0.072 & 0.055 & 0.078 & 0.065 & 0.099 & 0.070\\ 
    & Depth L1 [cm]$\downarrow$ & 0.793 & 0.561 & 0.914 & 0.702 & 1.099 & 0.951 & 0.968 & 1.661 & 0.956 \\
    & F1 [\%]$\uparrow$ & 79.92 & 83.60 & 81.26 & 86.51 & 82.18 & 78.20 & 79.86 & 58.01 & 78.69\\
    \midrule
    \multirow{5}{*}{SplaTAM\cite{SplaTAM}}
    & PSNR$\uparrow$    & 32.86 & 33.89 & 35.25 & 38.26 & 39.17 & 31.97 & 29.70 & 31.81 & 34.11\\
    & SSIM$\uparrow$    & 0.980 & 0.970 & 0.980 & 0.980 & 0.980 & 0.980 & 0.95 & 0.950 & 0.970\\
    & LPIPS$\downarrow$ & 0.070 & 0.100 & 0.080 & 0.090 & 0.090 & 0.100 & 0.120 & 0.150 & 0.100\\ 
    & Depth L1 [cm]$\downarrow$ & 0.425 & 0.364 & 0.516 & 0.414 & 0.646 & 0.980 & 1.234 & 0.609 & 0.648 \\
    & F1 [\%]$\uparrow$ & 89.95 & 89.13 & 88.86 & 92.32 & 90.01 & 86.10 & 79.17 & 80.96 & 87.06 \\
    \midrule
    \multirow{5}{*}{\textbf{GauS-SLAM(Ours)}} 
    & PSNR$\uparrow$    & \cellcolor{top1}38.04 & \cellcolor{top1}39.89 & \cellcolor{top1}40.25 & \cellcolor{top1}43.44 & \cellcolor{top2}41.22 & \cellcolor{top1}39.31 & \cellcolor{top1}39.55 & \cellcolor{top1}40.34 & \cellcolor{top1}40.25\\
    & SSIM$\uparrow$    & \cellcolor{top1}0.989 & \cellcolor{top1}0.991 & \cellcolor{top1}0.993 & \cellcolor{top1}0.994 & \cellcolor{top1}0.986 & \cellcolor{top1}0.993 & \cellcolor{top1}0.994 & \cellcolor{top1}0.992 & \cellcolor{top1}0.991\\
    & LPIPS$\downarrow$ & \cellcolor{top1}0.020 & \cellcolor{top1}0.030 & \cellcolor{top1}0.023 & \cellcolor{top1}0.022 & \cellcolor{top1}0.051 & \cellcolor{top1}0.023 & \cellcolor{top1}0.019 & \cellcolor{top1}0.029 & \cellcolor{top1}0.027\\ 
    & Depth L1 [cm]$\downarrow$ & \cellcolor{top1}0.343 & \cellcolor{top1}0.198 & \cellcolor{top1}0.451 & \cellcolor{top1}0.287 & \cellcolor{top1}0.365 & \cellcolor{top2}0.828 & \cellcolor{top2}0.736 & \cellcolor{top1}0.259 & \cellcolor{top1}0.433 \\
    & F1 [\%]$\uparrow$ & \cellcolor{top1}91.46 & \cellcolor{top2}92.09 & \cellcolor{top1}91.11 & \cellcolor{top2}93.59 & \cellcolor{top2}89.96 & \cellcolor{top1}89.14 & \cellcolor{top1}88.90 & \cellcolor{top1}88.47 & \cellcolor{top1}90.59 \\
    \bottomrule
    \end{tabular}
    \caption{\textbf{Rendering and reconstruction performance on Replica\cite{Replica}.}}
    \label{tab: replica_rendering}
    \end{table*}

%

\end{document}